\documentclass[Afour,sageh,times]{sagej}

\usepackage{moreverb,url}
\usepackage{array}
\usepackage{textcomp}
\usepackage{stfloats}
\usepackage{url}
\usepackage{verbatim}
\usepackage{graphicx}
\usepackage{amsmath}
\usepackage{tabularx} 
\usepackage{amsthm}
\usepackage{bm}
\usepackage{subcaption}

\usepackage{algorithm}
\usepackage{algpseudocode}
\usepackage{amssymb}
\usepackage{xcolor}

\newcommand{\trsp}{{\scriptscriptstyle\top}}

\newtheorem{theorem}{Proposition}
\newtheorem{lemma}{Lemma}
\newtheorem{corollary}{Corollary}
\newcommand{\minew}[1]{\textcolor{black}{#1}}

\newcommand{\mc}{\mathcal}

\usepackage[colorlinks,bookmarksopen,bookmarksnumbered,citecolor=red,urlcolor=red]{hyperref}

\newcommand\BibTeX{{\rmfamily B\kern-.05em \textsc{i\kern-.025em b}\kern-.08em
T\kern-.1667em\lower.7ex\hbox{E}\kern-.125emX}}

\def\volumeyear{2025}
\begin{document}

\runninghead{Xue et al.: Tensor Train Tree Search}

\title{Monte Carlo Tree Search with Tensor Factorization for Optimization Problems in Robotics}

\author{Teng Xue\affilnum{1, 2}, Yan Zhang\affilnum{1, 2}$^*$, Amirreza Razmjoo\affilnum{1, 2}$^*$, and Sylvain Calinon \affilnum{1, 2}}

\affiliation{\affilnum{1}Idiap Research Institute, Martigny, Switzerland\\
\affilnum{2}École Polytechnique Fédérale de Lausanne (EPFL), Lausanne, Switzerland \\
$^*$ Contributed equally to this work.
}

\corrauth{Teng Xue, Idiap Research Institute, Rue Marconi 19, 1920
Martigny, Switzerland.}

\email{teng.xue@idiap.ch}

\begin{abstract}

Many robotic tasks, such as inverse kinematics, motion planning, and contact-rich manipulation, can be formulated as optimization problems. Solving these problems requires addressing inherent nonlinear kinematics, complex contact dynamics, long-horizon correlations, and multimodal optimization landscapes, each posing distinct challenges for state-of-the-art optimizers. While existing methods tackle these issues through problem-specific strategies, such specialization inherently limits cross-task generalization, requires heavy engineering effort in problem reformulation, and hinders multi-task autonomy. Monte Carlo Tree Search (MCTS) offers a compelling framework that generalizes across diverse robotic tasks via strategic exploration of the solution space. However, it typically suffers from combinatorial complexity when applied naively, resulting in slow convergence and excessive storage space in high-dimensional domains. To address this limitation, we propose Tensor Train Tree Search (TTTS), which leverages tensor factorization to exploit implicit correlations among different branches within the decision tree. By utilizing the resulting compact, linear-complexity representation, TTTS significantly reduces both computation and storage overhead, thereby enabling highly efficient global decision making. Experimental results across inverse kinematics, motion planning around obstacles, legged robot manipulation, multi-stage motion planning, and bimanual whole-body manipulation demonstrate the efficiency of TTTS for generalized robot optimization over a diverse set of tasks.\\
Project website: \href{https://sites.google.com/view/tt-ts}{https://sites.google.com/view/tt-ts}.

\end{abstract}

\keywords{Monte Carlo Tree Search, Generalized Robot Optimization, Contact-rich Manipulation, Tensor Factorization}

\maketitle

\section{Introduction}

\begin{figure*}[htbp]
\centering
\includegraphics[width=1.0\textwidth]{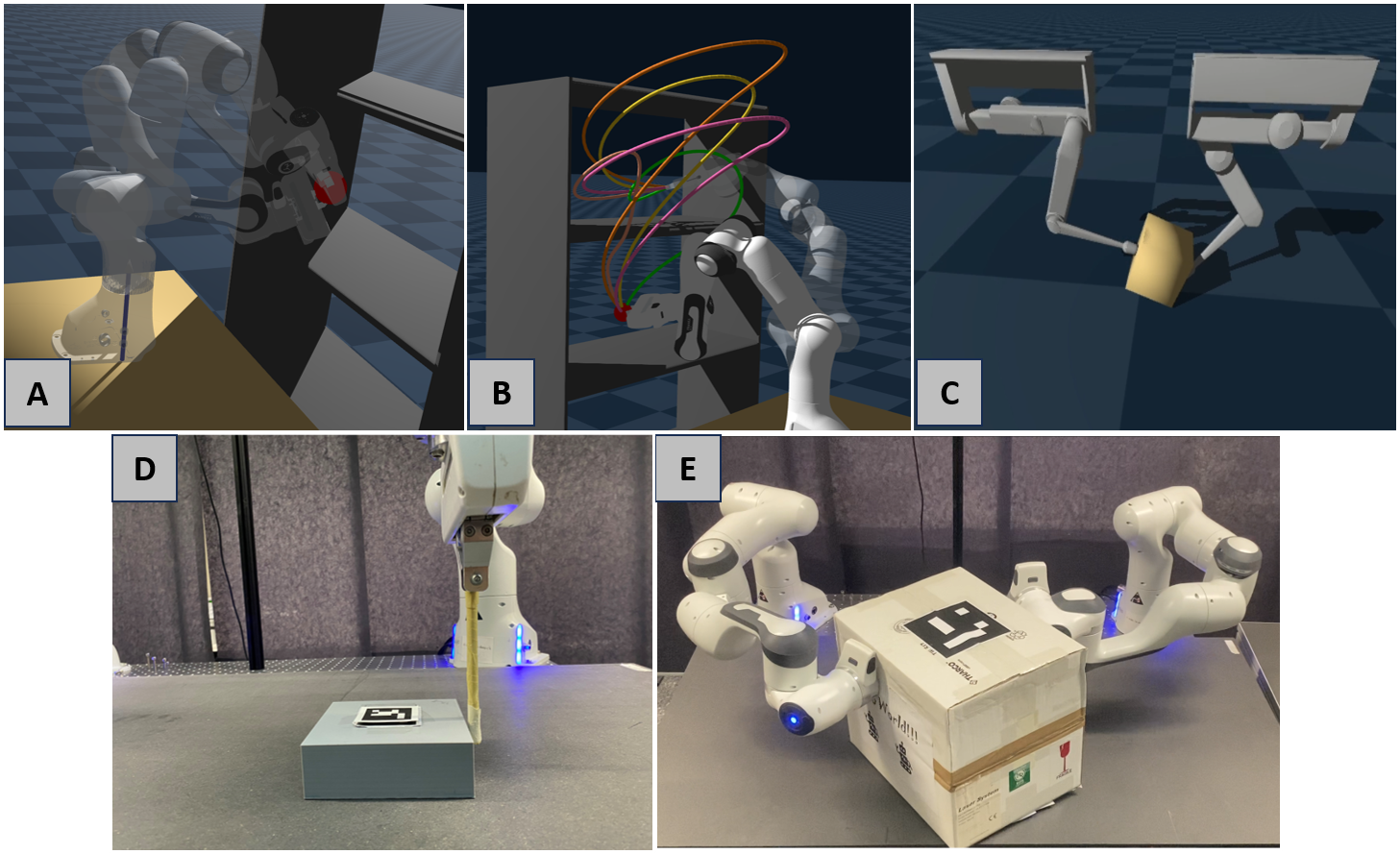}
\caption{\textbf{Overview of diverse applicable domains.} We demonstrate that TTTS is widely applicable in many tasks, such as Inverse Kinematics (A), Motion Planning (B), Legged Robot Manipulation (C), Multi-stage Motion Planning (D) and Bimanual Whole-body Manipulation (E).}
\label{fig:div_domain}
\end{figure*}

Optimization plays a critical role in robotics and serves as the foundation for a wide range of tasks, including inverse kinematics \citep{goldenberg2003complete}, obstacle avoidance \citep{marcucci2023motion}, multi-stage motion planning \citep{toussaint2015logic}, and contact-rich manipulation \citep{mason1986mechanics, mason1999progress}, see Figure~\ref{fig:div_domain}. Solving these problems requires addressing inherent nonlinear system dynamics, nonconvex constraints, joint reasoning over discrete contact modes and continuous motion trajectories, as well as complex interactions with the environment, each of which presents substantial challenges for state-of-the-art optimization methods. Specifically, the nonlinearity intrinsic to robotics renders many of these problems nonconvex, making gradient-based methods prone to getting trapped in poor local optima \citep{lembono2020memory}. Similarly, the need to jointly optimize over discrete modes and continuous motion introduces significant combinatorial complexity, substantially increasing computational costs and slowing convergence for both sampling-based \citep{lavalle2006planning, kavraki1996probabilistic} and search-based \citep{hart1968formal} approaches. Another key challenge is multimodal solution discovery, where multiple distinct feasible solutions may exist due to task redundancies or environmental symmetries. Identifying and reasoning over such diverse solutions is essential for both robustness and global optimization. To cope with these challenges, general formulations are often discarded to instead focus on a small set of problems by exploiting the specific structures. For example, trajectory optimization is often addressed through convex optimization with local linearization \citep{wang2017constrained, malyuta2022convex}. Obstacle avoidance is typically tackled using sampling-based methods, assuming that the environment contains a sufficiently large free space \citep{lin2017sampling}. Contact-rich manipulation, which involves significantly more challenging hybrid dynamics, is often tackled through local smoothing \citep{pang2023global} or the use of complementarity constraints \citep{moura2022non}.

Although these approaches are effective for individual tasks, they lack generalizability across domains. This limitation necessitates substantial human effort in problem design and imposes a heavy burden on problem decomposition in order to coordinate multiple specialized solvers for multi-task autonomy. Moreover, certain problems, such as multi-stage motion planning \citep{toussaint2015logic, Xue24ICRA} (also referred to as task and motion planning \citep{garrett2021integrated}), are inherently coupled and cannot be decomposed into independent subproblems that can be solved in isolation. In such cases, discrete mode sequencing (typically addressed with search-based methods) and continuous motion planning (typically addressed with gradient-based or sampling-based methods) must be considered jointly.

These challenges highlight the need for a general formulation for robot optimization. Monte Carlo Tree Search (MCTS) \citep{coulom2006efficient} offers a compelling framework: it eliminates reliance on gradient information, thereby avoiding poor local optima and making it well-suited for handling nonconvex constraints, nonlinear kinematics and dynamics. Moreover, by strategically exploring multiple branches of the search tree, MCTS is naturally capable of multimodal solution discovery, identifying diverse feasible solutions across different modes. Furthermore, it enables joint modeling of both discrete and continuous decision variables by discretizing the continuous domain into discrete grid points. This discretization, however, often results in excessively large trees and correspondingly slow convergence. To alleviate this issue, the spectrum of locally linearized controllability Gramians has been used in \citep{riviere2024monte} to efficiently decompose a continuous dynamical system into a few discrete sets. While effective in drones and ground vehicles, this approach may still incur substantial combinatorial complexity for high-dimensional nonlinear systems such as robotic manipulators, where local linearization becomes less reliable. Neural network heuristics have also been explored for guiding MCTS, and have shown impressive results on very challenging problems, such as Go \citep{silver2017mastering}, but they typically require large amounts of training data and substantial training time. Even when a simulator is available, obtaining informative training data that covers all modes of a multimodal landscape remains non-trivial. While such overhead may be justified for a single, well-defined, high-difficulty task, it becomes unnecessarily costly for the broad range of general robot optimization problems, which span tasks of moderate complexity, as shown in Figure~\ref{fig:div_domain}.

We argue that decision trees arising from domain discretization in robotics are typically not arbitrary, as different branches share substantial redundancy induced by shared robot kinematics, dynamics, and environmental constraints \citep{roy2021machine}. We propose to exploit this structure via tensor factorization, analogous to its use in quantum physics for capturing localized correlations \citep{eisert2010colloquium}. Specifically, we propose Tensor Train Tree Search (TTTS), which exploits the structure of the decision tree through tensor factorization, automatically identifying the most informative points without a pre-collected dataset or task-specific design, making it far more data-efficient than neural network heuristics and broadly applicable across diverse robotic tasks. This tree can be then expressed as a sequence of small third-order separable cores, each corresponding to a specific layer. In this way, the full decision tree can be represented using only linearly many parameters. This compact model provides Upper Confidence Bound (UCB) statistics for all nodes via efficient tensor contractions, directing search toward promising regions even from the first iteration and ensuring asymptotic completeness. Unlike conventional node- and table-based representations, the separable TT structure transforms the combinatorial complexity of tree search into linear time and space complexity, while also enabling full parallelization of MCTS.

Our main contributions are as follows:
\begin{enumerate}
\item We propose a general formulation for optimization problems in robotics that can handle nonlinear dynamical systems, nonconvex constraints, hybrid state-action spaces, and multimodal solution discovery.
\item We introduce the use of Tensor Train (TT) to approximate decision trees by exploiting their inherent structural redundancies, yielding a highly compact representation. This formulation enables fully parallelized tree search with linear time and space complexity, supporting efficient decision-making.
\item We provide theoretical analysis establishing TTTS's computational efficiency and asymptotic completeness for global optimum identification.
\item We evaluate TTTS against state-of-the-art methods and demonstrate its effectiveness on inverse kinematics, motion planning around obstacles, multi-stage motion planning, legged manipulation, and real-world bimanual whole-body manipulation tasks.
\end{enumerate}

\section{Related Work}
\label{sec:related}

\subsection{Gradient-based Methods}
Gradient-based and Newton-based methods (i.e., first and second-order optimization) are the most popular approaches in robot optimization, characterized by rapid convergence when gradients of the objective and constraints are available. Standard methods include Differential Dynamic Programming (DDP) \citep{mayne1966second,tassa2012synthesis}, iterative Linear Quadratic Regulators (iLQR) \citep{li2004iterative}, and trajectory optimization frameworks such as CHOMP \citep{ratliff2009chomp} and TrajOpt \citep{schulman2014motion}. The main advantage of gradient-based and Newton-based optimization is its efficiency in smooth and convex problem formulations, enabling practical deployment in real-time robot control scenarios. In these applications, gradients and Hessians provide useful local guidance, significantly speeding up convergence.

However, the critical limitation of gradient-based and Newton-based methods arises in scenarios involving highly nonlinear, nonconvex cost landscapes or discontinuous constraints, such as hybrid modes, contacts, and collision avoidance conditions \citep{xue2023guided, posa2014direct}. These methods are often too sensitive to initialization and often get trapped in poor local minima, demanding accurate initial guesses or sophisticated initialization schemes.

In contrast to these approaches, TTTS does not rely on convex structures and initial guesses. It can address highly nonlinear, nonconvex and multimodal landscapes, as well as mixed-integer decision variables. For example, the planar pushing task with face switching mechanism in \citep{xue2023guided} requires optimization on both discrete and continuous decision variables, which is not solvable with typical gradient-based and Newton-based methods. TTTS tackles this problem through TT approximation and strategic tree search, where gradient-based approaches can still be integrated within the framework as a refinement step for continuous variables, thereby combining the global exploration capability of tree search with the local optimization efficiency of gradient methods.

\subsection{Sampling-based Methods}
Sampling-based methods, including Rapidly-exploring Random Trees (RRT) \citep{lavalle2006planning} and Probabilistic Roadmaps (PRM) \citep{kavraki1996probabilistic}, offer an alternative to gradient-based approaches by constructing feasible paths through sampled configurations. These methods are particularly useful when gradient information is unavailable or difficult to obtain, and are widely used in motion planning and related robotics applications. Variants like RRT$^*$ \citep{karaman2011sampling}, BIT$^*$ \citep{gammell2015batch} and CMA-ES \citep{hansen2003reducing} further improve path optimality and convergence rates, demonstrating remarkable flexibility and scalability in challenging environments.

Nevertheless, a key limitation of sampling-based methods is that they are inherently tied to narrowly defined objectives, such as path length minimization, feasibility, or collision avoidance. Many robot optimization tasks, however, require optimizing general objective functions that go beyond geometric path planning, including multi-stage sequential costs, contact dynamics, and mixed-integer objectives. In such settings, sampling-based methods lack the structured guidance needed to efficiently explore the objective landscape, limiting their broader applicability.

In contrast, TTTS inherits from MCTS the ability to optimize a wide range of objective functions, making it broadly applicable to diverse robotic tasks beyond shortest-path problems. It leverages TT approximation to construct informative priors over the full tree and employs a UCB-based search that strategically identifies the global optimal solutions. In our framework, CMA-ES is integrated into TTTS to refine the solutions found by TT-Tree search and bridge the gap between the continuous decision domain and the discretized tree nodes.

\subsection{Search-based Methods}
Search-based planners, such as A$^*$ \citep{hart1968formal} and Monte Carlo Tree Search (MCTS) \citep{coulom2006efficient}, formulate planning as explicit search over discrete state or state-action spaces \citep{hart1968formal,browne2012survey}. These methods are particularly effective in environments with discrete or readily discretized action sets, and can offer strong guarantees under appropriate assumptions, such as optimality for A$^*$ with admissible heuristics or asymptotic convergence guarantees for variants of MCTS. Furthermore, methods like MCTS effectively balance exploration and exploitation, significantly improving planning efficiency in complex domains \citep{silver2017mastering}. Recent studies have also applied MCTS to contact-rich hybrid planning problems in robotic manipulation \citep{zhu2023efficient,cheng2023enhancing}, demonstrating its potential to handle challenging contact dynamics and improve dexterity through hierarchical exploration. 

While these works highlight the promise of MCTS in such domains, they also expose its limited scalability, largely due to the need to discretize continuous spaces. This discretization leads to exponential growth in computational complexity as dimensionality increases (i.e., the curse of dimensionality). Moreover, the discretization itself introduces approximation errors and scaling issues, rendering these approaches less effective or computationally prohibitive in high-dimensional continuous domains without carefully designed schemes or heuristics \citep{choset2005principles}. Rivi{\`e}re \emph{et al.} addresses this issue by decomposing continuous dynamical systems through the spectrum of the locally linearized controllability Gramian \citep{riviere2024monte}, but the discretization is achieved through local linearization, which can still lead to node explosion for highly nonlinear dynamical system, such as robot manipulator. Neural networks have also been utilized to guide MCTS to alleviate computational bottlenecks \citep{guez2018learning, silver2017mastering, kemmerling2024beyond}, showcasing remarkable results on challenging tasks such as Go. However, such networks inherently lack the flexibility to handle diverse, unseen scenarios without undergoing costly, per-task re-training, thereby limiting their practicality for generalized robot optimization.

TTTS leverages Tensor Train (TT) to reduce combinatorial complexity to linear complexity in both storage and computation. By exploiting the inherent tree structure, TT approximation is computationally efficient, providing global search guidance and enabling parallel rollouts for faster convergence. The strategic search from MCTS is preserved for asymptotic global convergence.

\subsection{Hybrid Methods}

Hybrid methods have also been developed that integrate multiple planning paradigms to overcome the limitations of purely gradient-based, sampling-based, or search-based methods. For example, combining trajectory optimization with sampling-based planners (such as RRT followed by trajectory optimization \citep{deits2015computing}) enables more effective exploration in high-dimensional spaces. In the context of mixed-integer programming, approaches like Logic-Geometric Programming (LGP) \citep{toussaint2015logic} alternate between discrete symbolic search and continuous geometric optimization. More recent hybrid methods combine global search strategies (e.g., MCTS or heuristic-based planners) with local refinement techniques or learning-based value estimation \citep{marcucci2024shortest, Xue24RSS}, demonstrating strong performance in solving complex planning problems that involve both discrete and continuous variables. By leveraging the strengths of both local optimization and global search, these methods achieve practical efficiency across a wide range of applications, including obstacle avoidance, contact-rich manipulation and legged locomotion \citep{anthony2017thinking, kim2020monte}.

Despite these benefits, hybrid methods often require careful engineering and tuning of parameters to balance computational resources effectively. Additionally, the integration of different planning paradigms introduces additional algorithmic complexity and implementation overhead, complicating both theoretical analysis and practical deployment.

TTTS can be seen as a hybrid approach that combines global search with local refinement, but different from the vanilla combination, TT provides a global approximation of decision tree with separable structure, enabling more efficient and parallelizable tree search. Moreover, different from the hierarchical framework that alternates between high-level discrete search and low-level continuous optimization, TTTS addresses the mixed-integer optimization jointly, by considering a joint distribution.

\section{Background}
\label{sec:background}

\subsection{Tensor Train Function Approximation}
A multivariate function $F(x_1,\ldots,x_d)$ defined on a Cartesian product domain $I_1 \times \cdots \times I_d$ can be discretized into a tensor $\bm{\mathcal{F}}$ 
by sampling it on a grid, where each entry is given by
\[
\bm{\mathcal{F}}_{i_1,\ldots,i_d} = F(x^{i_1}_1, \ldots, x^{i_d}_d), 
\quad i_k \in \{1,\ldots,n_k\}.
\]
The continuous function $F$ can then be approximated by interpolating the entries of $\bm{\mathcal{F}}$. However, direct storage and computation of high-dimensional tensors is infeasible due to their $\mathcal{O}(n^d)$ complexity. Analogous to matrix factorizations, tensor networks provide compact representations through factorization. In particular, the \emph{Tensor Train} (TT) decomposition expresses a $d$-dimensional tensor as a sequence of low-rank three-dimensional tensors, called \emph{cores}. In TT format, a tensor entry is represented as
\[
\bm{\mathcal{F}}(i_1,\ldots,i_d) 
= \bm{\mathcal{F}}^1_{:,i_1,:}\,\bm{\mathcal{F}}^2_{:,i_2,:}\cdots \bm{\mathcal{F}}^d_{:,i_d,:},
\]
where $\bm{\mathcal{F}}^k_{:,i_k,:} \in \mathbb{R}^{r_{k-1}\times r_k}$ 
denotes the $i_k$-th slice of the $k$-th core. TT decompositions are guaranteed to exist and can significantly reduce computational complexity \citep{oseledets2011tensor}. 

Algorithms such as TT-SVD~\citep{oseledets2011tensor} and TT-Cross~\citep{oseledets2010_ttcross1, savostyanov2011_ttcross2} provide efficient frameworks for computing and storing TT decompositions. TT-SVD relies on successive matricizations of the tensor and truncated singular value decompositions (SVDs), yielding quasi-optimal low-rank approximations but requiring access to the full tensor, which can be impractical in very high dimensions. In contrast, TT-Cross avoids the need for full tensor evaluation by constructing the decomposition from a relatively small, adaptively chosen set of tensor entries. The method is based on the principle of cross approximation, where one iteratively selects “skeleton” rows and columns in appropriate unfolding matrices. By exploiting the maximal-volume submatrices (\emph{maximum volume} principle), TT-Cross identifies the most informative entries of the tensor and uses them to construct the TT \emph{cores}. This process is carried out sequentially across tensor dimensions, updating ranks adaptively and ensuring numerical stability. As a result, TT-Cross can efficiently approximate the full tensor while querying and storing only a fraction of its entries, making it particularly well-suited for high-dimensional tensors (e.g., discretized function approximations). Crucially, the resulting TT model is a \emph{global} surrogate: once the TT cores are fitted from the sampled entries, the model can be evaluated at \emph{any} index tuple $(i_1,\ldots,i_d)$ in the full grid $\{1,\ldots,N_1\}\times\cdots\times\{1,\ldots,N_d\}$, not only at the entries queried during construction. This is analogous to fitting a regression model from a small sample and using it to predict at all points. For more details, please refer to \citep{oseledets2010_ttcross1, savostyanov2011_ttcross2}.


Given a discretized TT representation, it can be used to approximate continuous functions by interpolating across the TT \emph{cores}. For instance, by using linear interpolation between core slices, each core defines a matrix-valued interpolation:
\[
\bm{F}^k(x_k) = 
\frac{x_k - x_k^{i_k}}{x_k^{i_k+1} - x_k^{i_k}}\,\bm{\mathcal{F}}^k_{:,i_k+1,:} 
+ \frac{x_k^{i_k+1}-x_k}{x_k^{i_k+1}-x_k^{i_k}}\,\bm{\mathcal{F}}^k_{:,i_k,:},
\]
valid for $x^{i_k}_k \le x_k \le x^{i_k+1}_k$. 
The resulting continuous approximation then takes the form
\[
F(x_1,\ldots,x_d) \approx \bm{F}^1(x_1)\cdots \bm{F}^d(x_d),
\]
allowing efficient representation and approximation of functions defined on mixed discrete–continuous domains.

\subsection{Global Optimization via Tensor Train (TTGO)}

The goal of global optimization is to identify decision variables $\bm{x}$ that maximize a target function $f(\bm{x})$. TTGO \citep{shetty2016tensor} addresses this task by mapping $f(\bm{x})$ into an unnormalized density function $F(\bm{x})$ through a monotone transformation that preserves the ordering of optima. The function $F(\bm{x})$ is then approximated in Tensor Train (TT) format using the TT-Cross algorithm, resulting in a compact, structured representation:
\begin{align}
    &F(x_1, \dots, x_d) 
    \approx \sum_{\gamma_1=1}^{r_1} \sum_{\gamma_2=1}^{r_2} \cdots \sum_{\gamma_{d-1}=1}^{r_{d-1}} \notag \\
    &\quad \bm{\mc{F}}^1(1,\, x_1,\,\gamma_1)\,\bm{\mc{F}}^2(\gamma_1,\, x_2,\,\gamma_2)\,\cdots\,\bm{\mc{F}}^d(\gamma_{d-1},\, x_d,\,1),
\end{align}
where each TT core $\bm{\mc{F}}^k$ is a tensor of size \(r_{k-1} \times N_k \times r_k\), with \(N_k\) denoting the discretization resolution of the $k$-th variable and $r_k$ the associated TT ranks.

This TT representation yields a low-rank surrogate that enables efficient optimization. Instead of exhaustive grid search, which suffers from exponential complexity \(\mathcal{O}(N^d)\), TTGO leverages the tensor structure to perform coordinated dimension-wise search, reducing the computational cost to \(\mathcal{O}(Ndr^2)\), where $N$ is the typical discretization size per dimension and $r$ the maximal TT rank. However, TTGO relies heavily on accurate TT approximation, which may not hold under limited storage capacity.

\subsection{Monte Carlo Tree Search}

Given a decision tree, MCTS begins at the root and selects a promising node at each layer by balancing exploitation and exploration, typically via an upper confidence bound (UCB):

\begin{equation}
i_{[j]} \gets \arg\max\nolimits_{i_{[j]} \in \mc{I}_{[j]}} \left( \frac{q_{i_{[j]}}}{v_{i_{[j]}}} + c \sqrt{\frac{\log v_{i_{[j-1]}}}{v_{i_{[j]}}}} \right),
\label{eq:ucb}
\end{equation}
\noindent where $j$ denotes the current depth in the tree, and $i_{[j]} = (i_1, i_2, \cdots, i_j)$ represents the complete sequence from the root to the selected node. $\mc{I}$ denotes the index set of all complete node sequences corresponding to branches in the decision tree, and $\mc{I}_{[j]}$ denotes the subset of index sequences truncated at depth $j$, i.e., all partial paths from the root to depth $j$. Here, $q_{i_{[j]}}$ and $v_{i_{[j]}}$ denote the cumulative reward and visit count of the terminal node in path $i_{[j]}$, while $v_{i_{[j-1]}}$ refers to the visit count of its parent. The constant $c$ regulates the trade-off between exploration and exploitation.

After selection, MCTS expands the chosen node by generating a new child, followed by a simulation stage in which a rollout policy estimates the outcome from that child. The simulation results are then propagated back through the visited nodes, updating their value estimates and visit counts to guide subsequent searches. Through strategic exploitation and exploration, MCTS can efficiently explore large search spaces and asymptotically converge to the global optimum.

To support strategic searching behavior, it is essential to store and update the value and visit count for each node. This leads to a combinatorial complexity of $\mathcal{O}(N^d)$ in both computation time and memory usage, where $N$ is the number of nodes per layer and $d$ is the depth of the tree. Such complexity severely limits the applicability of MCTS in many tasks.

\section{Problem Formulation}
\label{sec:problem}


We consider a general robot optimization formulation that can address diverse problems in robotics:
\begin{align}
\max_{\bm{x} \in \Omega_x} \; & J(\bm{x}), \label{eq:general_opt} \;
\text{s.t.} \;  \phi(\bm{x}) \leq 0, \;
                  \psi(\bm{x}) = 0,
\end{align}
where $\bm{x}$ is the \emph{decision variable}, $\Omega_x$ is the \emph{decision domain}, $J: \Omega_x \to \mathbb{R}$ is the objective function, and $\phi(\bm{x}) \leq 0$, $\psi(\bm{x}) = 0$ are inequality and equality constraints encoding system dynamics and other physical limits. We assume that $J$ is \emph{evaluable} and \emph{deterministic}: given the same $\bm{x}$, $J(\bm{x})$ always returns the same value, with no assumption on gradient, convexity, or closed-form structure. Any black-box function that can be queried suffices, including those evaluated through black-box physical simulators. The decision domain $\Omega_x$ is flexible and may take different forms:

\begin{itemize}
\item \textbf{Discrete:} $\bm{x}$ is drawn from a finite set (e.g., combinatorial sequences, integer choices), such as hybrid contact modes (i.e., push, pivot) in contact-rich manipulation tasks.
\item \textbf{Continuous:} $\bm{x} \in \mathbb{R}^n$ represents the continuous decision variables to be optimized (e.g., in trajectory optimization). The coarse grid solution can subsequently be refined by a local continuous optimizer.
\item \textbf{Hybrid:} $\bm{x} = (\bm{x}_d, \bm{x}_c)$ combines discrete choices and continuous variables, which corresponds to the multi-stage manipulation problems that require joint reasoning over discrete modes and continuous motion trajectories.
\end{itemize}

To handle all three cases in a unified manner, we keep the discrete domain unchanged and discretize the continuous domain onto a finite grid of $N_j$ candidate values, thereby reducing all domain types to the same discrete structure. The resulting index set $\mathcal{I} \subseteq \{1,\ldots,N_1\}\times\cdots\times\{1,\ldots,N_d\}$ defines a $d$-layer decision tree with branching factor $N_j$ at layer $j$: each root-to-leaf path encodes a candidate solution $\bm{x}$ with associated value $J(\bm{x})$. The optimization problem thus reduces to finding the maximum-value branch, a tree search problem over $\mathcal{O}(N^d)$ leaves.

MCTS is a natural and powerful approach for solving this problem. It selectively searches branches in a strategic manner, but suffers from the combinatorial complexity of the underlying tree: both memory and search time scale as $\mathcal{O}(N^d)$, becoming intractable as depth $d$ grows.

\section{Tensor Train Tree Search}
\label{sec:ttts}

\begin{figure*}[htbp]
  \centering
  \includegraphics[width=0.9\textwidth]{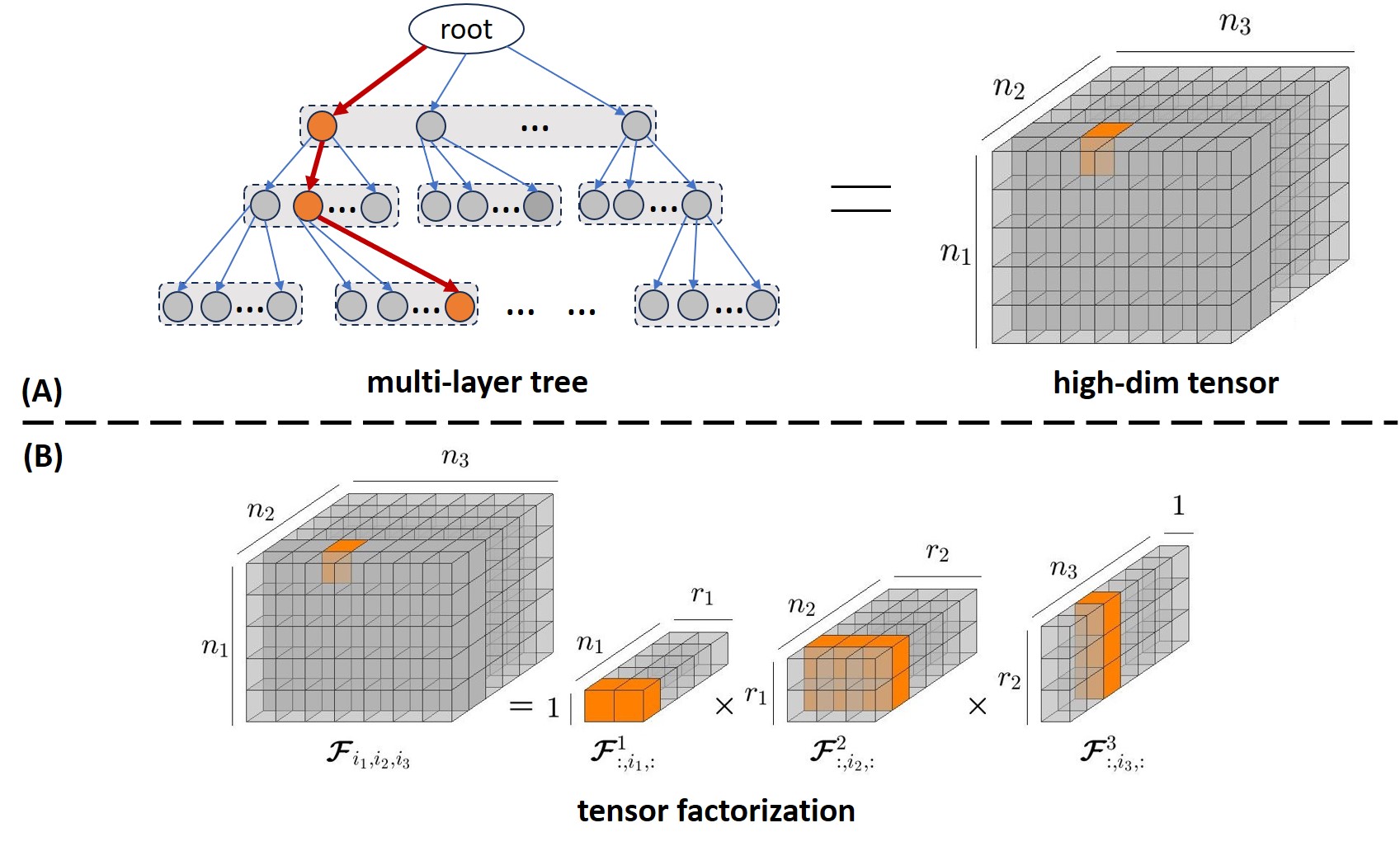}
  \caption{\textbf{Tree-Tensor-TT transformation.} \textbf{(A)} A multi-layer decision tree can be equivalently represented as a high-dimensional tensor, where each tensor element corresponds to the value at the terminal node of a branch. \minew{The figure shows a tree of depth three, in which each layer corresponds to one decision variable and hence to one tensor dimension. A tree of depth $d$ thus corresponds to a $d$-dimensional tensor.} \textbf{(B)} Tensor decomposition in TT format. The TT cores provide a compact, global surrogate for this entire tensor: any leaf value can be approximated by contracting the corresponding core slices, enabling efficient UCB computation over the full tree without explicitly storing all $N^d$ entries.}
  \label{fig:tree_tensor}
\end{figure*}

To reduce this combinatorial complexity, our objective is to express the decision tree with linear complexity $\mathcal{O}(\lambda N d)$, where $\lambda$ is a scaling factor. This is equivalent to representing a high-dimensional tensor of size $\mathcal{O}(N^d)$ using a compact, low-rank decomposition, which is a well-studied problem in numerical mathematics \citep{cichocki2016tensor}. Specifically, we employ tensor train (TT) to exploit the separable structure and reduce inter-layer dependencies. As a result, the original high-dimensional tensor can be efficiently represented using significantly fewer parameters through a sequence of third-order TT cores, namely:
\begin{equation}
	\label{eq:tt_rep}
	\widehat{\bm{\mc{T}}}_{(i_{1},\ldots,i_{d})} \approx \bm{\mc{T}}^1_{:,i_1,:}\bm{\mc{T}}^2_{:,i_2,:}\cdots \bm{\mc{T}}^d_{:,i_d,:},
\end{equation}
where $\widehat{\bm{\mc{T}}}$ denotes the original high-dimensional tensor, and $\bm{\mc{T}}^j$ is the core corresponding to the $j$-th dimension (i.e., the $j$-th layer of the tree). Figure \ref{fig:tree_tensor} (B) presents an illustrative example of a three-dimensional tensor.

Algorithm~\ref{alg:TTTS} presents the pseudocode of the proposed method. \minew{It consists of three phases, which correspond to the three blocks of Algorithm~\ref{alg:TTTS}. (i)~\emph{Offline TT-tree initialization}: TT-Cross approximates the objective in TT format, yielding a compact representation of the node values of the entire decision tree. (ii)~\emph{Online TT-tree search}: a task-conditioned TT model is obtained for the given problem instance, followed by an MCTS-style tree search with node values and visit counts computed via TT contractions. (iii)~\emph{Refinement}: the best solutions found on the grid are refined by a local continuous optimizer. The remainder of this section describes the TT representation of the tree used in phases~(i) and~(ii), the individual steps of the online search, and the refinement.}

\begin{figure*}[htbp]
  \centering
  \includegraphics[width=0.8\textwidth]{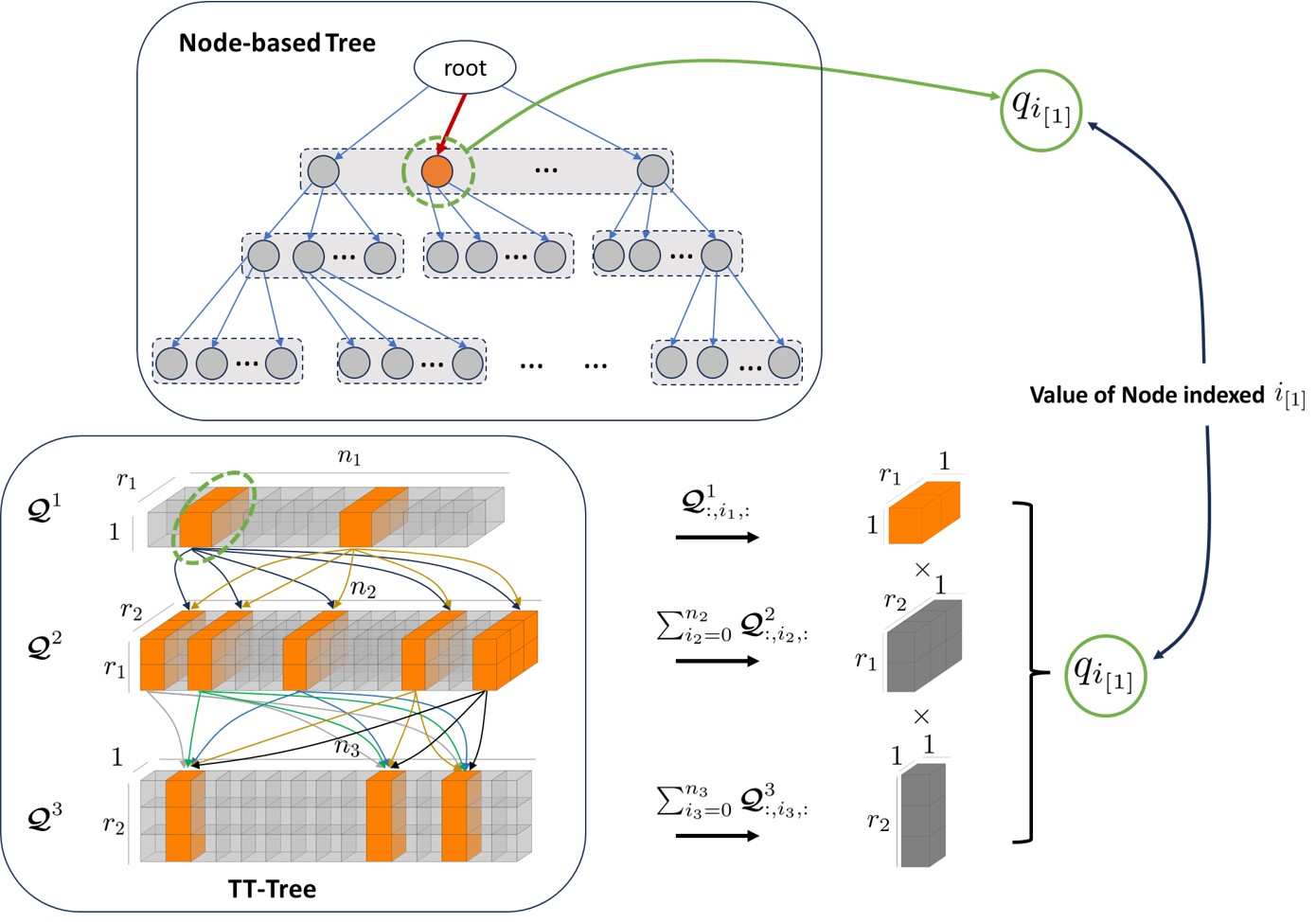}
  \caption{\textbf{Node value computation for a tree represented in TT format.} This example illustrates how the value of a first-layer node is computed using TT cores. The value of node $i_1$ is obtained by fixing the slice $\bm{\mc{Q}}^1_{:,i_1,:}$ from the first TT core and summing over all indices of the remaining cores ($\bm{\mc{Q}}^2$ through $\bm{\mc{Q}}^3$) to aggregate all leaf completions, as in Eq.~\eqref{eq:w_compute}.}
  \label{fig:tree_node}
\end{figure*}



\subsection{\minew{Offline Construction and Context Conditioning}}
\label{sec:tt_offline}

\minew{To make the offline construction reusable, we consider a family of optimization problems that share the same decision variables but differ in their initial states or target parameters. Examples include pushing an object from different initial poses and reaching different end-effector target poses. We express this family as $\max_{\bm{x}\in\Omega_x} J(\bm{x};\bm{z})$, where the context variable $\bm{z}\in\Omega_z$ specifies a problem instance and remains fixed while $\bm{x}$ is optimized. For a single instance, $\bm{z}$ can be omitted.}

\minew{We first construct an augmented TT model $\bm{\mc{Q}}^+$ offline by applying TT-Cross to the objective over the joint domain $\Omega_z\times\Omega_x$. A single set of TT cores thus approximates the objective across the prescribed family of instances, avoiding a separate TT-Cross construction for each instance. The maximum-volume principle of TT-Cross guides the selection of informative objective evaluations, producing a global surrogate that can guide the subsequent search from its first iteration.}

\subsection{\minew{Online Tree Search and Refinement}}
\label{sec:tt_online}

\minew{For online search, the given context $\bm{z}$ specifies the instance to be solved. Partial evaluation of the augmented model yields its value model, $\bm{\mc{Q}}=\bm{\mc{Q}}^+(\bm{z})$. Specifically, we evaluate the context-related cores at the given context, using linear interpolation between neighboring grid points when needed, and contract them into the remaining cores. The resulting TT represents the objective values over the decision variables, without rerunning TT-Cross or making additional objective evaluations.}

\minew{The online search uses this value model $\bm{\mc{Q}}$ together with a visit-count model $\bm{\mc{V}}$ to compute the UCB scores in Eq.~\eqref{eq:ucb}. We denote the corresponding full value and visit-count tensors by $\widehat{\bm{\mc{Q}}}$ and $\widehat{\bm{\mc{V}}}$. Their TT representations avoid explicitly storing tensors whose size grows exponentially with tree depth. The visit-count model is initialized to zero for each instance and updated as branches are explored, while the context $\bm{z}$ remains fixed.}


Due to dependence on the parent nodes, the value and visitation statistics of a node are path-dependent, introducing non-Markovian behavior. Querying the value of node $n_{i_{[j]}}$ requires summing over all completions of the branch from layer $j+1$ to depth $d$. Given the full tree approximated in TT format, the value of a node at level \( j \), denoted as \( q_{i_{[j]}} \), can be efficiently computed via tensor contractions as
\begin{equation}
\begin{aligned}
    q_{i_{[j]}} &= \sum_{i_{j+1}=0}^{N_{j+1}} \cdots \sum_{i_d=0}^{N_d} \widehat{\bm{\mc{Q}}}_{(i_1, \ldots, i_j, i_{j+1}, \ldots, i_d)} \\
    &\approx \bm{\mc{Q}}^1_{:,i_1,:} \cdots \bm{\mc{Q}}^j_{:,i_j,:} \sum_{i_{j+1}=0}^{N_{j+1}} \cdots \sum_{i_d=0}^{N_d}  \bm{\mc{Q}}^{j+1}_{:,i_{j+1},:} \cdots \bm{\mc{Q}}^d_{:,i_d,:},
\end{aligned}
\label{eq:w_compute}
\end{equation}
where each TT core \( \bm{\mc{Q}}^l \) encodes the factorized structure of the tree at layer \( l \). To compute the node value at depth \( j \), we extract the corresponding slices from the first \( j \) TT cores (representing the visited layers), and perform summation over indices in the remaining cores from \( j+1 \) to \( d \). The final node value is obtained by multiplying the resulting matrices across all layers. Figure \ref{fig:tree_node} illustrates how to compute the value of a node in the first layer of a three-layer tree using TT cores. The multiple arrows between TT cores indicate the ability to perform parallel selection in the TT-Tree. 

Similarly, the visit count $v_{i_{[j]}}$ is computed as:
\begin{equation}
\begin{aligned}
    v_{i_{[j]}} &= \sum_{i_{j+1}=0}^{N_{j+1}} \cdots \sum_{i_d=0}^{N_d} \widehat{\bm{\mc{V}}}_{(i_1, \ldots, i_j, i_{j+1}, \ldots, i_d)} \\
    &\approx \bm{\mc{V}}^1_{:,i_1,:} \cdots \bm{\mc{V}}^j_{:,i_j,:} \sum_{i_{j+1}=0}^{N_{j+1}} \cdots \sum_{i_d=0}^{N_d} \bm{\mc{V}}^{j+1}_{:,i_{j+1},:}  \cdots \bm{\mc{V}}^d_{:,i_d,:}.
\end{aligned}
\label{eq:v_compute}
\end{equation}

With $q_{i_{[j]}}$ and $v_{i_{[j]}}$ efficiently available via TT contractions (Eqs.~\eqref{eq:w_compute}--\eqref{eq:v_compute}), the online phase follows the standard MCTS pipeline: selection, expansion, simulation, and backpropagation, with an additional top-$\tau$ solutions maintenance step. We now describe each step in detail.

\paragraph{Selection and Expansion.}
Given $q_{i_{[j]}}$, $v_{i_{[j]}}$, and $v_{i_{[j-1]}}$, node expansion follows the UCB rule from \eqref{eq:ucb}. The TT-based representation enables parallel selection, which is typically non-trivial in traditional node-based or table-based implementations \citep{chaslot2008parallel}. If no further expansion is possible, the current branch terminates. Otherwise, the branch is extended by one layer and followed by a simulation to the leaf.

\paragraph{Simulation.}
Rather than performing a random rollout, we leverage the TT value model $\bm{\mc{Q}}$ as a global approximation of the decision tree to guide the simulation strategy via stochastic sampling, treating TT values as unnormalized probabilities. This enables parallel simulation and yields a top-$\tau$ set of candidate solutions.

\paragraph{Backpropagation.}
Following the simulation step, the TT models associated with the traversed branches are updated through backpropagation. Specifically, the visit count tensor $\bm{\mc{V}}$ is updated incrementally via a residual strategy. \minew{Let $\Delta \widehat{\bm{V}}_{\gamma}^{\ell}$ denote the residual visit count tensor that tracks node visitations at layer $\gamma$ during iteration $\ell$, where an element is $1$ if the corresponding node is visited and $0$ otherwise.} \minew{Since at most $\tau$ distinct nodes are visited at layer $\gamma$ in each iteration, $\Delta \widehat{\bm{V}}_{\gamma}^{\ell}$ is a sum of at most $\tau$ one-hot (i.e., rank-one) tensors and therefore admits an exact TT representation of rank at most $\tau$.} Leveraging the fact that the newly visited indices are precisely known a priori, we tailor the standard TT-Cross algorithm into an efficient, guided variant. \minew{Standard TT-Cross initializes its left and right index sets randomly and refines them over several sweeps by the maxvol principle. The guided variant instead initializes them directly with the prefixes $(i_1,\ldots,i_k)$ and suffixes $(i_{k+1},\ldots,i_\gamma)$ of the visited index tuples $i_{[\gamma]}$. Because these index sets contain all nonzero entries of $\Delta \widehat{\bm{V}}_{\gamma}^{\ell}$, the cross interpolation is exact, and the guided variant recovers its TT representation $\Delta \bm{\mc{V}}_{\gamma}^{\ell}$ in a single iteration. It only evaluates $\Delta \widehat{\bm{V}}_{\gamma}^{\ell}$ on the fibers through the visited nodes, which is cheap and requires no evaluation of $J$. The residual is then added to the visit count model, $\bm{\mc{V}}_{\gamma}^{\ell} = \bm{\mc{V}}_{\gamma}^{\ell-1} + \Delta \bm{\mc{V}}_{\gamma}^{\ell}$, by TT addition (i.e., block-wise concatenation of the cores). This update is also exact, so the visit counts grow exactly as in standard MCTS. Since the ranks of the TT sum add up over iterations, the visit count model can optionally be compressed by TT truncation when storage is limited; with a small relative tolerance (e.g., $10^{-5}$), the truncation leaves the visit counts practically unchanged. The visit counts of the ancestor layers $1,\ldots,\gamma-1$ are updated in the same way, with residuals obtained by summing $\Delta \bm{\mc{V}}_{\gamma}^{\ell}$ over its last indices, i.e., contracting its last cores with all-ones vectors, so that each visit is propagated to all nodes along the traversed path.} The value model $\bm{\mc{Q}}$ can also be updated in a similar manner to enhance approximation accuracy, but we omit this step in practice to reduce runtime. 

\paragraph{Top-$\tau$ Update and Refinement.}
We maintain a candidate set $\mathcal{S}$ to dynamically track the top-$\tau$ elite solutions, which is updated iteratively with newly discovered solutions. For continuous decision variables, the coarse solutions obtained from TT-Tree Search are subsequently optimized using CMA-ES as a local refiner.

\noindent\textbf{Remark (Parallelism in TTTS).} Standard MCTS faces two fundamental challenges when parallelizing tree search: (1) all threads tend to follow the same UCB-greedy path from the root, causing redundant exploration; and (2) concurrent updates to shared per-node statistics require expensive locking mechanisms \citep{chaslot2008parallel}. TTTS addresses both challenges through the separable structure of its TT representation, which decouples the tree across layers and enables global, layer-wise operations.
For challenge (1): in standard MCTS, selection walks down the tree one node at a time along a single root-to-leaf path, so parallel threads all tend to visit the same nodes. In TTTS, the separable structure allows UCB scores for \emph{all} nodes at a given layer to be computed simultaneously via TT contractions (Eq.~\eqref{eq:w_compute}--\eqref{eq:v_compute}), and the top-$\tau$ nodes are selected in one shot. This layer-wise parallel selection naturally diversifies exploration across $\tau$ branches at once.
For challenge (2): in standard MCTS, both visit counts and action-value estimates are stored as per-node counters, so threads updating these counters simultaneously must contend for locks. In TTTS, both statistics are encoded as TT models ($\bm{\mc{V}}$ for visit counts and $\bm{\mc{Q}}$ for node values) and updated via TT-Cross in a single bulk operation, eliminating the need for per-node synchronization. The simulation phase likewise draws multiple complete trajectories in parallel by treating $\bm{\mc{Q}}$ as an unnormalized probability model, yielding $\tau$ candidate solutions per iteration without thread contention.

\begin{algorithm}[htbp]
\scalebox{0.9}{
\caption{Tensor Train Tree Search (TTTS)}
\label{alg:TTTS}
\begin{minipage}{1.25\textwidth} 
\begin{algorithmic}[1]
\Function{TensorTrainTreeSearch}{$\bm{x}_0$, $\bm{z}$, $J$, $\Omega_x$, $L$,  $\mathcal{I}$}
\State {\textbf{Input: }{Initial state $\bm{x}_0$, Condition var. $\bm{z}$$\,\in\Omega_z$,  \\
\quad \quad Maximum iter. $L$, Objective function $J$, \\
\quad \quad Search domain $\Omega_x = \{ \bm{x}^{(i_1,\ldots,i_d)} : i_j \in \{1,\ldots,N_j\} \}$, \\
\quad \quad Index set $\mathcal{I} \subseteq \{1,\ldots,N_1\} \times \cdots \times \{1,\ldots,N_d\}$, }}
\State {\textbf{Hyperparameters:} {Number of solutions $\tau$,  \\
\quad \quad Exploration param. $c$} \textcolor{gray}{\# default: $\tau=10$, $c=3$}}
\State{\textbf{Output: }{ top-$\tau$ solutions: solution set $\mathcal{S}$}}

\State {\color{gray}\textbf{// == \minew{Phase I: TT-Tree Initialization (offline)} ==}}
\State $\bm{\mc{Q}}^+ \leftarrow \text{TT-Cross}(J, \Omega_z \times \Omega_x)$ \textcolor{gray}{\# once per task}

\State {\color{gray}\textbf{// == \minew{Phase II: TT-Tree Search (online)} ==}}
\State $\bm{\mc{Q}} \leftarrow \bm{\mc{Q}}^+(\bm{z})$ \textcolor{gray}{\# per problem instance $\bm{z}$}
\State $i_0 \gets \text{Node}(\bm{x}_0)$, $\bm{\mc{V}_0} \leftarrow \bm{0}$,  $\mathcal{S} = [\;]$
\For{$\ell = 1, 2, \ldots$, L}
    \For{$j = 1, \ldots, d$}
        \State $q_{i_{[j]}} \leftarrow $ Eq. \eqref{eq:w_compute}, $v_{i_{[j]}}, v_{i_{[j-1]}} \leftarrow $ Eq. \eqref{eq:v_compute}
        \State $i_{[j]} \gets \arg\max\nolimits^{\tau}_{i_{[j]} \in \mc{I}_{[j]}} \left( \frac{q_{i_{[j]}}}{v_{i_{[j]}}} + c \cdot \sqrt{\frac{\log v_{i_{[j-1]}}}{v_{i_{[j]}}}} \right)$
        \State \textbf{if} $i_j$ is not expanded \textbf{then} \textbf{break}
    \EndFor

    \For{$s = j+1, \ldots, d$}
        \State $q_{i_{[s]}} \gets $ Eq. \eqref{eq:w_compute}
        \State $i_{[s]} \gets  \arg\max_{i \in \mc{I}_{[s]}}^{\tau}  q_{i_{[s]}}$
    \EndFor

    \State  $\bm{x}^\star \gets \arg\max\nolimits^{\tau}_{\bm{x} \in \Omega_x(i_{[d]})} J(\bm{x})$

    \For{$\gamma = 1, \ldots, j$}
        \State $\bm{\mc{V}}_{\gamma}^{\ell} = \bm{\mc{V}}_{\gamma}^{{\ell}-1} + \text{Guided TT-Cross}(\Delta \widehat{\bm{V}}_\gamma^{\ell}, i_{[\gamma]})$
    \EndFor

    \State  $\mathcal{S}^+ \gets \text{append}(\mathcal{S}, \bm{x}^\star)$
    \State  $\mathcal{S} \gets \arg\max\nolimits^{\tau}_{\bm{x} \in \mathcal{S}^+} J(\bm{x})$
\EndFor
\State {\color{gray}\textbf{// == \minew{Phase III: TT-Tree Refinement (online)} ==}}
\State  $\mathcal{S} \gets \text{Refine}(\mathcal{S})$ \textcolor{gray}{\# e.g., CMA-ES on continuous dims}
\State \Return  $\mathcal{S}$
\EndFunction
\end{algorithmic}
\end{minipage}
}
\end{algorithm}

\subsection{Algorithm Walkthrough via an Illustrative Example}
\label{sec:walkthrough}

\begin{figure}
    \centering
    \includegraphics[width=0.8\linewidth]{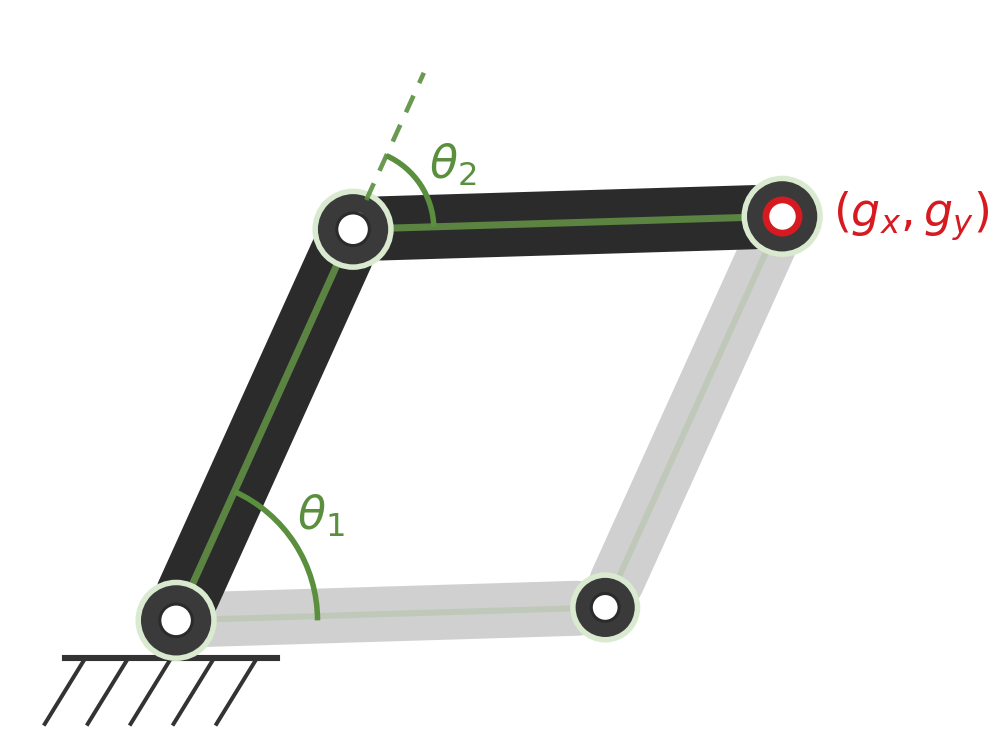}
    \caption{\textbf{Two-joint inverse kinematics with multimodal solutions.} 
    Two distinct arm configurations, shown in dark and light gray, reach the same end-effector goal $(g_x, g_y)$ due to the nonlinear kinematics, illustrating the multimodal and nonconvex nature of the problem.
    }
    \label{fig:ik_setup}
\end{figure}

\begin{figure*}
    \centering
    \begin{subfigure}{0.45\textwidth}
        \centering
        \includegraphics[width=\linewidth]{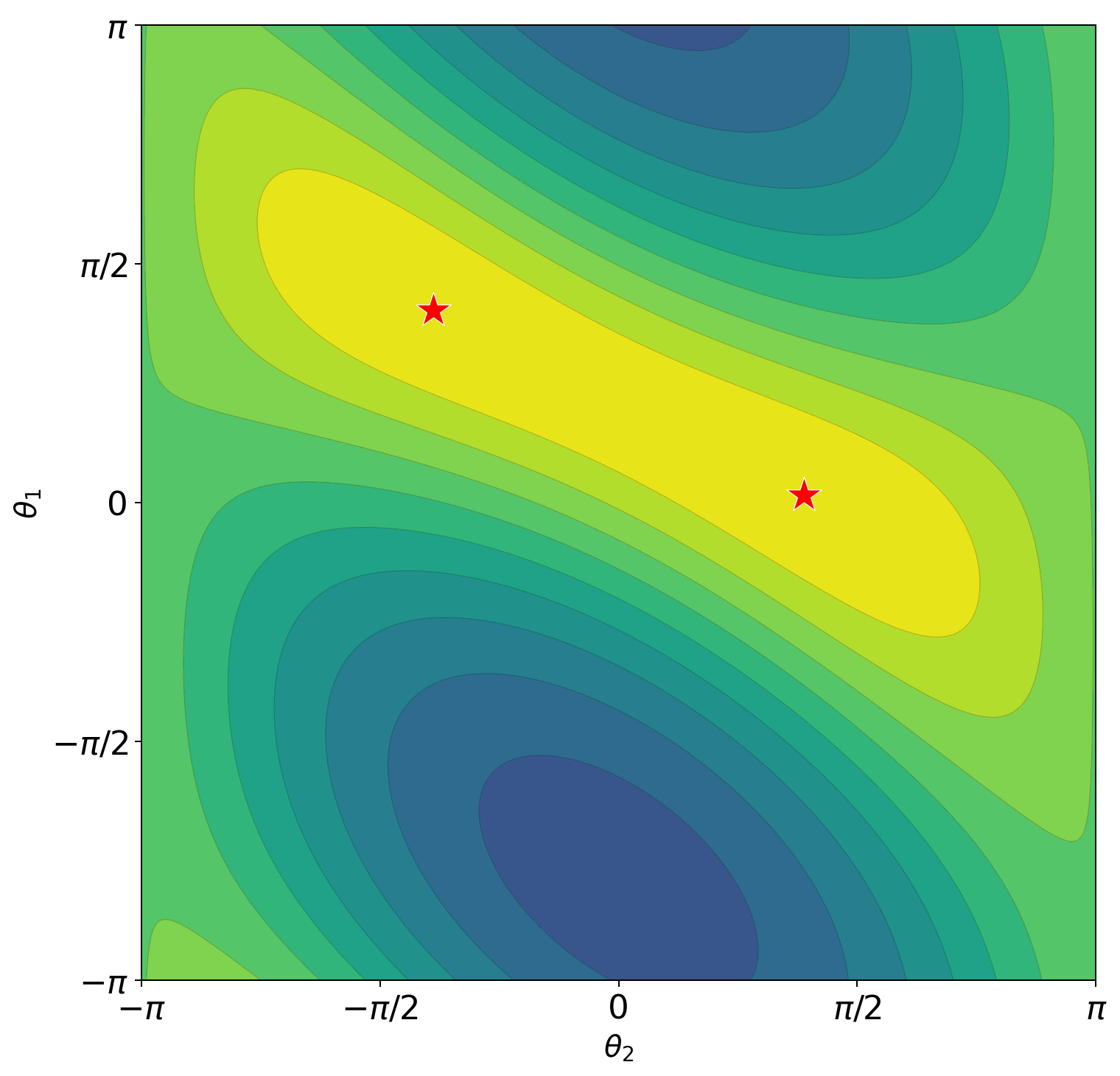}
        \caption{Top-down view of $J(\theta_1, \theta_2)$ over the range $[-\pi, \pi]$ for each joint. Lighter colors (yellow) indicate higher objective values.}
        \label{fig:ik_value}
    \end{subfigure}
    \hfill
    \begin{subfigure}{0.5\textwidth}
        \centering
        \includegraphics[width=\linewidth]{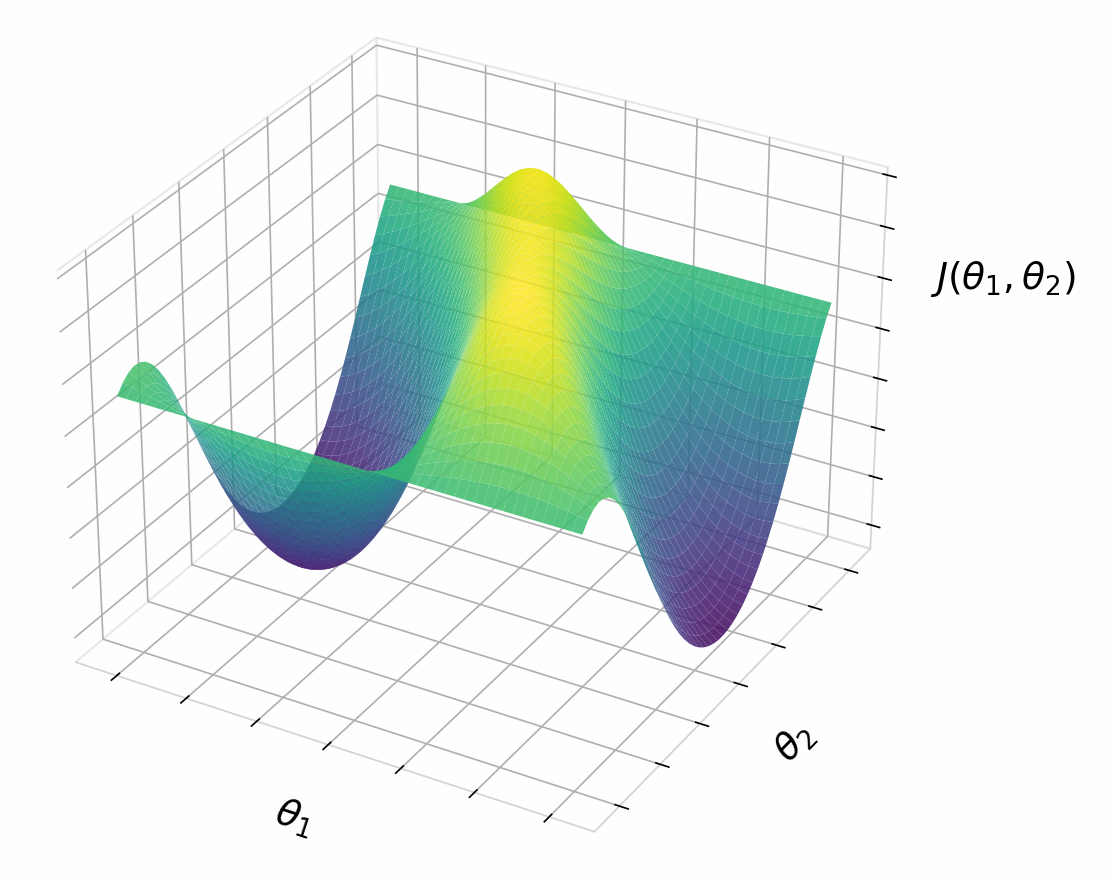}
        \caption{3D surface of $J(\theta_1, \theta_2)$, revealing its nonconvex, multimodal structure with two distinct peaks.}
        \label{fig:ik_value_3d}
    \end{subfigure}

    \vspace{0.5em}

    \begin{subfigure}{0.45\textwidth}
        \centering
        \includegraphics[width=\linewidth]{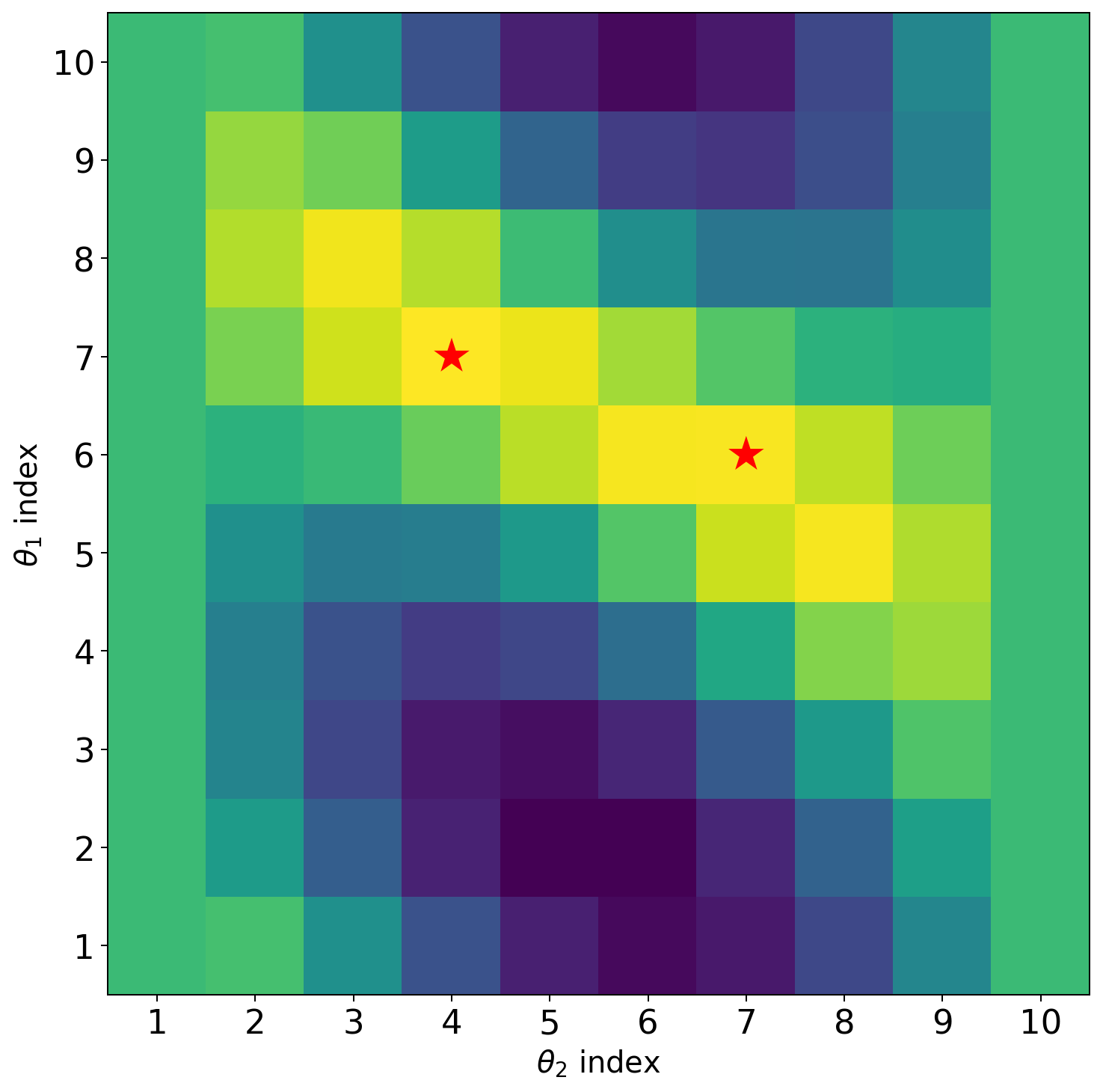}
        \caption{Discretized $10\!\times\!10$ value matrix. Red-star markers indicate the two optimal configurations.}
        \label{fig:ik_matrix}
    \end{subfigure}
    \hfill
    \begin{subfigure}{0.5\textwidth}
        \centering
        \includegraphics[width=\linewidth]{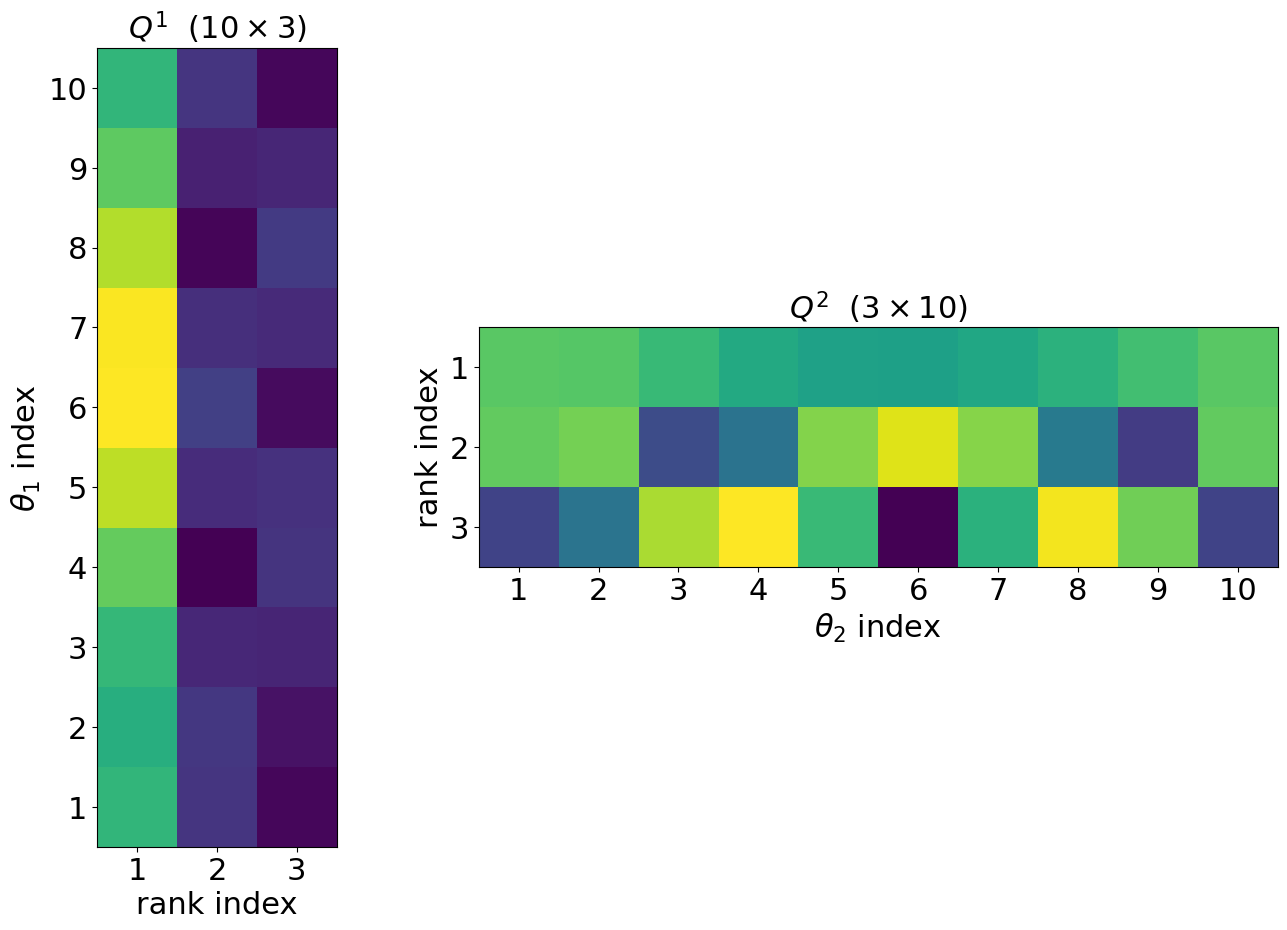}
        \caption{TT factorization of the value matrix into $\bm{Q}^1\!\in\!\mathbb{R}^{10\times 3}$ and $\bm{Q}^2\!\in\!\mathbb{R}^{3\times 10}$ at rank $r=3$. TT factorization corresponds to matrix factorization in this low-dimensional illustrative example.}
        \label{fig:matrix_decomp}
    \end{subfigure}
    \caption{\textbf{Objective function, discretization, and TT approximation for the 2-DOF IK example.} \textbf{(a)} Top-down heatmap of the value function $J(\theta_1,\theta_2)$ over $[-\pi,\pi]^2$. \textbf{(b)} 3D surface view of $J$, clearly showing its nonconvex, multimodal landscape with two distinct peaks, each corresponding to one kinematic solution. \textbf{(c)} Discretizing onto a $10\!\times\!10$ grid yields a matrix representation; red-star markers indicate the two optimal configurations. \textbf{(d)} TT-Cross decomposes the value matrix into two low-rank factors $\bm{Q}^1\!\in\!\mathbb{R}^{10\times3}$ and $\bm{Q}^2\!\in\!\mathbb{R}^{3\times10}$ at rank $r=3$, capturing the dominant structure with significantly fewer parameters.}
    \label{fig:ik_value_matrix_decomp}
\end{figure*}

\begin{figure*}
    \centering
    \includegraphics[width=1.0\linewidth]{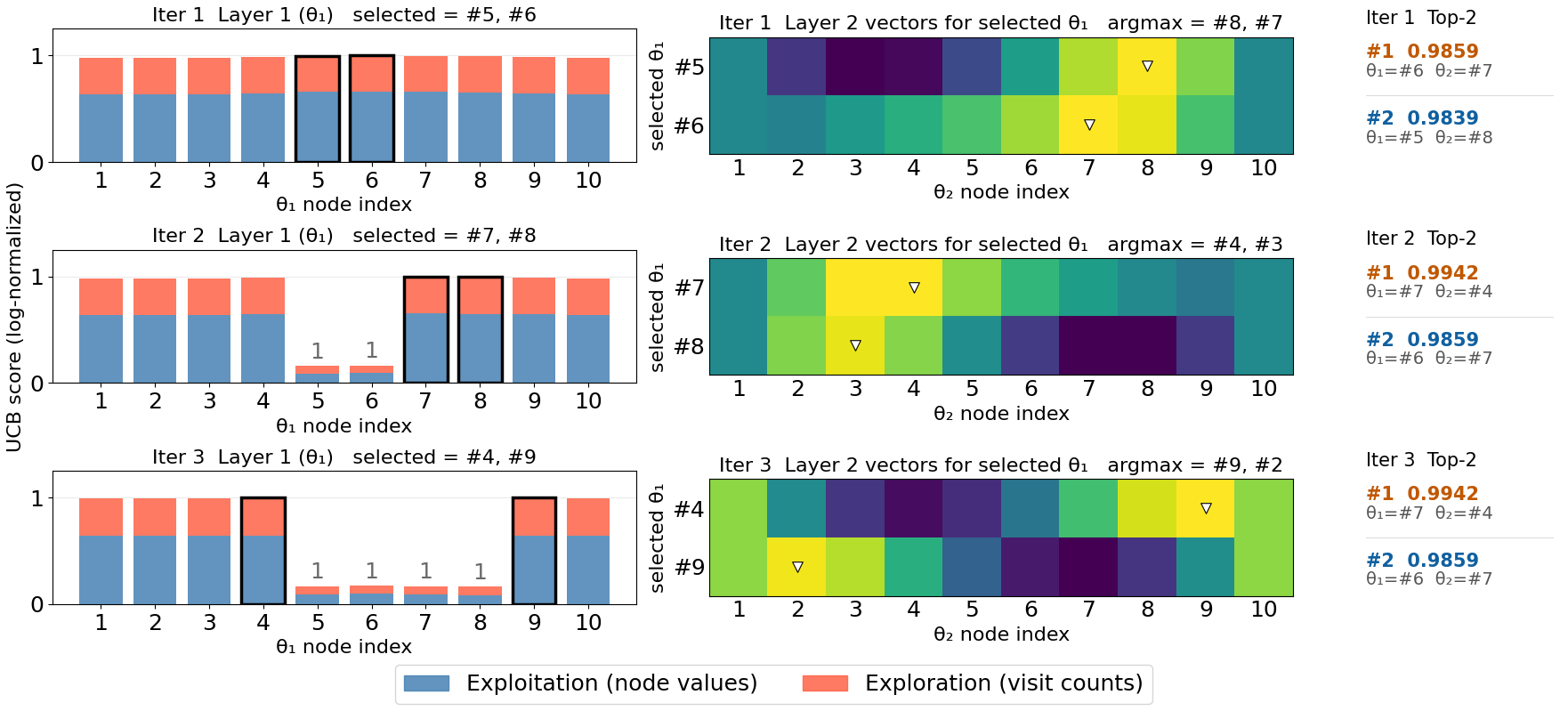}
    \caption{\textbf{TTTS applied to the 2-DOF IK problem across three iterations.} \textbf{Left:} UCB bar charts for Layer-1 ($\theta_1$) node selection, decomposed into exploitation (blue) and exploration (red) components; nodes with black borders are selected, and gray labels denote accumulated visit counts. \textbf{Middle:} Layer-2 ($\theta_2$) value heatmaps conditioned on each selected first-layer node; triangular markers indicate the top-$\tau$ selected second-layer nodes. \textbf{Right:} maintained top-$\tau$ candidate solutions after each iteration. As visit counts accumulate across iterations, the exploration bonus progressively shifts selection toward unvisited regions, enabling systematic coverage of the full solution space.}
    \label{fig:ik_ttts}
\end{figure*}

We walk through Algorithm~\ref{alg:TTTS} step-by-step using a 2-DOF inverse kinematics (IK) problem. As shown in Fig.~\ref{fig:ik_setup}, the task is to find joint angles $(\theta_1, \theta_2) \in [-\pi, \pi]^2$ such that the robot end-effector position, denoted by $(r_x, r_y)$, reaches the target position $(g_x, g_y)$. Because of the nonlinear kinematics, the same goal position can be reached by two distinct arm configurations, shown in dark and light gray in Fig.~\ref{fig:ik_setup}. This yields a representative nonlinear problem with a multimodal solution set, well suited for illustrating the key properties of TTTS.

\paragraph{Objective function and discretization.} We define the cost function as $c(\theta_1, \theta_2) = (r_x - g_x)^2 + (r_y - g_y)^2$, which represents the squared end-effector positioning error, and convert it into a maximization objective:
\begin{equation}
    J(\theta_1, \theta_2) = e^{-c(\theta_1, \theta_2)}.
\end{equation}
Figs.~\ref{fig:ik_value} and~\ref{fig:ik_value_3d} show the landscape of $J$ from top-down and perspective views, respectively. The 3D surface in Fig.~\ref{fig:ik_value_3d} clearly reveals the nonconvex, multimodal structure of $J$, with two distinct peaks corresponding to the two kinematic solutions, a landscape in which gradient-based methods are prone to converging to a single optimum. Discretizing $(\theta_1, \theta_2)$ onto a $10 \times 10$ grid yields the objective matrix in Fig.~\ref{fig:ik_matrix}, where the two red-star entries mark the optimal grid configurations.

\paragraph{Initialization.}
We start TTTS by calling TT-Cross to construct a compact TT approximation of the objective function, with the maximum rank budget set to 3. This produces two low-rank factor matrices $\bm{Q}^1 \!\in\! \mathbb{R}^{10 \times 3}$ and $\bm{Q}^2 \!\in\! \mathbb{R}^{3 \times 10}$ with TT rank $r = 3$, as illustrated in Fig.~\ref{fig:matrix_decomp}. Despite using far fewer parameters than the full $10 \times 10$ matrix, this factorization captures the dominant structure of the value landscape and concentrates probability mass near high-value regions, providing an informative global prior before any tree traversal begins.

\paragraph{Selection.}
At each iteration $\ell$, TTTS selects the top-$\tau$ nodes in Layer~1 ($\theta_1$) by maximizing the UCB score. The node value for the first layer is computed from the TT cores as:
\begin{equation}
    q_{i_{[1]}} = \bm{Q}^1_{i_1, :} \sum_{i_2=1}^{N_2} \bm{Q}^2_{:, i_2},
\end{equation}
which marginalizes over all second-layer completions via $\bm{Q}^2$, serving as the exploitation term. The visit count $v_{i_1}$ provides the exploration bonus. Given the selected first-layer node $i_1^*$, the second-layer node values are:
\begin{equation}
    q_{i_{[2]}} = \bm{Q}^1_{i_1^*, :}\, \bm{Q}^2_{:, i_2},
\end{equation}
and the top-$\tau$ nodes in $\theta_2$ are selected accordingly.

Fig.~\ref{fig:ik_ttts} traces this process across three iterations (left: Layer-1 UCB bar charts; middle: Layer-2 value heatmaps; right: maintained top-$\tau$ solutions). In Iteration~1, all visit counts are zero, so the UCB score is dominated entirely by the exploitation term (blue bars). Nodes $i_1 = 5$ and $i_1 = 6$ achieve the highest $q_{i_1}$ values and are selected. Conditioned on these selections, the Layer-2 heatmaps identify $i_2 = 8$ for $i_1 = 5$ and $i_2 = 7$ for $i_1 = 6$. The TT model successfully guides search toward high-value regions from the very first iteration. However, due to the low-rank approximation, not all optima may be captured precisely in a single pass. The top-$\tau$ solution list is maintained and updated across iterations as the UCB exploration bonus progressively steers search toward other promising regions.

\paragraph{Backpropagation.}
After each iteration, the visit count model $\bm{\mc{V}}$ is updated via Guided TT-Cross to record the explored nodes. In Iteration~2, nodes $i_1 = 5$ and $i_1 = 6$ carry visit count~1 (indicated by the ``1'' labels in Fig.~\ref{fig:ik_ttts}), which reduces their exploration bonus and raises the UCB scores of their neighbors $i_1 = 7$ and $i_1 = 8$. By Iteration~3, nodes $i_1 = 4$ and $i_1 = 9$ are selected, revealing a systematic outward expansion: TTTS uses the exploitation term to focus on high-value regions and the exploration term to spread coverage as nodes are visited, together driving strategic search across the full solution space.

\paragraph{Refinement.}
The top-$\tau$ discrete candidate solutions from TT-Tree search are passed to CMA-ES for continuous refinement. This corrects for grid discretization errors and yields the two precise IK configurations shown in Fig.~\ref{fig:ik_setup}.

This IK example could of course be solved differently; it is used here to didactically illustrate how TTTS can combine a compact TT representation as a global prior with the UCB-driven exploration--exploitation mechanism of MCTS, simultaneously discovering all optimal modes through linear-complexity tensor operations.
\subsection{Theoretical Analysis}
We establish three theoretical results that together show TTTS achieves MCTS-like optimality with greater efficiency. Proposition~1 shows that TT structure enables exact MCTS node value computation in $\mathcal{O}(Ndr^2)$, an exponential improvement over the $\mathcal{O}(N^d)$ complexity of explicit tree evaluation. Proposition~2 establishes that the TT model provides a near-optimal surrogate: its best solution deviates from the true global optimum by at most $2\epsilon$, where $\epsilon$ is the TT approximation error, which can be obtained while performing TT-Cross. Proposition~3 establishes that TTTS achieves the same asymptotic completeness as MCTS: given sufficient iterations and storage, it asymptotically approaches the true global optimum on the decision grid. Overall, TTTS enables efficient MCTS while providing asymptotic global convergence.



\begin{theorem} Given a TT model $\bm{\mc{T}}$, the exact MCTS node values at every layer can be computed in $\mathcal{O}(Ndr^2)$, and the greedy-optimal solution on $\bm{\mc{T}}$ can be retrieved with the same complexity.
\label{thm:TT-global-opt}
\end{theorem}

\begin{proof}
As shown in \eqref{eq:tt_rep}, each element of $\bm{\mc{T}}$ is a product of TT cores. The MCTS node value at layer $j$, defined as the sum of $\bm{\mc{T}}$ over all completions of the current path, is given by \eqref{eq:w_compute}. Computing $q_{i_{[j]}}$ requires contracting the $d-j$ right-side cores, which costs $\mathcal{O}(r^2)$ per contraction; the right-side contraction can be precomputed once and reused for all $N_j$ siblings, giving $\mathcal{O}(N_j r^2)$ per layer and $\mathcal{O}(Ndr^2)$ for all layers.

The greedy-optimal solution is then obtained by traversing the tree layer by layer, at each step selecting $x_k^* = \arg\max_{x_k} q_{i_{[k]}}$, which adds only $\mathcal{O}(Nd)$ overhead. This is equivalent to running MCTS on $\bm{\mc{T}}$ to full convergence: since the node values are computed exactly from the TT model rather than estimated via rollouts, a single greedy pass recovers the solution that standard MCTS would identify in the limit of infinite simulations.
\end{proof}

\begin{theorem}
Let $\widehat{\bm{\mc{T}}}$ be the true objective landscape \minew{(i.e., the objective function $J(\bm{x})$ on the grid $\Omega_x$)} and $\bm{\mc{T}}$ be its Tensor Train (TT) approximation \minew{(i.e., the TT value model $\bm{\mc{Q}}$)} with $\|\bm{\mc{T}} - \widehat{\bm{\mc{T}}}\|_\infty \leq \epsilon$. Define the respective optimal solutions as:
\begin{equation*}
    \bm{x}^\dagger = \arg\max_{\bm{x}} \bm{\mc{T}}(\bm{x}), \quad \text{and} \quad \bm{x}^* = \arg\max_{\bm{x}} \widehat{\bm{\mc{T}}}(\bm{x}).
\end{equation*}
Then, the true objective value evaluated at the optimum of the TT approximation deviates from the true global optimum by at most $2\epsilon$, i.e.,
\begin{equation}
    \widehat{\bm{\mc{T}}}(\bm{x}^*) - \widehat{\bm{\mc{T}}}(\bm{x}^\dagger) \leq 2\epsilon.
\end{equation}
\end{theorem}

\begin{proof}
Let $\bm{x}^* = \arg\max_{\bm{x}} \widehat{\bm{\mc{T}}}(\bm{x})$. By algebraic decomposition:
\begin{align*}
    \widehat{\bm{\mc{T}}}(\bm{x}^*) - \widehat{\bm{\mc{T}}}(\bm{x}^\dagger)
     &\leq \bigl|\widehat{\bm{\mc{T}}}(\bm{x}^*) - \bm{\mc{T}}(\bm{x}^*)\bigr|
     + \bigl[\bm{\mc{T}}(\bm{x}^*) - \bm{\mc{T}}(\bm{x}^\dagger)\bigr] \\
     &+ \bigl|\bm{\mc{T}}(\bm{x}^\dagger) - \widehat{\bm{\mc{T}}}(\bm{x}^\dagger)\bigr|. \notag
\end{align*}
The first and third terms are each bounded by $\epsilon$ via the $L_\infty$ bound. The middle term is non-positive since $\bm{x}^\dagger = \arg\max \bm{\mc{T}}$, so $\bm{\mc{T}}(\bm{x}^*) - \bm{\mc{T}}(\bm{x}^\dagger) \leq 0$. Therefore $\widehat{\bm{\mc{T}}}(\bm{x}^*) - \widehat{\bm{\mc{T}}}(\bm{x}^\dagger) \leq 2\epsilon$.
\end{proof}

\minew{Proposition~2 quantifies how surrogate approximation error affects the quality of a surrogate-optimal solution. We next examine how the subsequent tree search achieves asymptotic global convergence even when the objective surrogate is approximate. The argument leading to Proposition~3 has two steps: establishing that the TT updates preserve exact visit counts, and showing that UCB selection with these counts visits every node infinitely often.
}

\begin{lemma}
\label{lem:exact_visits}
\minew{Under exact residual construction and exact TT updates, the visit-count model equals the true visit count at every node and iteration.}
\end{lemma}

\begin{proof}
\minew{The increments admit an explicit exact representation because the selected indices are known. Let $\mathcal{B}_{\gamma}^{\ell}$ be the collection of branches selected in iteration $\ell$ that terminate at layer $\gamma$. Their count increment is}
\begin{equation*}
\minew{\Delta\widehat{\bm{V}}_{\gamma}^{\ell}
=\sum_{\bm{i}\in\mathcal{B}_{\gamma}^{\ell}}
\bm{e}_{i_1}\otimes\cdots\otimes\bm{e}_{i_\gamma},}
\end{equation*}
\minew{where $\bm{e}_{i_k}$ is the corresponding coordinate vector and repeated selections, if any, are counted with multiplicity. Each summand has TT rank one, so their sum has an exact TT representation with ranks at most $|\mathcal{B}_{\gamma}^{\ell}|$. This gives a direct construction of the residual without approximating its entries.}

\minew{The guided TT-Cross introduced in Algorithm~\ref{alg:TTTS} performs exactly this construction. Leveraging the fact that the newly visited indices are precisely known a priori, it constructs its nested index sets directly from the prefixes and suffixes of the selected branches. These sets contain all rows $R$ and columns $C$ supporting the nonzero entries of each unfolding $\bm{A}$ of the residual. Hence the skeleton decomposition $\bm{A}=\bm{A}_{:,C}\bm{A}_{R,C}^{+}\bm{A}_{R,:}$ is exact, where $+$ denotes the Moore--Penrose pseudoinverse \citep{goreinov1997theory}. Using these support-covering index sets without rank truncation yields an exact TT representation of the residual \citep{oseledets2010_ttcross1}.}

\minew{The count identity now follows by induction. Initially, both the counts and their TT representations are zero. Exact TT addition inserts the residual at its terminal layer, while exact contractions propagate its increments to all ancestor layers. Each selected branch therefore adds one to the count of each node on its path and leaves all other counts unchanged. Applying these updates for all selected branches preserves the identity at iteration $\ell$.}
\end{proof}

\begin{lemma}
\label{lem:inf_visits}
\minew{Consider a finite decision tree with bounded node values and exact visit counts. Suppose that the search uses the UCB rule in Eq.~\eqref{eq:ucb} with $c>0$ and assigns infinite priority to unvisited children. Then every node of the tree is visited infinitely often as the number of search iterations tends to infinity.}
\end{lemma}

\begin{proof}
\minew{A logarithmic lower bound on arm visits is stated in \citep[Theorem~3]{kocsis2006bandit}. Consider an infinitely visited parent $s$. After finitely many expansions, each visit continues to a child, and infinite priority ensures that all children are visited at least once. Suppose some children are visited only finitely often. Since each visit to $s$ eventually continues to one of its finitely many children, at least one child is visited infinitely often. Dividing the UCB scores by $\sqrt{\log v_s}$ preserves their ordering: bounded exploitation terms vanish, while exploration terms tend to positive constants for finitely visited children and to zero for infinitely visited children. Hence a finitely visited child eventually has a higher UCB score than every infinitely visited sibling, so UCB selection must revisit it, a contradiction. Since the root is visited in every iteration, induction over tree depth proves the claim.}

\minew{This argument also covers TTTS's layer-wise top-$\tau$ selection, which retains and extends each selected parent's highest-UCB child to a complete branch.}
\end{proof}

\begin{theorem}\minew{Let $\Omega_x$ be a finite discrete decision domain and $J$ a deterministic objective. Under the exact-count and UCB selection conditions of Lemmas~\ref{lem:exact_visits} and~\ref{lem:inf_visits}, the best objective value found by TTTS converges to the global optimum as the number of search iterations tends to infinity.}
\end{theorem}
\begin{proof}
\minew{By Lemma~\ref{lem:inf_visits}, every terminal node is eventually visited and evaluated using the true objective. Since $\Omega_x$ is finite, a global maximizer is evaluated after finitely many iterations. The best objective value retained by TTTS is nondecreasing and cannot exceed the global
optimum. It therefore reaches the optimal value and remains equal to it thereafter, establishing asymptotic global convergence.}
\end{proof}

\minew{\textbf{Remark (Effect of TT truncation).} The analysis assumes that the visit-count models are updated without TT truncation. If TT truncation (e.g., TT rounding) is applied to control their ranks, it introduces approximation error in the visit counts. A small truncation tolerance can limit this error in practice, but the theoretical guarantee of Proposition~3 does not directly apply.}


\begin{corollary}
Propositions~1--3 together characterize TTTS's efficiency and asymptotic global convergence. Propositions~1 and~2 are conditioned on the TT approximation error $\epsilon$: they guarantee near-optimal initialization when the TT model accurately captures the objective landscape, with $\epsilon$ depending on how accurately the objective can be represented by a TT model with the chosen rank. Proposition~3 shows the asymptotic global convergence thanks to the UCB strategy. Together, TTTS provides: \emph{(i)} exact MCTS node value computation in $\mathcal{O}(Ndr^2)$ per iteration (Proposition~1), an exponential improvement over $\mathcal{O}(N^d)$ brute-force evaluation; \emph{(ii)} a TT-guided initialization concentrating search on $2\epsilon$-near-optimal regions from the first iteration (Proposition~2); and \emph{(iii)} asymptotic convergence to the global optimum (Proposition~3).
\end{corollary}

\minew{\textbf{Remark (conditioning on the context variable).} For a discrete context $\bm{z}$ taken from the grid of $\Omega_z$, conditioning extracts an exact slice of $\bm{\mc{Q}}^+$, so the conditioned model $\bm{\mc{Q}}=\bm{\mc{Q}}^+(\bm{z})$ has the same error bound $\epsilon$ with respect to $J(\cdot\,;\bm{z})$ as the augmented model $\bm{\mc{Q}}^+$. For a continuous context $\bm{z}$, values between grid points are handled by interpolating neighboring slices, which introduces an additional interpolation error in Proposition~2. This interpolation error is typically small when the objective varies smoothly with $\bm{z}$, and it decreases as the grid of $\Omega_z$ is refined.}

\subsection{Toy Examples on Function Optimization}
\label{sec:toy}
\subsubsection{Continuous nonconvex optimization.}
\label{sec:cont_opt}

\begin{figure*}[htbp]
    \centering
    \vspace{1em} 
    \begin{minipage}{0.35\linewidth}
        \subfloat[\centering Continuous optimization]{\includegraphics[width=0.95\linewidth]{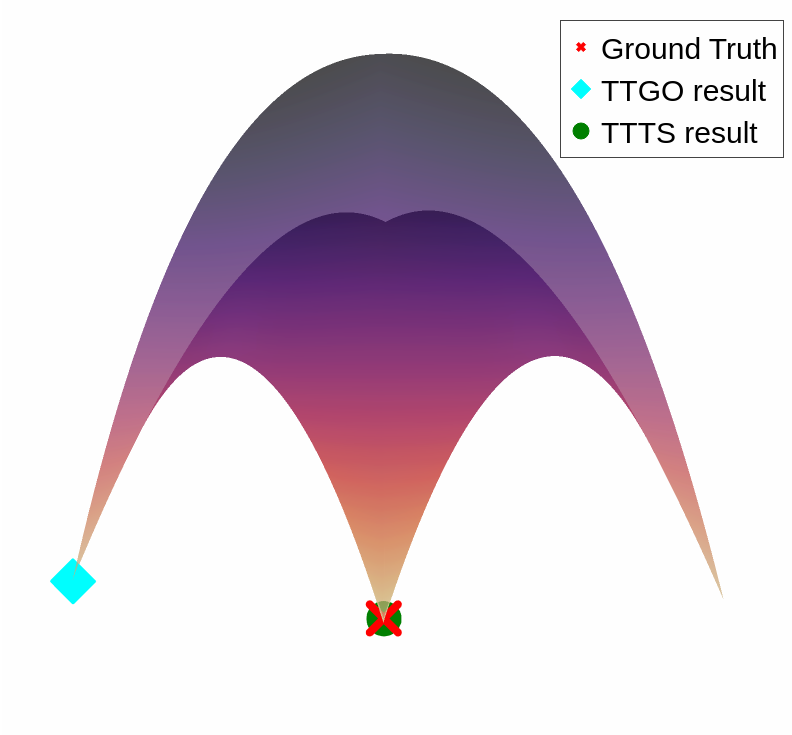}\label{fig:cont_opt}}
    \end{minipage}
    \hfill
    \begin{minipage}{0.32\linewidth}
        \subfloat[\centering Original continuous function]{\includegraphics[width=0.95\linewidth]{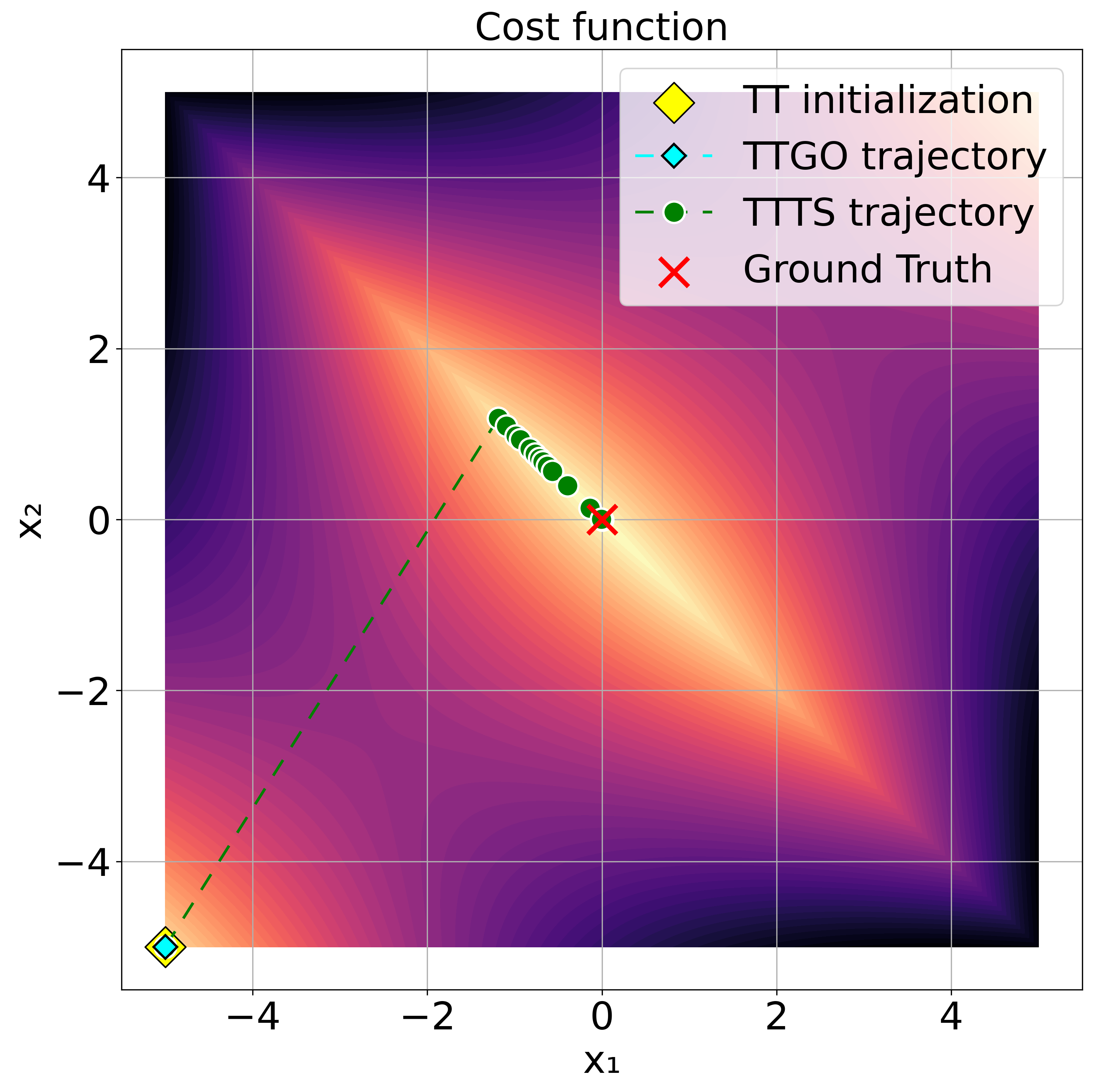}\label{fig:cont_cost_fun}}
    \end{minipage}
    \hfill
    \begin{minipage}{0.32\linewidth}
        \subfloat[\centering TT approximation]{\includegraphics[width=0.95\linewidth]{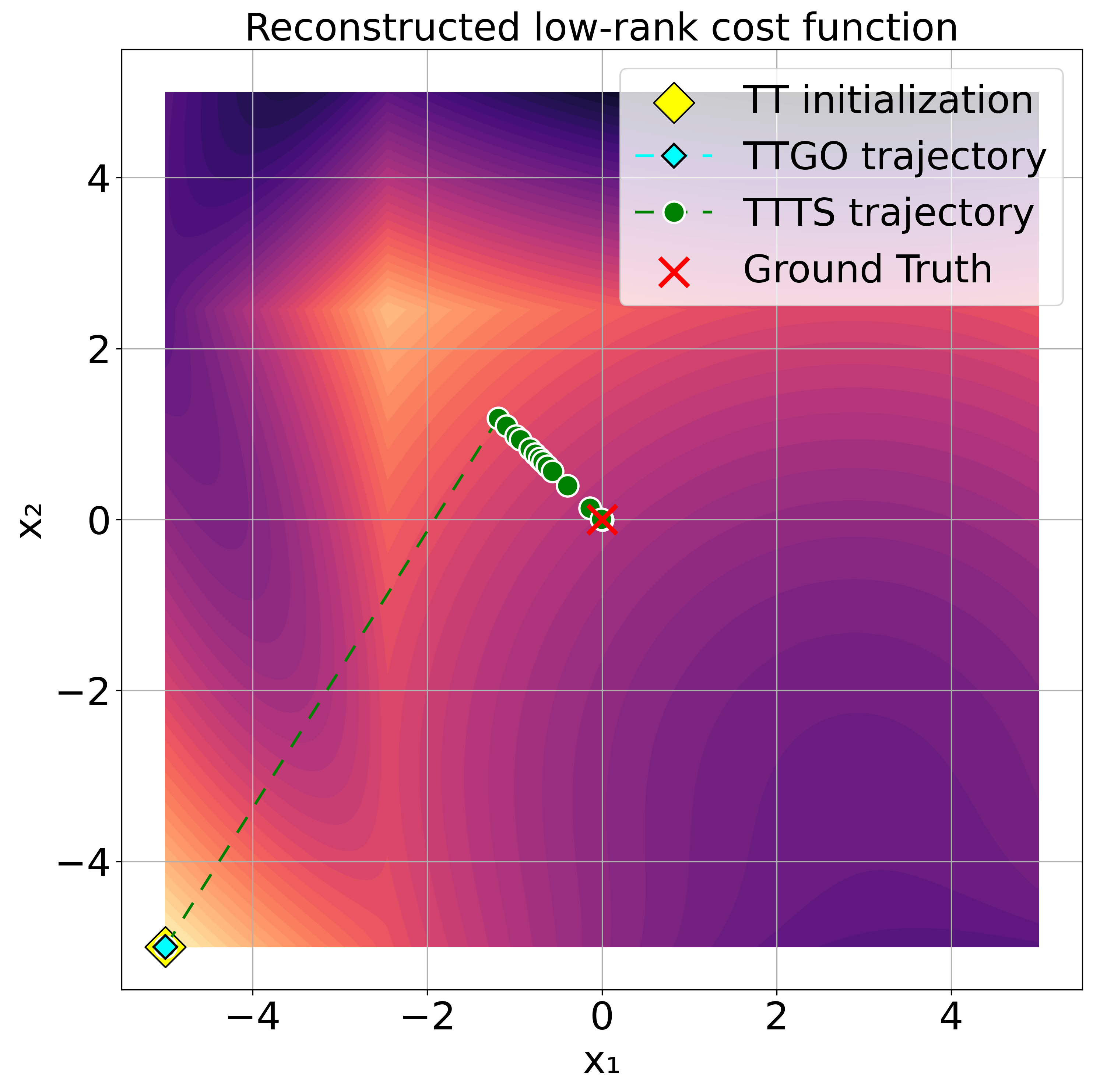}\label{fig:cont_recon_cost_fun}}
    \end{minipage}

    \begin{minipage}{0.32\linewidth}
        \subfloat[\centering Mixed-integer optimization]{\includegraphics[width=0.95\linewidth]{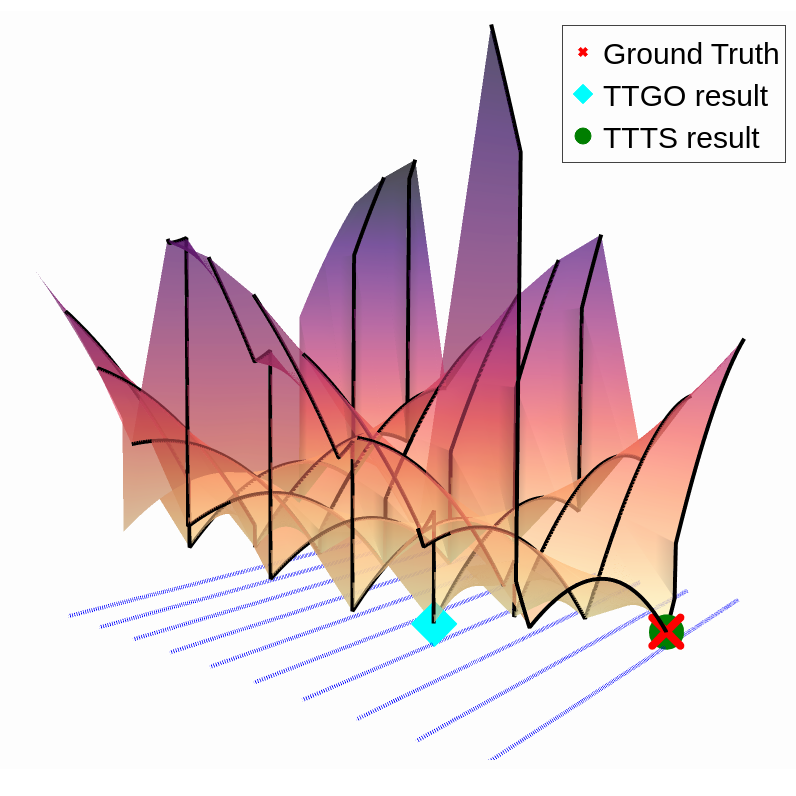}\label{fig:mix_opt}}
    \end{minipage}
    \hfill
    \begin{minipage}{0.32\linewidth}
        \subfloat[\centering Original mixed-integer function]{\includegraphics[width=0.95\linewidth]{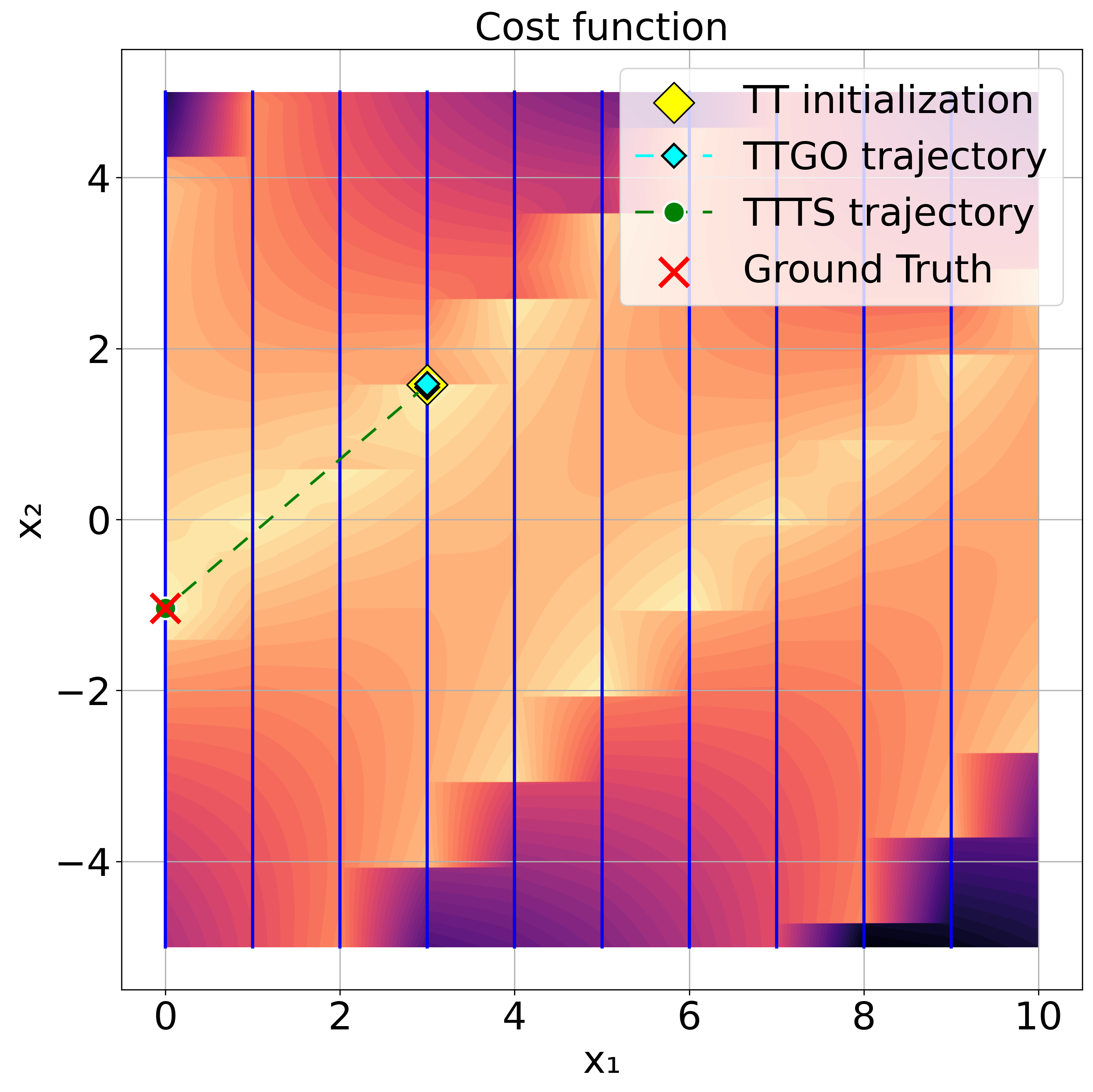}\label{fig:mix_cost_fun}}
    \end{minipage}
    \hfill
    \begin{minipage}{0.32\linewidth}
        \subfloat[\centering TT approximation]{\includegraphics[width=0.95\linewidth]{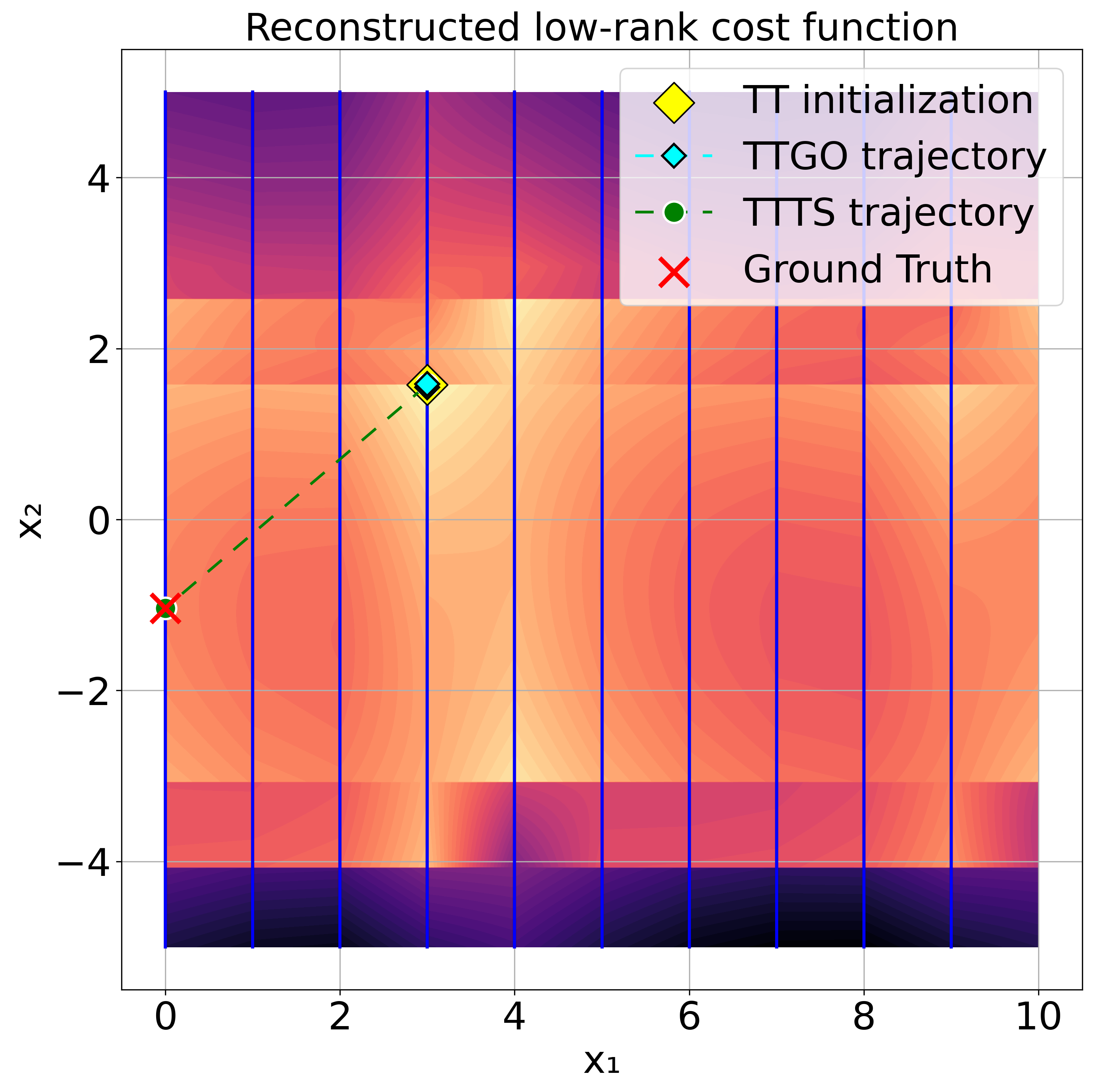}\label{fig:mix_recon_cost_fun}}
    \end{minipage}
    \caption{\textbf{Toy examples of function optimization with low-rank tensor-train (TT) approximations.}
    (a–c) Continuous nonconvex optimization: (a) landscape of a multimodal nonconvex function $f_1(x_1, x_2)$; (b) its contour map; (c) TT-Cross reconstruction with maximal TT rank $r_{\max}=2$, which captures the dominant structure but smooths out fine details.
    (d–f) Mixed-integer nonconvex optimization: (d) landscape of $f_2(x_1, x_2)$ with both discrete ($x_1$) and continuous ($x_2$) variables; (e) its grid-aligned contour plot; (f) TT approximation, which highlights promising regions but may miss the true global optimum.
    \minew{Markers show the initialization obtained from the TT model, the trajectories of TTGO and TTTS (best solution after each iteration), and the ground-truth optimum.}}
    
    \label{fig:toy_example}
\end{figure*}

Many optimization problems in robotics are nonconvex due to obstacles and nonlinear system dynamics. To illustrate both the strengths and limitations of low-rank TT approximation, we consider a full-rank nonconvex continuous function \eqref{eq: cont_func} and approximate it using a low-rank TT model with rank $r_{max} = 2$ via TT-Cross. The discrete matrix analogue of this function, shown in Figure~\ref{fig:cont_cost_fun}, clearly exhibits full-rank characteristics. Note that the general objective $J$ in \eqref{eq:general_opt} is defined as $J(\bm{x}) = e^{-c(\bm{x})}$, where $c$ is the cost function. For better visual clarity, we directly illustrate the landscape of $c$ in Figure~\ref{fig:cont_cost_fun}. The detailed definitions are provided in the Appendix.

\minew{Figure~\ref{fig:cont_recon_cost_fun} shows that the low-rank TT approximation captures useful structure in the objective landscape, allowing promising regions to be identified quickly. However, approximation errors can distort the relative quality of these regions, so optimizing the TT surrogate alone may not recover the global optimum of the original objective. TTTS addresses this limitation by combining TT-guided search with UCB-driven exploration, which allows the search to revisit regions underestimated by the surrogate and supports the asymptotic global convergence discussed above. The trajectories in Figures~\ref{fig:cont_cost_fun} and~\ref{fig:cont_recon_cost_fun}, which track the best solution found at each iteration, illustrate this mechanism. Both TTGO and TTTS start from a TT-based initialization at $(-5,-5)$, which is favored by the TT approximation but is only a local optimum of the objective function. TTGO applies gradient-based refinement and remains at this corner. TTTS, by contrast, explores beyond the region favored by the surrogate and discovers better solutions along the diagonal valley of $f_1$, eventually reaching the global optimum at $(0,0)$ (Figure~\ref{fig:cont_opt}). This example shows how the TT approximation provides initial guidance while tree search enables TTTS to overcome errors in that approximation.}

\subsubsection{Mixed-integer programming.}
\label{sec:mi_opt}

We further demonstrate the effectiveness of our approach in tackling a more challenging problem: mixed-integer nonlinear programming (MINLP). Such problems include both nonconvexity due to nonlinear constraints and the combinatorial complexity imposed by integer variables. This setting is particularly difficult because integer variables are discrete and lack gradient information, rendering standard nonlinear programming (NLP) solvers ineffective. 

Figure~\ref{fig:mix_opt} illustrates a simple MINLP example, where $x_1$ is an integer variable ranging from $0$ to $10$, and $x_2$ is a continuous variable. The cost function \eqref{eq: mix_func} is nonlinear and exhibits multiple local optima. Figure~\ref{fig:mix_cost_fun} presents the discrete matrix analogue of the continuous cost function. We approximate the function in TT format with rank $r_{\max}=2$, as shown in Figure~\ref{fig:mix_recon_cost_fun}. The results indicate that even with a low TT rank, TT-Cross can identify high-quality local optima, demonstrating the strong modeling and optimization capabilities of TT. However, the results obtained by TTGO do not correspond to the global optimum, highlighting that a low-rank TT approximation may fail to fully capture the complexity of the original function. In such cases, TTTS leverages strategic search to efficiently converge to the global optimal solution. \minew{The trajectories in Figures~\ref{fig:mix_cost_fun} and~\ref{fig:mix_recon_cost_fun} illustrate this distinction: TTGO converges to a local optimum favored by the TT approximation despite gradient-based refinement, whereas TTTS explores beyond this region through subsequent tree search and reaches the global optimum of the original objective.}

\textbf{Remark on the TT approximation.}
The accuracy of the TT approximation governs the \emph{speed} of convergence, not convergence itself, and two complementary regimes arise in practice.
When the TT model approximates the decision tree accurately, it captures the global landscape from a small number of function evaluations, and TTTS identifies near-optimal solutions from the very first iterations (Proposition~2 with small $\epsilon$).
When the approximation is less accurate, the TT model introduces nonzero $\epsilon$ but still captures the dominant features of the objective landscape, providing an informative warm-start that concentrates early MCTS exploration near promising regions.
This is illustrated in Figure~\ref{fig:toy_example}(c), where even a rank-$2$ approximation of a high-rank (rank 1000 on its $1000\times1000$ grid) continuous function effectively highlights high-value regions. By combining this TT-guided initialization with UCB-driven tree search, TTTS achieves both efficient exploration and global convergence.
\minew{In both regimes, TT-Cross efficiently exploits the correlations among decision variables to construct a global approximation of the entire decision tree from a limited number of function evaluations, which provides informative guidance for the subsequent tree search. A low TT rank may lead the approximation to retain only a subset of the solution modes rather than all of them. However, the retained modes still provide multiple diverse candidates, richer than the single solution provided by local approximation methods. The subsequent tree search and refinement further explore beyond these candidates and can discover solutions not covered by the TT approximation.}

\section[TTTS for Generalized Robot Optimization]{TTTS for Generalized Robot Optimization}
\label{sec:robot_problem}

We now instantiate \eqref{eq:general_opt} for generalized robot optimization, such as inverse kinematics, motion planning, multi-stage motion planning, and model predictive control. These problems correspond to a large set of mathematical programs, including nonlinear programming (NLP), large-scale (a.k.a.\ high-dim) NLP and mixed-integer nonlinear programming (MINLP). In particular, we leverage basis functions to reduce the dimensionality in large-scale NLP formulations, and the choice of basis functions remains flexible.

We describe the notation and variables:

\begin{itemize}
\item  $K$  the number of discrete modes (a.k.a.\ stages).
\item  $m_k \in \mathcal{M}$  the discrete \emph{mode} at stage $k$, chosen from a finite (or countable) set $\mathcal{M}$.
\item  $a_k \in \mathcal{A}$  the discrete action at stage $k$, chosen from the finite action set $\mathcal{A}$.
\item $\bm{x}_k^t \in  \Omega_{\bm{x}} \subseteq \mathbb{R}^n$  a continuous state (or configuration) at stage $k$ at time $t$.
\item $\bm{u}_k^t \in  \Omega_{\bm{u}} \subseteq \mathbb{R}^p$ a continuous control input at stage $k$ at time $t$.
\item $T$ trajectory length for each stage.
\item $B = B_1 + \cdots + B_K$, the total number of basis functions, where $B_k$ denotes the number of basis functions at stage $k$.
\item $\bm{\Psi}_k \in \mathbb{R}^{T \times B_k}$ a chosen set of basis functions for stage $k$, used to reconstruct the continuous control from weights.
\item $\bm{w}_k \in \Omega_{\bm{w}} \subseteq  \mathbb{R}^{B_k}$ the vector of basis \emph{weights} at stage $k$. We let $\bm{u}^{[T]}_k \;=\; \bm{\Psi}_k\, \bm{w}_k$, encoding the continuous trajectory in basis form.
\item $c(m_k, a_k, x_k^t, u_k^t)$ stage cost (e.g., energy, distance, penalty) at time $t$ in stage $k$.
\item $c_{\text{terminal}}(x_K^T)$ terminal cost, e.g.\ capturing the final configuration error.
\item $\phi(\cdot) \le 0$, $\psi(\cdot) = 0$ general inequality/equality constraints (kinematic limits, collision avoidance, dynamics, boundary conditions).
\end{itemize}

The decision variable is $\bm{x} = (a_1, \ldots, a_K,, \bm{w}_1, \ldots, \bm{w}_{B})$, collecting both the discrete action modes and the continuous basis weights. We use a bracket subscript for sequences (e.g., $\bm{x}_{k}^{[T]}$ represents the full trajectory at stage $k$, and $\bm{u}_{[K]}^{[T]}$ represents the control variables across all stages and time steps). The robot optimization problem is:

\begin{align}
\label{eq:general_opt_robot}
\min_{\substack{a_{[K]}, \bm{u}_{[K]}^{[T]}}}
& \quad
\sum_{k=0}^K \int_{0}^{T} c\bigl(m_k,\;a_k, \;\bm{x}_k^t,\;\bm{u}_k^t\bigr) dt
\;+\; c_{\text{terminal}}\bigl(\bm{x}_K^T\bigr)
\\[6pt]
\text{s.t.}
& \quad
\bm{u}_{k}^{[T]} = \sum_{b=0}^B \bm{\Psi}_k^b\,\bm{w}_k^b,
\label{eq:basis_encoding_robot}\\
& \quad
\forall_{k=0}^{K} \quad  m_{k+1} \;\in\; \text{succ} \; \!\bigl(m_k,\;a_k,\; \bm{x}_k^{[T]},\;\bm{u}_k^{[T]}\bigr),
\label{eq:mode_transition_robot}\\
& \quad
\forall_{k=0}^{K} \quad  \phi\bigl(m_k,\;a_k,\; \bm{x}_k^{[T]},\;\bm{u}_k^{[T]}\bigr) \;\le\; 0,
\label{eq:ineq_constraints_robot}\\
& \quad
\forall_{k=0}^{K} \quad  \psi\bigl(m_k,\;a_k,\; \bm{x}_k^{[T]},\;\bm{u}_k^{[T]}\bigr) \;=\; 0,
\label{eq:eq_constraints_robot}\\
& \quad
(m_0,\; \bm{x}_0^0) = (m_{\text{init}},\; \bm{x}_{\text{init}}),
\end{align}
where:
\begin{itemize}
    \item \eqref{eq:basis_encoding_robot} encodes the full trajectory of decision variables using basis functions.
    \item \eqref{eq:mode_transition_robot} enforces the allowed transition to $m_{k+1}$ in the discrete mode set $\mathcal{M}$, given the current mode $m_k$, discrete action $a_k$, and the continuous trajectories $\bm{x}_k^{[T]}$, $\bm{u}_k^{[T]}$.
    \item \eqref{eq:ineq_constraints_robot} and \eqref{eq:eq_constraints_robot} represent the system dynamics, consistency of different modes, and other physical constraints.
    \item $m_{\text{init}}$ and $x_{\text{init}}$ denote the initial mode and state.
\end{itemize}

This formulation covers a wide range of robotic tasks:

\paragraph{Inverse Kinematics (IK)} Set $T=1$, $K=1$. The decision variable is $\bm{w}_1$ with $\bm{u}_1 = \bm{\Psi}_1\bm{w}_1$, and the cost measures end-effector positioning error subject to joint-limit and collision constraints.

\paragraph{Motion Planning (MP)} Set $K=1$ with trajectory length $T$. The decision variable is $\bm{w}$ encoding the full joint trajectory via basis functions.

\paragraph{Multi-stage Motion Planning (MsMP)} Use $K$ stages, where at each stage $k$ the discrete action $a_k\in\mathcal{A}$, mode $m_k\in\mathcal{M}$, and continuous trajectory $\bm{u}_k^{[T]}=\bm{\Psi}_k\bm{w}_k$ must satisfy collision-free motion, piecewise dynamics, and valid contact transitions.

To apply TTTS to \eqref{eq:general_opt_robot}, we construct the search space $\Omega_x$ as a $d = K + B$ layer decision tree: the first $K$ layers index the discrete action choices (each with $|\mathcal{A}|$ options), and the remaining $B$ layers index the discretized basis weights (each with $N$ grid points). TTTS then optimizes over this index space as described in Section~\ref{sec:ttts}, recovering the optimal $\bm{x}^* = (a^*_{[K]}, \bm{w}^*)$, followed by CMA-ES refinement of the continuous weights.

\section{Experimental Results on Generalized Robot Optimization}

The toy examples in Section~5.2 establish the algorithm's core properties on simple function optimization. In this section, we further evaluate TTTS on a diverse set of robotic tasks to demonstrate its broad applicability. We conduct ablation studies to assess the contribution of each component and compare against state-of-the-art baselines. Hyperparameters and cost functions for each task are detailed in the appendix.

\subsection{Inverse kinematics}



Building on the 2-DOF IK example in Section~\ref{sec:walkthrough}, we here evaluate TTTS on 3-DOF and 7-DOF manipulators (denoted \emph{IK1} and \emph{IK2}) with collision avoidance. This corresponds to $T = 1$, $K = 1$ in \eqref{eq:general_opt}, with continuous decision variables $\bm{u}$ subject to joint limits, collision avoidance, and reachability constraints.

\begin{figure*}[htbp]
  \centering
  \includegraphics[width=0.8\textwidth]{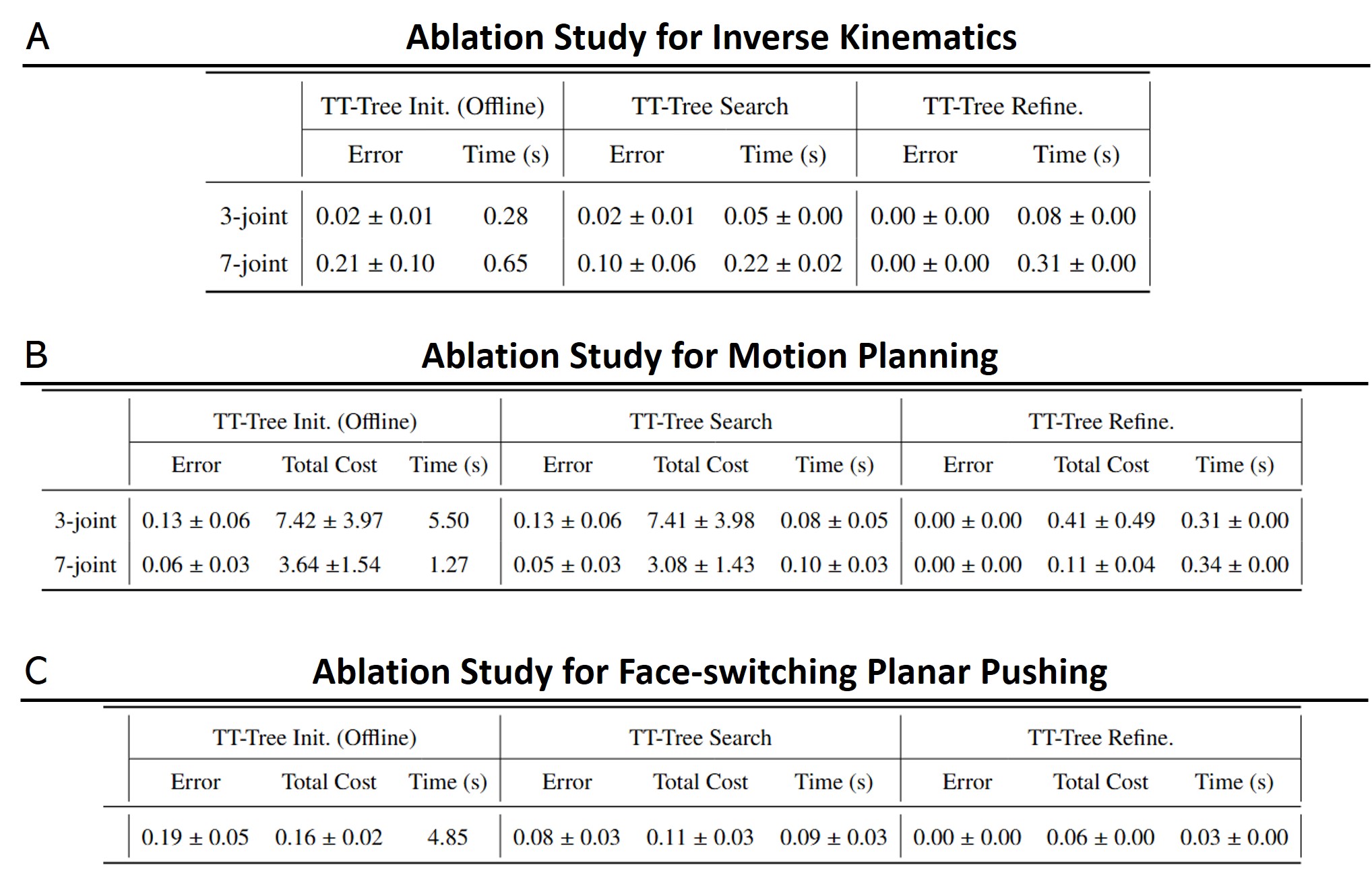}
  \caption{\textbf{Ablation studies for diverse robotics tasks.} \emph{Error} denotes the $\ell_2$ norm between the final and target configurations. 
    \emph{Total Cost} refers to the value of the cost function for the obtained solution.
    \emph{Time} indicates the net algorithmic runtime required to find the solution.}
  \label{fig:ablation_study}
\end{figure*}

We randomly generated five targets in the ablation study to analyze the necessity of each component of TTTS, which consists of three main components: TT-Tree Initialization, TT-Tree Search, and TT-Tree Refinement. TT-Tree Initialization typically requires some computation time due to TT-Cross approximation (particularly for high-dimensional systems), but this process is performed offline. At this step, the state space is augmented with task variables, enabling rapid task-conditioned retrieval for TT-Tree Search. In our experiments, we set $r_{\max} = 21$ for both the 3-joint and 7-joint manipulators (which we refer to as \emph{IK1} and \emph{IK2} in Figure \ref{fig:fullcomp}), resulting in a coarse approximation of the full decision tree. Notably, obstacle avoidance tends to make the TT approximation less accurate. As shown in Figure \ref{fig:ablation_study} (A), TT-Tree Initialization alone does not yield the global optimum, highlighting the limitations of using TT approximation by itself. However, after performing TT-Tree Search and Refinement, the final task-space error is zero, indicating that the globally optimal solution was successfully found.

We further compared TTTS against state-of-the-art baselines: TTGO, MCTS, and CMA-ES. TTGO and MCTS can each be viewed as special cases of TTTS. TTGO relies on direct TT sampling instead of UCB-driven tree search, while MCTS operates without TT guidance. CMA-ES is a gradient-free, sampling-based method well-suited to robotic problems with nonlinear objectives and obstacle avoidance constraints. As shown in Figure~\ref{fig:fullcomp}, TTTS, MCTS, and CMA-ES all recover the global optimum, whereas TTGO does not, as it lacks UCB-driven exploration and therefore does not achieve asymptotic convergence. Among the three successful methods, TTTS is the fastest, confirming that TT guidance substantially accelerates the search. 


\subsection{Motion planning around obstacles}
We further applied our framework to a motion planning problem in which a robot must generate a smooth, collision-free trajectory from a given start to a goal configuration. The problem is formulated over \( T \) time steps, where the continuous trajectory is represented using basis functions as $\bm{u}^{[T]} = \sum_{b=1}^B \bm{\Psi}^b\,\bm{w}^b$, and the optimization variables are the corresponding weights \( \{\bm{w}^b\}_{b=1}^B \). The objective is to minimize a cost function that encourages smooth motion (e.g., penalizing velocity or acceleration), while ensuring accurate goal reaching. The constraints include joint limits and collision avoidance with static obstacles.

\begin{figure*}[htbp]
  \centering
  \includegraphics[width=1.0\textwidth]{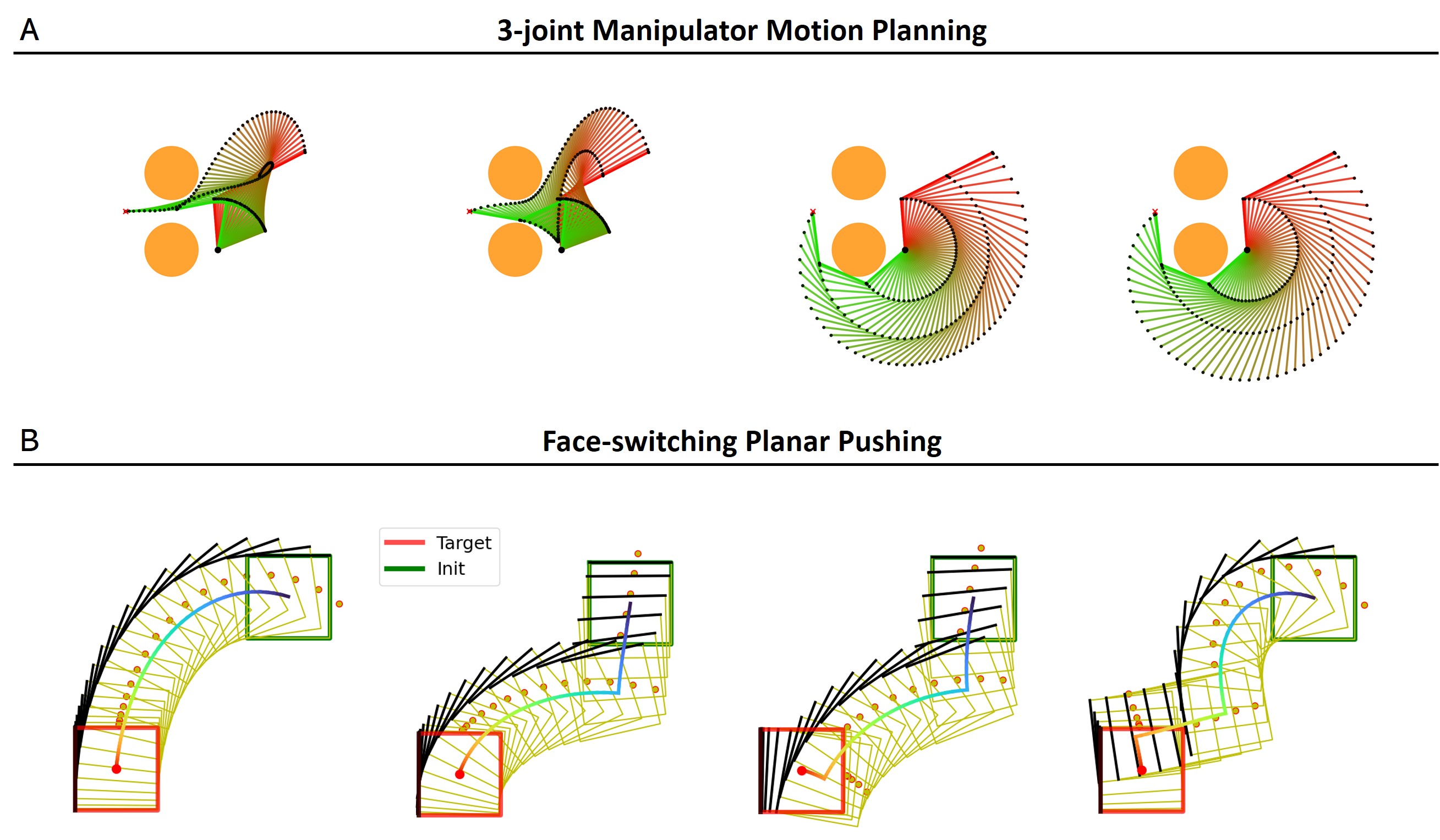}
  \caption{\textbf{multimodal solutions for motion planning around obstacles and face-switching planar pushing.} (A) 3-joint manipulator motion planning with multimodal solutions. The trajectory is visualized using a red-to-green color spectrum to indicate temporal evolution. (B) Face-switching planar pushing task with multimodal solutions. Given the same initial configuration $[0.25, 0.25, -\frac{\pi}{2}]$ and target $[0.25, 0.25, -\frac{\pi}{2}]$, our algorithm can find diverse solutions to accomplish the task, by jointly optimizing over discrete contact faces and continuous motion variables. The black edge of the rectangle indicates the cube’s orientation. The number of face switches varies from $0$ to $2$.}
  \label{fig:multi_modal}
\end{figure*}

Figure \ref{fig:multi_modal} (A) presents our first test scenario: a 3-joint manipulator reaching task, which we refer to as \textit{MP1}. This task is particularly challenging because the target lies on the opposite side of the robot, with two circular obstacles obstructing the direct path. The trajectories found by our approach exhibit multiple solution modalities under the same initial configuration and target. Notably, the first two trajectories shown in the figure require the manipulator to initially move away from the target and then pass through a narrow passage between the obstacles, which is not easy to find and requires long-horizon anticipation. This setting involves numerous local optima and a narrow feasible solution space, necessitating long-horizon planning rather than short-horizon control. The results illustrate our framework's ability to overcome local optima and support long-term decision-making. We further applied our approach to a 7-joint manipulator reaching task, which we refer to as \textit{MP2}. The robot arm must move its end-effector from one level of the shelf to another while avoiding obstacles, such as the shelf frame, across its entire body surface. This setup is representative of daily tasks such as pick-and-place or bookshelf arrangement. Figure \ref{fig:div_domain} (B) shows the resulting trajectories, where end-effector paths are visualized as curves. Different colors indicate distinct solutions discovered by the algorithm.

Figure \ref{fig:ablation_study} (B) presents the ablation study of TTTS applied to motion planning (MP) problems. As the three stages of TTTS are applied, both the final state error and the total trajectory cost consistently decrease, demonstrating the effectiveness of each component. The final errors for both tasks are zero, indicating that a collision-free joint trajectory was successfully found to reach the target. An interesting observation is that TT-Tree Initialization takes longer for task \textit{MP1} than for \textit{MP2}, as \textit{MP1} is a more challenging problem with more local optima. Accordingly, we set $r_{\text{max}} = 41$ for \textit{MP1} and $r_{\text{max}} = 21$ for \textit{MP2}. The TT approximation in \textit{MP1} achieves high accuracy, and further iterations of TT-MCTS provide limited improvement. This suggests that, with sufficient storage capacity, we can use higher TT ranks for more accurate tree approximation, accelerating online inference. In contrast, the results of \textit{MP2} highlight the effectiveness of TTTS under limited storage conditions, where low-rank TT approximations still capture informative representations of complex decision trees. The integration of TT approximation with UCB-driven exploration enables TTTS to effectively search complex, multimodal solution space and reliably converge to high-quality solutions.

Figure~\ref{fig:fullcomp} compares TTTS with other optimization methods in terms of \emph{Final Error}, \emph{Total Cost}, and \emph{Runtime}. \emph{Final Error} denotes the $\ell_2$ norm between the system’s final and target configurations. \emph{Total Cost} refers to the value of the cost functions evaluated at the obtained solution, with the cost functions described in Section~\ref{sec: cost_functions}. \emph{Runtime} indicates the net algorithmic runtime required to find the solution, excluding forward rollout evaluations (i.e., physics simulator calls), as rollout speed is task-dependent and does not reflect the computational efficiency of the algorithm itself. From the \emph{Final Error} and \emph{Total Cost}, we observe that TTTS achieves results comparable to MCTS when the latter is given sufficient time, highlighting TTTS’s ability to find similar global solutions. The \emph{Runtime} further shows that TTTS requires significantly less computation time. TTGO performs well for \textit{MP2}, but it struggles in \textit{MP1} due to the complex cost landscape, which shows TT approximation alone is insufficient to capture all necessary features. In contrast, TTTS leverages strategic tree search to explore the decision space more effectively, achieving global solutions almost surely. CMA-ES also fails in \textit{MP1} because the narrow passage between obstacles challenges its single-modality evolutionary strategy, causing it to be trapped in local optima. These experiments highlight both the computational efficiency of TTTS, enabled by the TT approximation of the decision tree, and its global solution-finding capability, enabled by TT contractions and the strategic search adopted from MCTS.

\begin{figure*}[htbp]
    \centering
    \begin{minipage}{0.32\linewidth}
        \subfloat[\centering TTTS]{\includegraphics[width=0.95\linewidth]{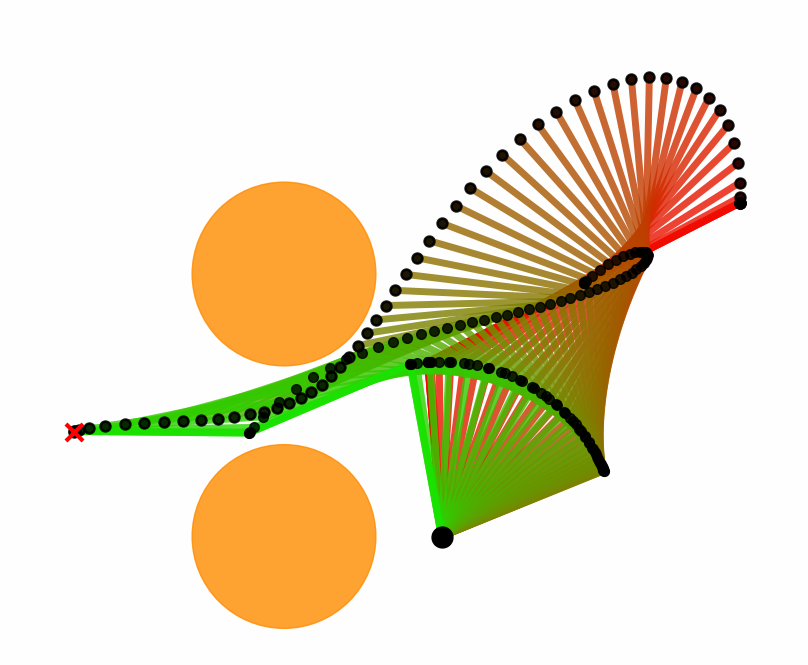}}\label{fig:ttts}
    \end{minipage}
    \hfill
    \begin{minipage}{0.32\linewidth}
        \subfloat[\centering VP-STO]{\includegraphics[width=0.95\linewidth]{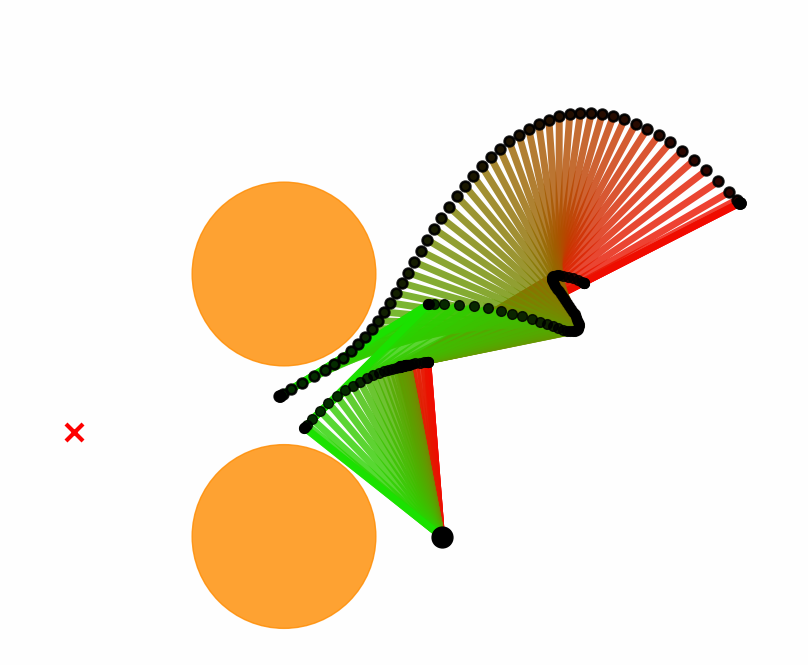}}\label{fig:vpsto}
    \end{minipage}
    \hfill
    \begin{minipage}{0.32\linewidth}
        \subfloat[\centering PRM+TO]{\includegraphics[width=0.95\linewidth]{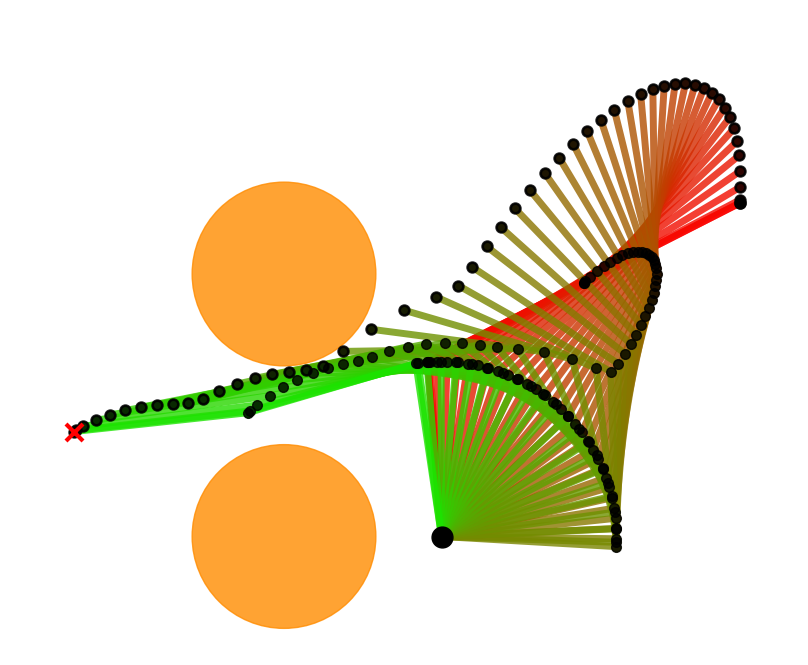}}\label{fig:prm}
    \end{minipage}

    \vspace{1em} 

    \begin{minipage}{0.95\linewidth}
        \centering
        \begin{small}
        \scalebox{0.9}{
        \begin{tabular}{l |c c c|c c c|c c c|}
            \toprule
            & \multicolumn{3}{c|}{TTTS} & \multicolumn{3}{c|}{VP-STO} & \multicolumn{3}{c|}{PRM+TO} \\
            \cline{2-10}
            & {Error} & {Total Cost} & {Comp. Time (s)} & {Error}  & {Total Cost} & {Comp. Time (s)}  & {Error}  & {Total Cost} & {Comp. Time (s)} \\
            \midrule
            {} & 0.00 $\pm$ 0.00 & 0.56 $\pm$ 0.74 &  1.00 $\pm$ 0.03 & 
            0.07 $\pm$ 0.13  & 3.67 $\pm$ 6.85 & 2.52 $\pm$ 2.43 & 
            0.00 $\pm$ 0.00  & 0.41 $\pm$ 0.51 & 1.87 $\pm$ 0.34 \\
            \bottomrule
        \end{tabular}
        }
        \end{small}
    \end{minipage}

    \caption{\textbf{Comparison of TTTS, VP-STO, and PRM+TO.} The red cross indicates the reaching target. TTTS and PRM+TO successfully generate optimal manipulator trajectories, while VP-STO fails in the narrow passage. The table reports reaching error, control cost, and computation time, highlighting TTTS’s superior efficiency.}
    \label{fig:comp_vpsto_prm_table}
\end{figure*}

In addition, we compared TTTS with two widely-used approaches for obstacle-avoidance motion planning: VP-STO \citep{jankowski2023vp} and the Probabilistic Roadmap combined with trajectory optimization (PRM+TO) \citep{gasparetto2015path}. To evaluate performance, we randomly generated five targets and applied the approaches to compute the joint trajectories required to reach them. The comparison results, including \emph{reaching error}, \emph{total cost}, and \emph{computation time}, are reported in Figure~\ref{fig:comp_vpsto_prm_table}. We also visualize the manipulator trajectories for one of the targets, where TTTS and PRM+TO both find a solution, while VP-STO struggles with the narrow passage. From the table, we observe that TTTS achieves performance comparable to PRM+TO while requiring less computation time, thereby demonstrating its computational efficiency. In contrast, VP-STO results in higher error due to its reliance on a good initial guess and its inability to handle multimodal solution spaces.

\subsection{Legged Robot Manipulation}

In addition to motion planning around obstacles, we also evaluated our approach on a contact-rich manipulation task using a legged robot \citep{zhu2023efficient}. As illustrated in Figure~\ref{fig:div_domain}(C), a solo robot manipulates a cube of size $10 \times 10$ cm under a joint impedance controller in the Genesis simulator \citep{Genesis}. The cube, weighing $0.05$kg, is required to pivot about the $y$-axis by a certain angle. This task is challenging because the robot must coordinate contact interactions with the object while maintaining stability, handle the hybrid dynamics arising from intermittent contacts, and plan a smooth motion under impedance control. Small errors in trajectory generation can cause the cube to slip, fail to pivot, or destabilize the supporting leg. To solve this task, we define the cost function as a weighted sum of the terminal pose error and the average cube velocity, with details provided in the appendix.
During planning, we optimize the Cartesian-space position trajectory for each leg tip. Both legs are then controlled via impedance control with default parameters of $K_p = 100~\text{N/m}$ and $K_v = 10~\text{N·s/m}$. To enable stable pivoting against gravity, we increase the stiffness of the left leg to $K_p = 2000~\text{N/m}$ along the $y$- and $z$-axes.


Figure~\ref{fig:fullcomp} reports the comparison of TTTS with other methods on this task. TTTS and MCTS achieve similar performance in terms of reaching error and total cost, highlighting TTTS’s asymptotic completeness comparable to MCTS, while TTTS requires significantly less computation time owing to TT approximation of the decision tree. In contrast, TTGO and CMA-ES perform worse. This further underscores the importance of TT factorization and strategic search for effectively solving this challenging task.

\subsection{Multi-stage motion planning}

Multi-stage motion planning encompasses a broad class of realistic robotic problems, such as multi-stage forceful manipulation \citep{holladay2024robust} and multi-primitive sequencing \citep{Xue24RSS}. The objective is to generate a trajectory that enables a robot to interact intelligently with its environment, typically involving contact mode switches (e.g., sticking, sliding, or transitioning between different manipulation primitives). In this article, the trajectory is parameterized by a sequence of basis weights ${\bm{w}_k}$, such that the continuous state at each stage is given by $\bm{u}_k = \bm{\Psi}_k \bm{w}_k$. Discrete modes ${m_k}$ represent different contact or task-specific phases. The optimization aims to minimize a total cost consisting of smoothness penalties, control effort, and task-specific objectives, while satisfying constraints such as collision avoidance, dynamics, and mode transitions.

We use face-switching planar pushing \citep{doshi2020hybrid, xue2023guided} as a representative hybrid manipulation task. In this problem, a robot needs to push a cube from an initial configuration to a target configuration under underactuated dynamics, while deciding both the continuous end-effector velocity trajectory and the discrete contact-face sequence. This task is difficult because the nonlinear dynamics and hybrid discrete-continuous decision structure pose significant challenges to both gradient-based and sampling-based optimization methods. The purpose of this experiment is to evaluate whether our approach can efficiently solve hybrid decision-making problems involving nonlinear dynamics. We formulate this task as an optimization problem with $4$ discrete contact-face variables, each taking values in $\{0,1,2,3\}$, and $4$ continuous B-spline control weights that parameterize the continuous motion using basis functions. After discretizing each continuous dimension into $30$ grid points and allowing up to $4$ contact-face selection stages, each with $4$ possible face choices, the resulting MINLP has a search space of $4^4 \times 30^4$ grid points. As shown in Fig.~\ref{fig:multi_modal}(B), our method discovers multiple cube trajectories for the same initial and target configurations, highlighting the multimodal structure of the solution space. These solutions exhibit different numbers of face switches, ranging from $0$ to $2$, together with diverse and smooth continuous trajectories.


Figure \ref{fig:ablation_study} (C) presents the ablation study for this task. The TT approximation of the decision tree is obtained with a rank setting of $r_{\text{max}} = 41$. The results show that the TT approximation alone is not sufficiently accurate to capture the full landscape of the objective function. However, the subsequent tree search significantly improves the solution quality, and the final refinement step enables convergence to the global optimum. An interesting observation is that, compared with the policy learning formulation (i.e., infinite-horizon dynamic programming) proposed in \citep{Xue24CORL, Xue25IJRR}, this finite-horizon planning formulation requires a significantly higher TT rank to accurately represent the objective function. This is because trajectory-level planning uses decision variables (i.e., basis weights) that influence the entire trajectory, where small changes can induce large, structured variations, necessitating higher representational capacity to capture complex temporal dependencies. This further motivates the necessity of combining TT approximation with tree search, rather than relying on TT alone.

\begin{figure*}[htbp]
  \centering
  \includegraphics[width=1.0\textwidth]{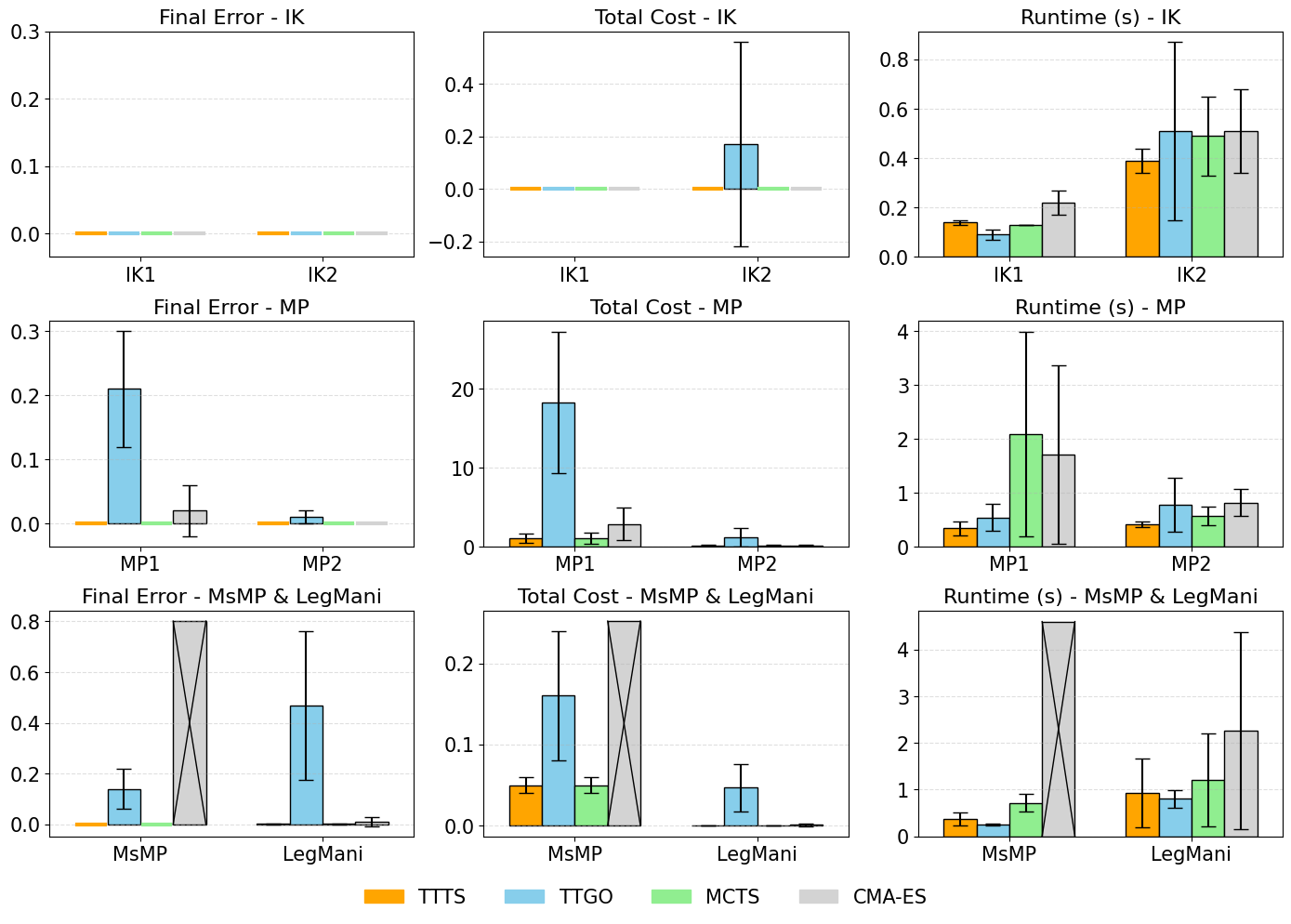}
  \caption{\textbf{Comparison of TTTS, TTGO, CMA-ES, and MCTS with respect to final reaching error, total control cost, and runtime.} \textit{IK1} and \textit{IK2} represent the 3-joint and 7-joint inverse kinematics tasks, respectively. \textit{MP1} and \textit{MP2} correspond to the 3-joint and 7-joint manipulator reaching tasks with obstacle avoidance. \textit{MsMP} denotes the planar pushing task with a face-switching mechanism, and \textit{LegMani} refers to the legged robot manipulation (cube pivoting) task. Diagonal patterns in the bar charts indicate that CMA-ES is incompatible with \textit{MsMP} task.}

  \label{fig:fullcomp}
\end{figure*}

Figure~\ref{fig:fullcomp} compares TTTS with other approaches. CMA-ES struggles with this problem (indicated by diagonal hatching in the bar chart) because it involves both discrete and continuous decision variables. Both TTTS and MCTS successfully reach the final target, but TTTS requires less computation time owing to the TT factorization of the decision tree. TTGO is highly efficient in terms of runtime, but its solutions are less accurate due to the absence of the strategic search incorporated in TTTS. Overall, these experiments highlight both the computational efficiency and the global solution-finding capability of TTTS.



\subsection{Comparison with Neural-Guided MCTS and MINLP Solvers}

\minew{We compare TTTS with AlphaZero-MCTS \citep{silver2017mastering}, NN-surrogate MCTS, plain MCTS, and Differential Evolution (DE) for MINLP \citep{storn1997differential,virtanen2020scipy}. The two neural-guided baselines use neural networks to guide MCTS in different ways. AlphaZero-MCTS trains policy and value networks through self-play to guide node selection and leaf evaluation, respectively. NN-surrogate MCTS replaces the TT surrogate with a neural network trained on the same data to approximate the same objective $J(\bm{x})$, while retaining the other search components to isolate the contribution of the surrogate representation. Plain MCTS uses the same decision tree without a surrogate and estimates node values through rollouts evaluated with the true objective. The planar pushing task and 3-joint manipulator motion planning are used to evaluate the performance of TTTS against these baselines.}

\minew{Unlike TTTS, AlphaZero-MCTS does not approximate the objective $J(\bm{x})$ itself. Instead, it learns policy and value networks that operate on partial assignments of the decision variables, and selects nodes using the PUCT rule. On planar pushing, both methods search over the same decision variables, and refine the continuous candidates with CMA-ES. The architectures, training procedures, query budgets, and search settings are detailed in Table~\ref{tab:neural_mcts_details} in the Appendix.}

\begin{figure*}[t]
  \centering

  \begin{minipage}[b]{0.48\textwidth}
    \centering
    \vspace{0pt}
    \includegraphics[width=\linewidth]{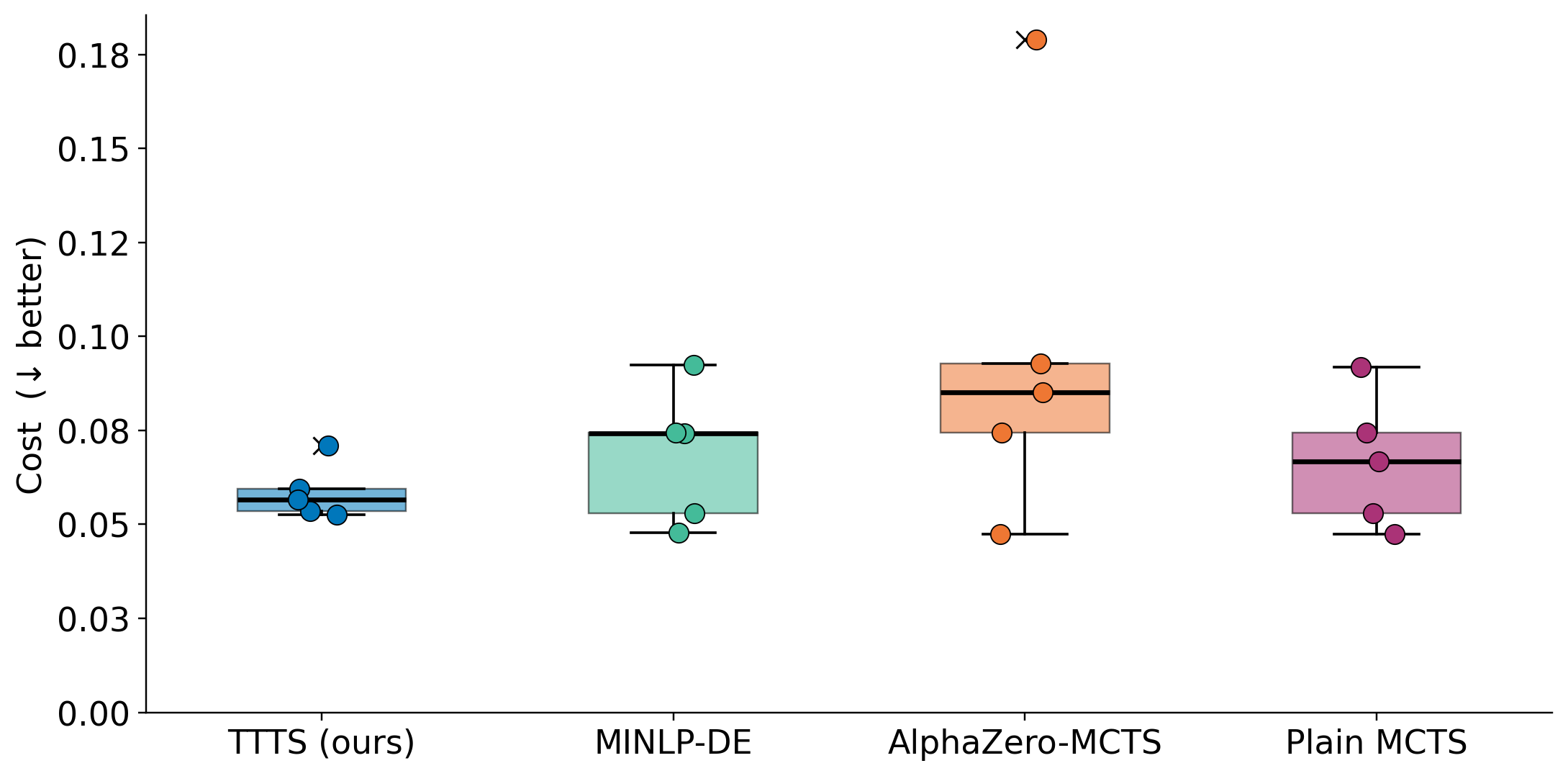}
    \subcaption{Distributions of final trajectory costs across five test scenarios. In most experiments, TTTS produces lower final trajectory costs with less variability than AlphaZero-MCTS, MINLP-DE, and \minew{plain MCTS}.}
    \label{fig:three_way_boxplot}
  \end{minipage}
  \hfill
  \begin{minipage}[b]{0.48\textwidth}
    \centering
    \small
    \begin{tabular}{@{}lccc@{}}
      \toprule
      Method & \begin{tabular}[b]{@{}c@{}}Pretraining\\Time\end{tabular} & \begin{tabular}[b]{@{}c@{}}Online\\Wall Time\end{tabular} & \begin{tabular}[b]{@{}c@{}}Online\\Net Time\end{tabular}\\
      \midrule
      AlphaZero-MCTS & 4.77 h & 26.1 $\pm$ 0.1 s & 0.72 $\pm$ 0.04 s \\
      MINLP-DE & \textbf{0} & 693.4 $\pm$ 7.0 s & 1.67 $\pm$ 0.03 s \\
      \minew{Plain MCTS} & \textbf{\minew{0}} & \minew{711.4 $\pm$ 6.5 s} & \minew{3.54 $\pm$ 1.48 s} \\
      TTTS (ours) & 10.5 s  & \textbf{3.1 $\pm$ 0.0 s} & \textbf{0.12 $\pm$ 0.05 s} \\
      \bottomrule
    \end{tabular}
    \subcaption{\minew{Pretraining time and online search time per scenario. Online wall time includes simulator rollouts; online net time excludes them. TTTS requires less pretraining time than AlphaZero-MCTS and less online search time than all three baselines.}}
    \label{tab:time_comparison}
  \end{minipage}

\caption{\minew{\textbf{Solution quality and computational efficiency on planar pushing.} We compare TTTS with AlphaZero-MCTS, MINLP-DE, and plain MCTS. (a) Final trajectory costs show that TTTS produces higher-quality solutions with less variability. (b) Pretraining and online search times highlight the computational efficiency of TTTS: AlphaZero-MCTS requires extensive offline training, while MINLP-DE and plain MCTS require substantially longer online searches.}}
  \label{fig:cost_time_comparison}
\end{figure*}

Figure~\ref{fig:cost_time_comparison} compares solution quality and computation time. AlphaZero-MCTS performs well in four of the five scenarios but fails in one, with a cost more than four times that of TTTS, resulting in greater variability across scenarios. This suggests that $4.77$ hours of pretraining are insufficient for reliable generalization across all tested instances, although additional training may improve performance. TTTS requires only $10.5$ seconds to construct a global TT surrogate that effectively guides online search. MINLP-DE requires no pretraining but achieves only moderate solution quality despite a substantially longer online search. TTTS gains efficiency by exploiting correlations among decision variables to construct a compact surrogate over the full search space, which directs the search toward low-cost regions with far fewer objective evaluations than MINLP-DE.

We note that AlphaZero-style methods are powerful in highly challenging settings with large decision spaces, such as Go, but require substantial training investment. The robot optimization problems considered in this work are individually of moderate difficulty, yet span a broad and heterogeneous range of task types, making it impractical to invest substantial training effort for each task. TTTS is well suited to this setting, as it replaces sample-inefficient RL training with TT approximation, which can effectively exploit the correlations among decision variables and enable efficient offline construction of a surrogate for subsequent tree search..

\label{sec:tt_vs_nn}
\minew{The comparisons in Fig.~\ref{fig:cost_time_comparison} evaluate the complete pipelines, without analyzing the specific contribution of the TT surrogate. To isolate this contribution, we compare TTTS with NN-surrogate MCTS using identical training data, equal parameter counts, and matched search budgets. The NN surrogate is a multilayer perceptron (MLP) trained on the query points and corresponding objective values collected by TT-Cross during construction of the TT surrogate. NN-surrogate MCTS uses a standard node-based search tree that explicitly stores the visit count $v$ and cumulative reward $q$ in Eq.~\eqref{eq:ucb} for each expanded node, without using a TT representation.}

\minew{Both methods use the UCB rule in Eq.~\eqref{eq:ucb} with $c=3$, but differ in how they use their surrogates to compute node values and sample completions. TTTS computes node values by exact TT contraction, summing the surrogate over all completions of a partial assignment, and samples completions directly from the TT. NN-surrogate MCTS estimates the same node values from neural-network predictions on sampled completions and draws completions through importance resampling, using those predictions as weights. Neither method applies continuous refinement, so the reported costs reflect tree search alone. We evaluate the methods on planar pushing (five initial slider poses) and 3-joint motion planning (three goal configurations), each with 10 random seeds. Appendix~\ref{sec:app_tt_vs_nn} provides further implementation details.}

\begin{figure*}[t]
  \centering
  \includegraphics[width=0.75\textwidth]{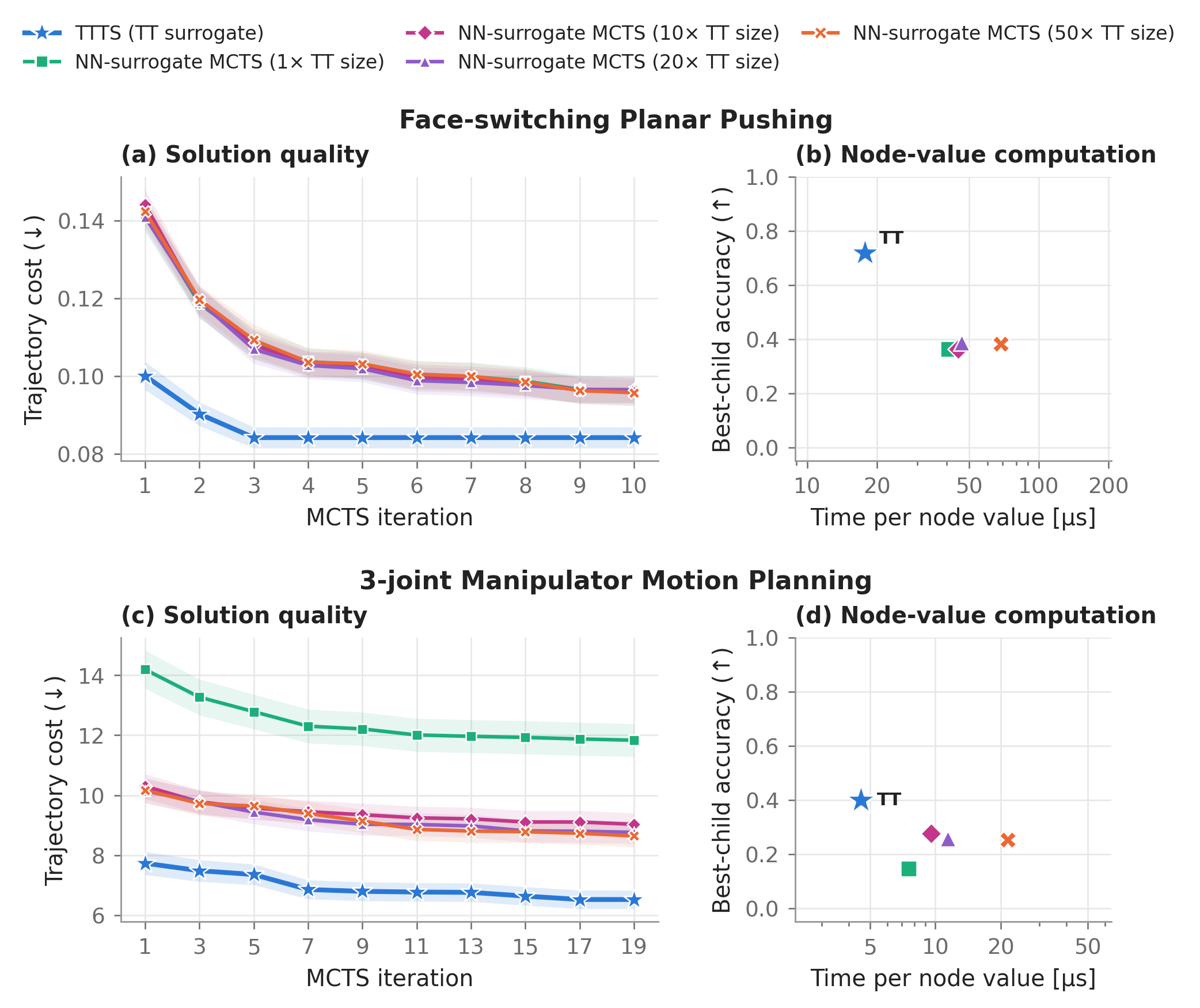}
    \caption{\minew{\textbf{Controlled comparison of TT and NN surrogates for MCTS.} Both methods use the same UCB rule and perform the same number of true objective evaluations per iteration. The surrogates use identical training data, with the NNs trained on all TT-Cross queries. The NNs have 1$\times$ to 50$\times$ as many parameters as the TT. (a, c) Best trajectory cost versus MCTS iteration (mean $\pm$ standard error). (b, d) Node-value accuracy versus computation time per node value. TT node values are computed by TT contraction, whereas NN node values are estimated from sampled completions.}}
  \label{fig:tt_vs_nn}
\end{figure*}

\minew{Figure~\ref{fig:tt_vs_nn} presents comparisons for matched training data, parameter counts, and search budgets, where the NN surrogate is an MLP with three hidden layers and the same number of parameters as the TT surrogate (1$\times$ TT size). Under these conditions, TTTS achieves lower costs than NN-surrogate MCTS at every iteration on both tasks. On planar pushing, TTTS finds a high-quality solution within the first iteration and converges after three iterations, whereas NN-surrogate MCTS yields higher costs and improves only gradually. On 3-joint motion planning, TTTS attains a low cost already in the first iteration and continues to improve, whereas NN-surrogate MCTS remains at substantially higher costs. We further consider NN surrogates with 10, 20, and 50 times the number of parameters of the TT surrogate. The 10$\times$ and 20$\times$ networks are MLPs with six hidden layers, and the 50$\times$ network is a residual MLP with twelve hidden layers. On planar pushing, the larger NNs bring no noticeable improvement. On 3-joint motion planning, they narrow the gap relative to the 1$\times$ MLP, but increasing the size beyond 10$\times$ yields only marginal gains, and even after 20 iterations, none of them reaches the cost that TTTS attains in its first iteration. These results indicate that the TT surrogate provides an effective global approximation of the objective function from limited training data, enabling efficient guidance of the tree search. In comparison, the NN surrogates are trained to fit the objective at the sampled points, but fail to capture the global structure needed for tree-search guidance, even with fifty times as many parameters.}

\minew{Figure~\ref{fig:tt_vs_nn}(b, d) further examines the accuracy and computational cost of node-value estimates obtained from the two surrogates. Each node value aggregates the objective over all completions of a partial assignment. TT contraction (Eq. \eqref{eq:w_compute}) computes this aggregate exactly for the TT surrogate, whereas the NN estimates it from predictions on sampled completions. We assess node-value accuracy using the best-child accuracy, i.e., how often the child with the highest estimated value coincides with the best child under a reference computed with the true objective. Since the search favors the most promising child, a higher best-child accuracy indicates that the surrogate more reliably directs the search toward the right branch. As shown in the figure, TT contraction yields both higher best-child accuracy and shorter computation time per node value on both tasks. On planar pushing, the larger NNs are no more accurate than the 1$\times$ MLP, while being slower. On 3-joint motion planning, the larger NNs improve the accuracy over the 1$\times$ MLP, consistent with their better search performance, but remain less accurate and slower than TT contraction. The efficiency of TT stems from computing the aggregate directly through the TT cores, without explicitly sampling completions. In comparison, NN-based estimation requires generating completed assignments and evaluating the network on these samples, so its computational cost grows with the network size. These results indicate that TT contraction provides more reliable and efficient node values for guiding tree search.}

\minew{These comparisons demonstrate the benefits of TTTS relative to neural-guided baselines in the evaluated tasks. Compared with AlphaZero-MCTS, TTTS reduces offline computation significantly and produces more consistent solutions. Compared with NN-surrogate MCTS under matched data and search budgets, TTTS achieves lower solution costs at every iteration, even against an NN surrogate with fifty times as many parameters, while computing node values more efficiently. These findings highlight two advantages of the TT surrogate in offline construction and online search. First, TT-Cross actively explores the decision tree to construct a global approximation of the objective function, providing informative guidance for subsequent search. Second, the TT representation enables accurate and efficient node-value computation through exact marginalization of the surrogate over unassigned decision variables, identifying the best branch more reliably and at lower computational cost than the NN surrogates trained on the same data.}

\subsection{Analysis of Offline-Online Computation Budget}

\begin{figure}[htbp]
  \centering
  \includegraphics[width=0.5\textwidth]{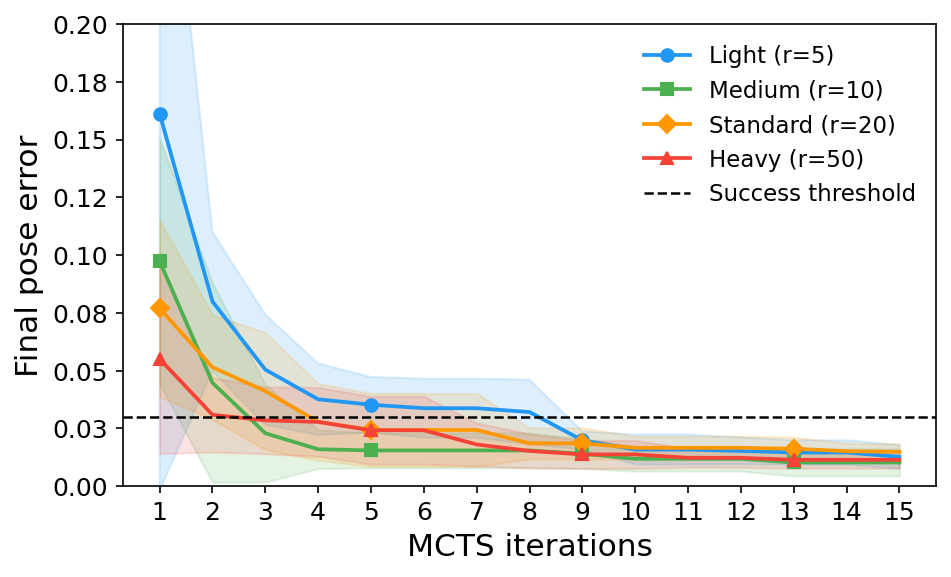}
\caption{Trade-off between offline TT-Cross pretraining and online MCTS search for planar pushing task. Final pose error is plotted against online MCTS iterations for four pretraining budgets (TT rank $r \in \{5,10,20,50\}$). Heavier pretraining (larger $r$) yields a better warm-start and faster convergence to the success threshold (dashed line); all budgets ultimately achieve comparable accuracy below this threshold.}
  \label{fig:online_offline}
\end{figure}

Fig.~\ref{fig:online_offline} illustrates the convergence behavior of MCTS under four offline TT-Cross pretraining budgets, measured by final pose error plotted against online MCTS iterations. The results reveal a clear trade-off: heavier offline pretraining (larger TT rank $r$) yields a substantially better warm-start for online search, enabling faster convergence and fewer iterations to reach the success threshold (dashed line). Specifically, the Heavy budget ($r{=}50$) produces an initial solution close to the threshold at iteration 1 and crosses it within one additional iteration, while the Light budget ($r{=}5$) requires approximately 9--10 iterations to cross the same threshold. Importantly, all budgets ultimately converge to comparable low-error solutions (below 0.02 m) given sufficient online iterations, demonstrating that the method remains applicable even under constrained hardware resources, at the cost of a larger online search budget. This flexibility allows practitioners to shift computation between the offline pretraining phase and the online computation phase according to available resources, making the approach suitable for a wide range of deployment scenarios. Furthermore, since TT approximation is inherently compact and requires only linear memory to approximate the full exponential tree, TTTS is naturally suited for memory-limited hardware.



\subsection{Model Predictive Control for bimanual whole-body manipulation}

We evaluate our approach using a Model Predictive Control (MPC) formulation applied to a bimanual whole-body manipulation task. This task is particularly challenging due to complex contact dynamics between objects, the whole-body geometry of the robot, and interactions with the environment, all of which make accurate modeling difficult. Physical simulators such as MuJoCo \citep{todorov2012mujoco} and IsaacGym \citep{liang2018gpu} can help address these challenges. However, the resulting sim-to-real gap necessitates the use of real-time MPC. Given the simulator as a black-box forward dynamics model, sampling-based MPC becomes a promising approach, as it does not rely on explicit gradient information. Nevertheless, such methods often suffer from high sample complexity and slow convergence, limiting their practicality for real-world deployment. In this experiment, we aim to demonstrate that TTTS can quickly find high-quality solutions to support real-time, sampling-based MPC. Specifically, we use Genesis \citep{Genesis} as the simulator due to its parallel simulation capabilities, and the number of environments is set to $500$.

\begin{figure}[htbp]
  \centering
  \includegraphics[width=0.5\textwidth]{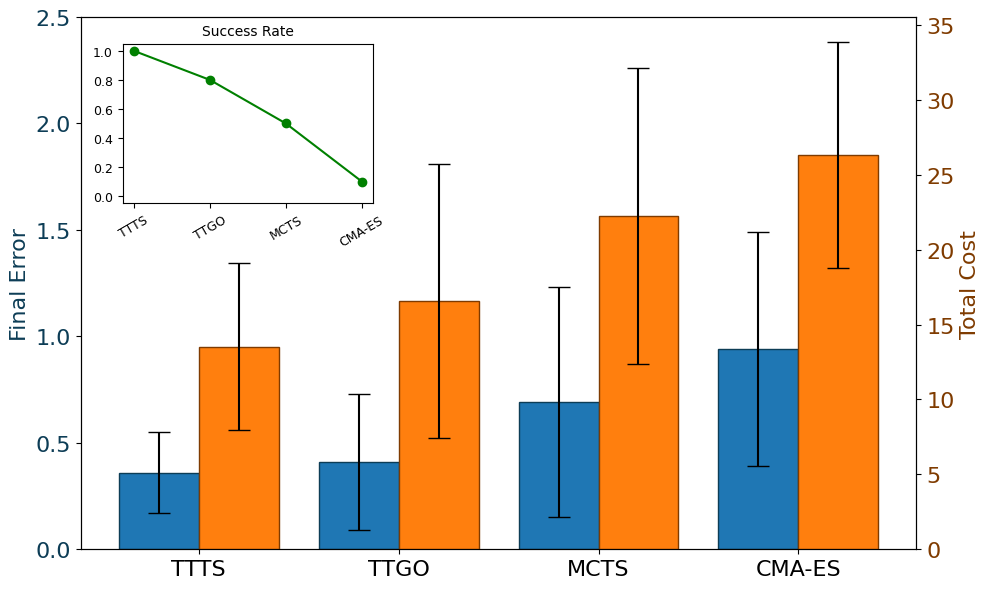}
  \caption{\textbf{Comparison for bimanual whole-body manipulation.} The blue bars represent the final state error achieved by different methods, while the orange bars correspond to the total cost.}
  \label{fig:mpc_bimanual}
\end{figure}

Figure \ref{fig:mpc_bimanual} illustrates the performance differences between TTTS, TTGO, MCTS, and CMA-ES. The computation time is limited to 1 second. A task is considered successful if the angular error at the final timestep satisfies $|\theta - \theta^*| < 3^\circ$, where $\theta$ is the object's final z-axis orientation and $\theta^*$ is the desired target angle. We evaluate five randomly generated configurations and report success rate, final state error, and total trajectory cost. TTGO achieves a decent success rate with a low TT rank ($r_{\max} = 10$), but its limited accuracy leads to occasional failures. MCTS offers asymptotic completeness guarantees but requires more time, making it impractical for real-time MPC. CMA-ES also performs poorly due to slow convergence. In contrast, TTTS first approximates the decision tree in TT format, which accelerates convergence toward promising regions. It then performs a TT-based tree search, enabling efficient exploitation and strategic exploration. The results, including final error and total cost, confirm the effectiveness of TTTS in supporting real-time MPC for contact-rich manipulation.

\subsection{Real-world experiments}

\captionsetup[subfloat]{justification=centering}
\begin{figure*}[htbp] 
    \centering
    \subfloat[]{\includegraphics[width=0.66\columnwidth]{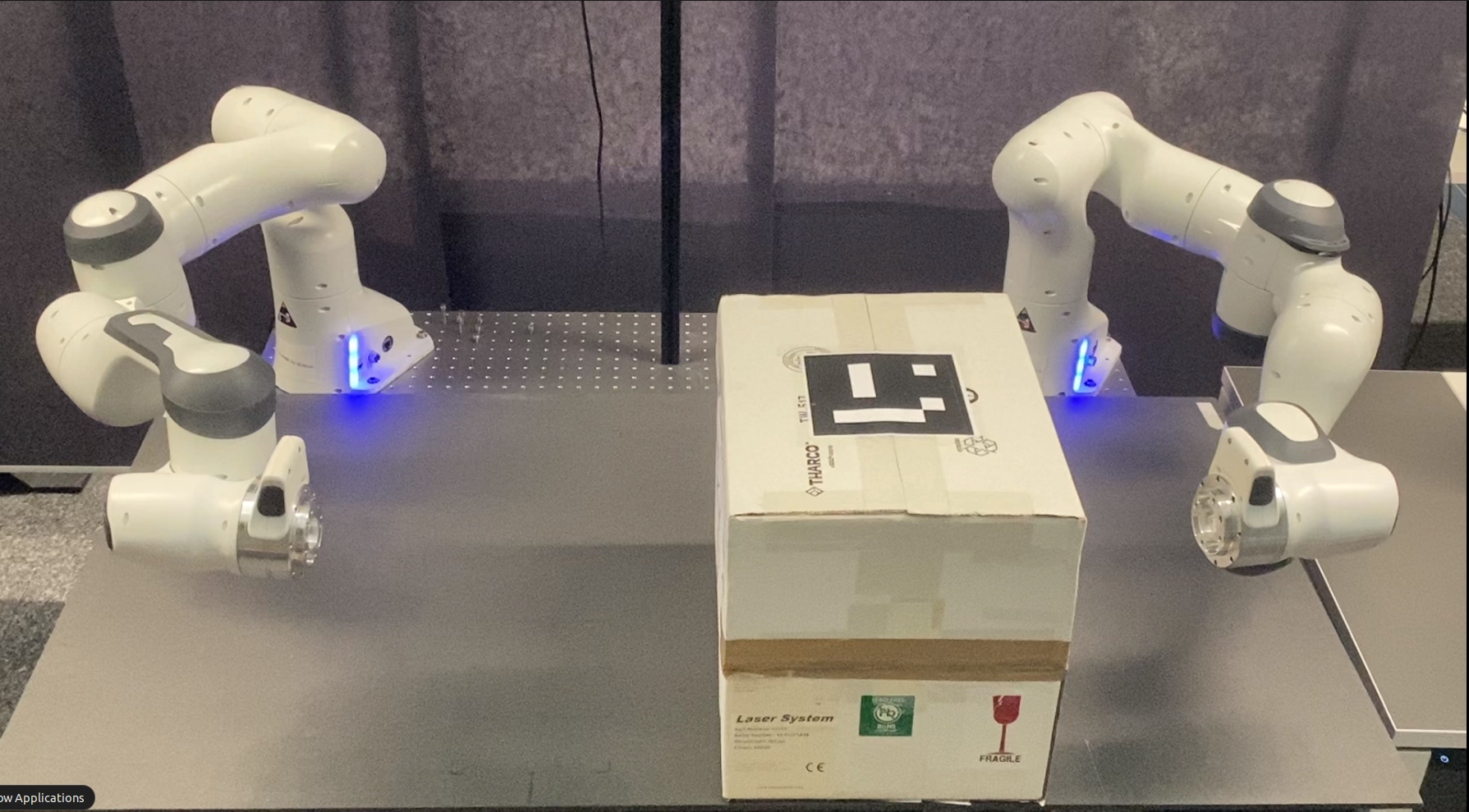}\label{fig:robot_init}}
    \hfill
    \subfloat[]{\includegraphics[width=0.66\columnwidth]{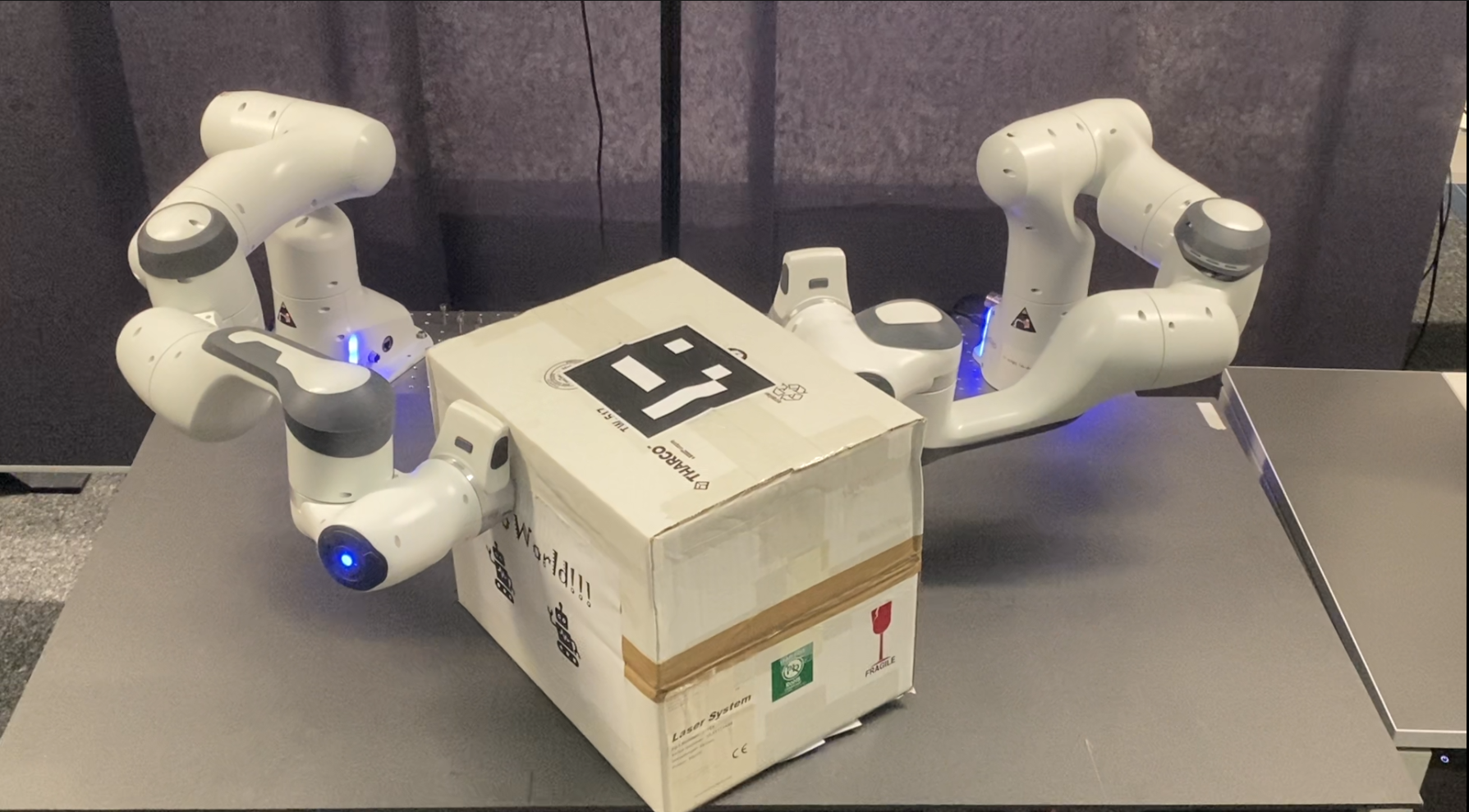}} 
    \hfill
    \subfloat[]{\includegraphics[width=0.66\columnwidth]{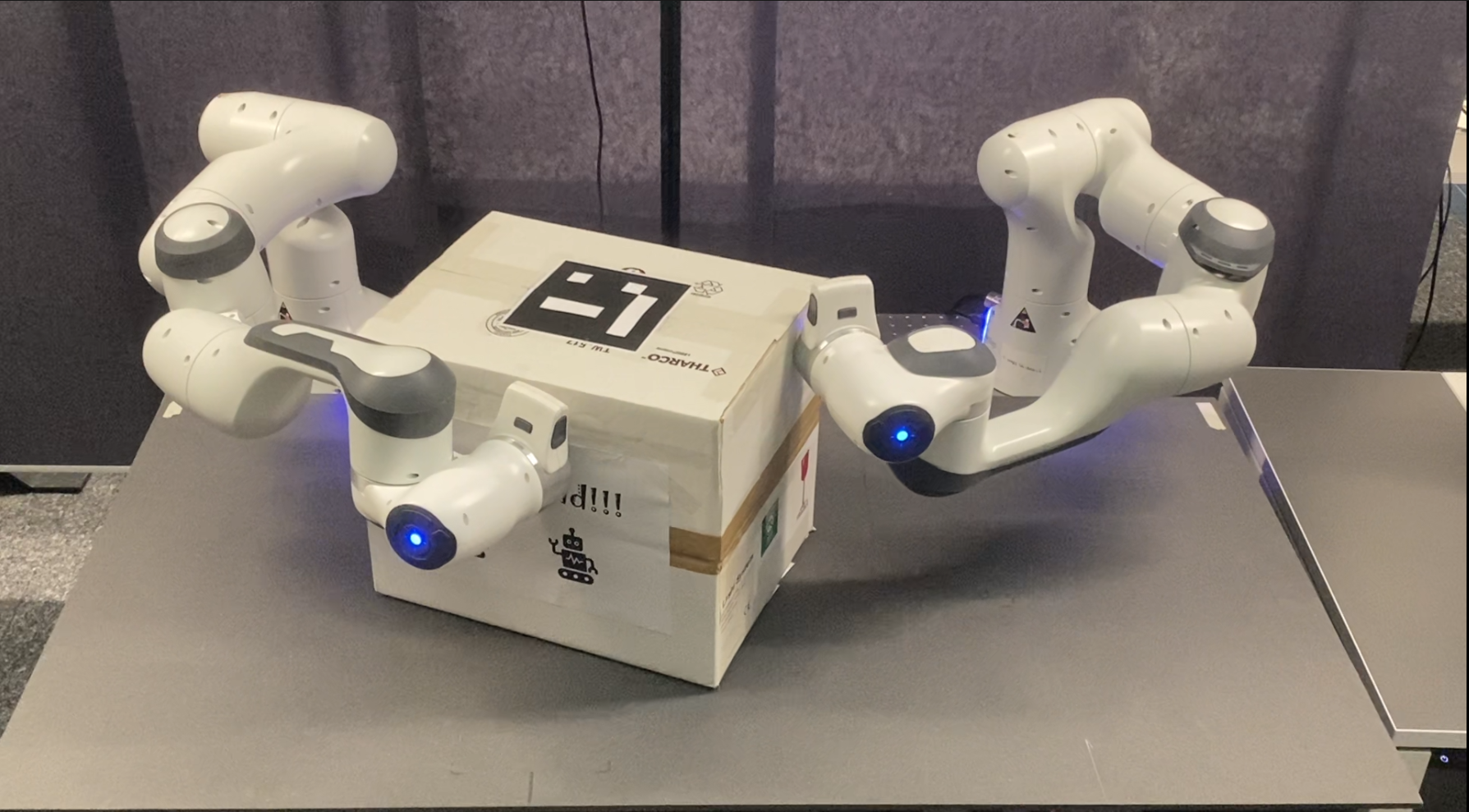}} 
    \\
    \subfloat[]{\includegraphics[width=0.66\columnwidth]{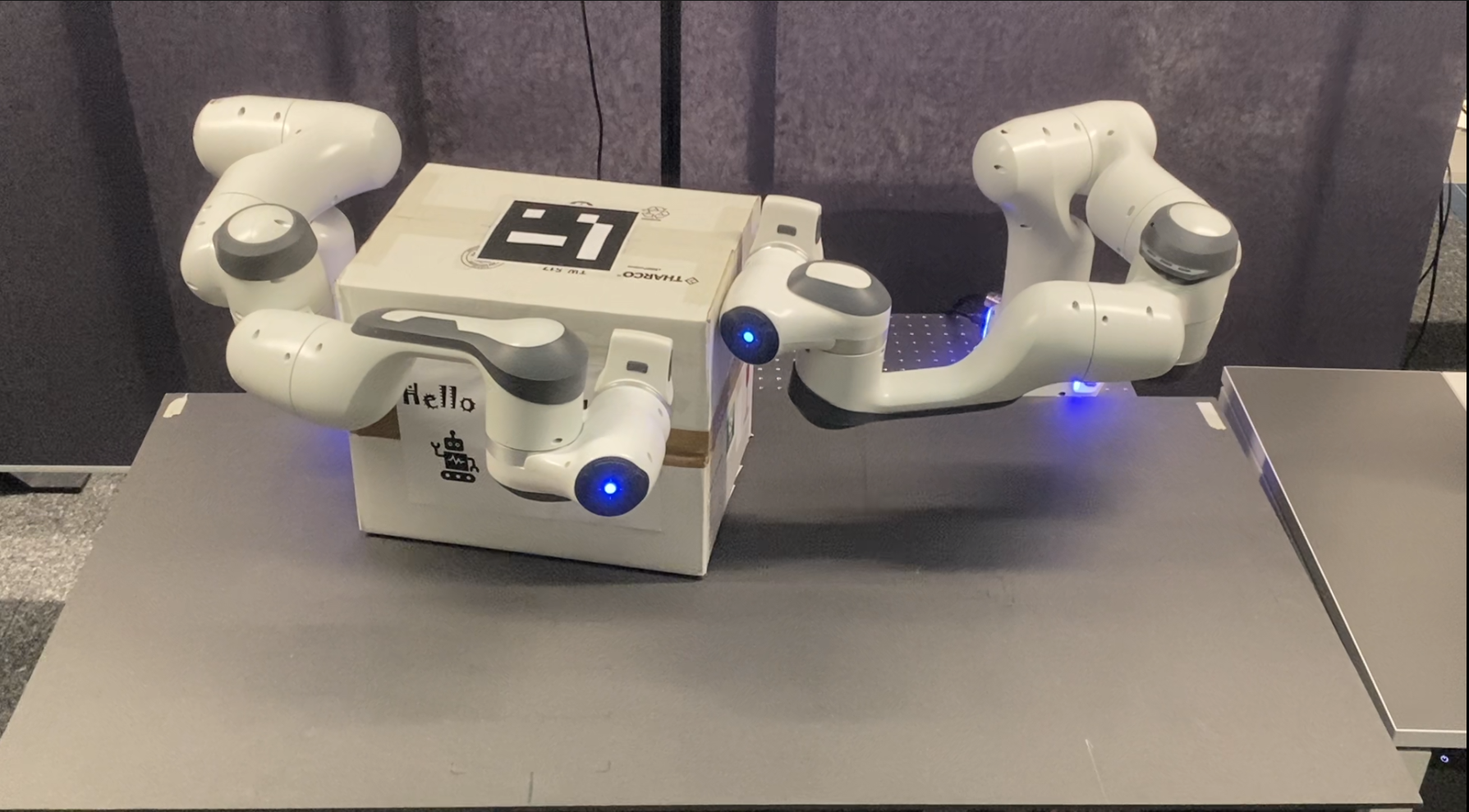}} 
    \hfill
    \subfloat[]{\includegraphics[width=0.66\columnwidth]{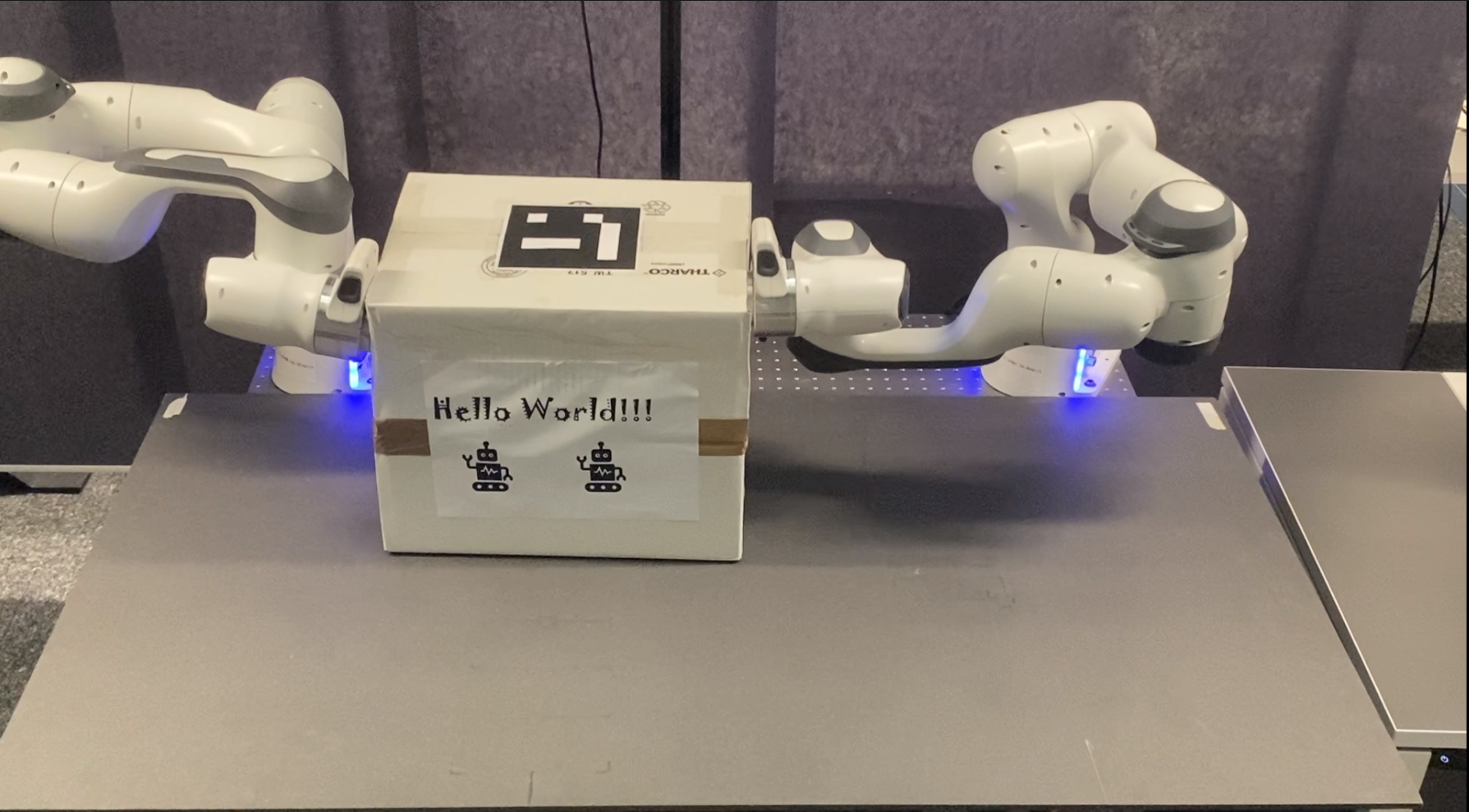}} 
    \hfill
    \subfloat[]{\includegraphics[width=0.66\columnwidth]{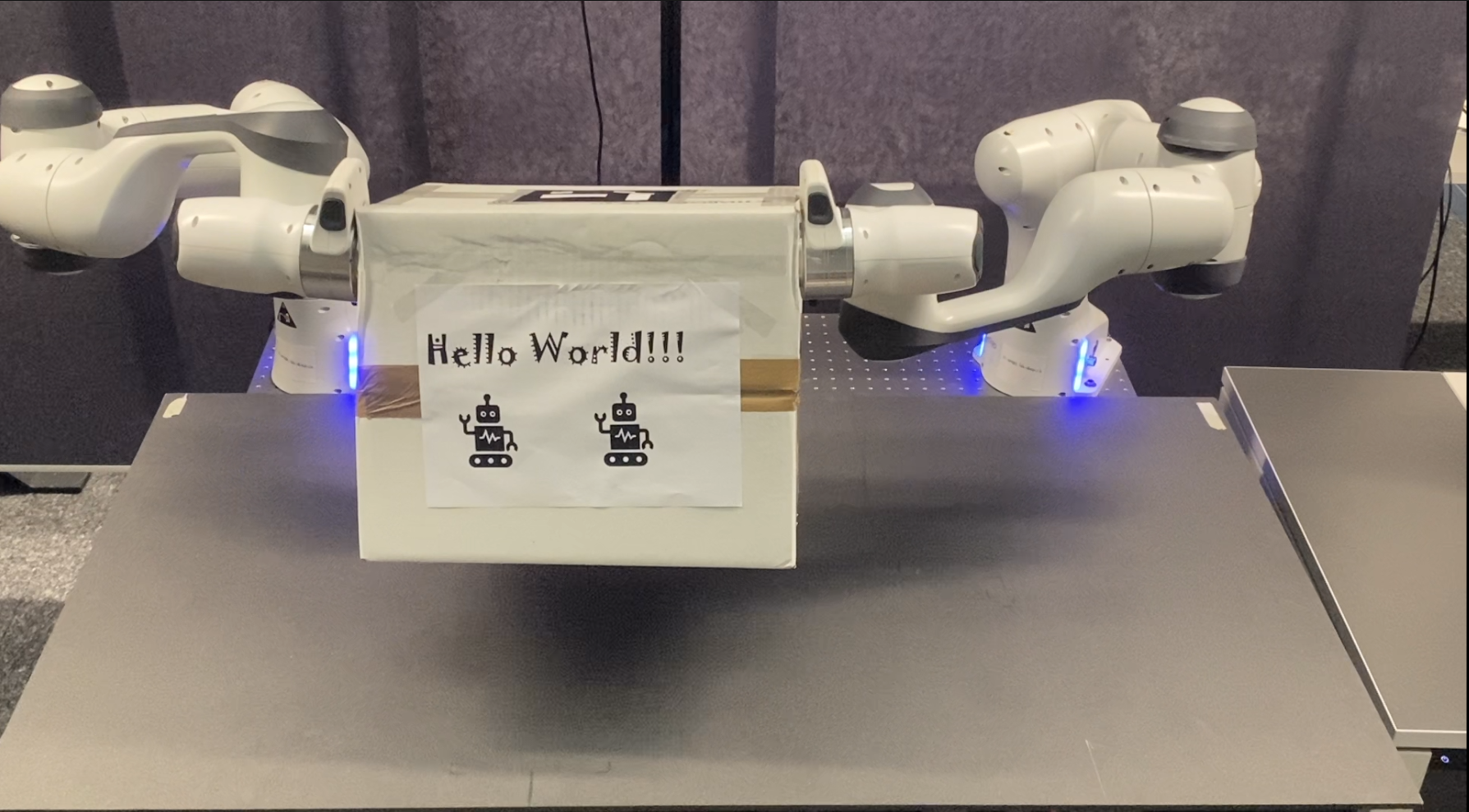}\label{fig:robot_end}}
    \caption{\textbf{Keyframes of bimanual whole-body manipulation.} 
    The system is initialized as (a), and the objective is to rotate the box $90^\circ$ and then lift it as (f). 
    We can observe that the robots can actively exploit whole-body geometry to make and break contacts with the object, completing the overall task successfully.}
    \label{fig:keyframe}
\end{figure*}

\begin{figure*}[htbp]
    \centering
    \begin{subfigure}{0.32\textwidth}
        \includegraphics[width=\linewidth]{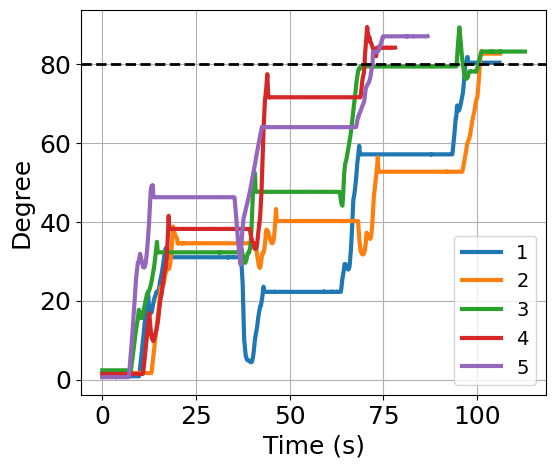}
        \caption{TTTS Results}
        \label{fig:ttts_real_res}
    \end{subfigure}
    \begin{subfigure}{0.32\textwidth}
        \includegraphics[width=\linewidth]{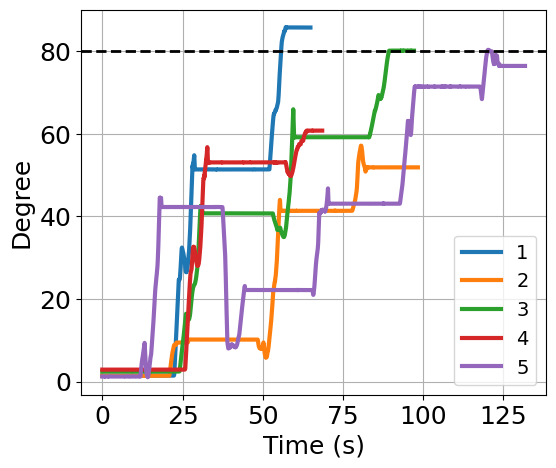}
        \caption{CMA-ES Results}
        \label{fig:cmaes_real_res}
    \end{subfigure}
    \begin{subfigure}{0.33\textwidth}
        \includegraphics[width=\linewidth]{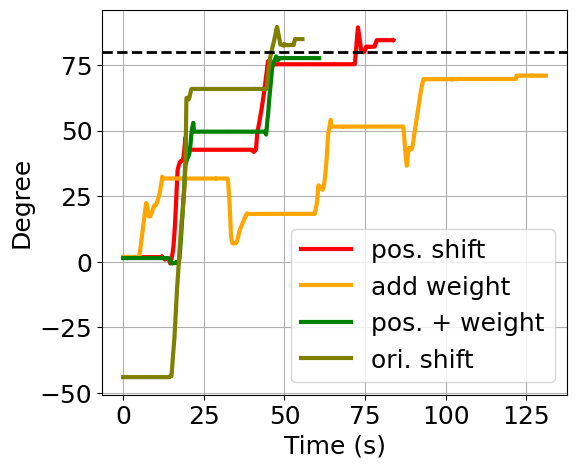}
        \caption{TTTS Robustness Evaluation}
        \label{fig:ttts_robustness}
    \end{subfigure}
    \label{fig:real_analysis}

    \caption{\textbf{Statistical analysis of real-world bimanual whole-body manipulation.}
    The objective is to rotate the box by $90^\circ$. A trial is considered successful if the box is rotated beyond $80^\circ$ (indicated by the dashed line in the plots). 
    (\subref{fig:ttts_real_res}) and (\subref{fig:cmaes_real_res}) present the box orientation trajectories produced by TTTS and CMA-ES, respectively, showing that TTTS achieves a significantly higher success rate. 
    To further assess TTTS robustness, additional experiments were conducted by varying initial position, adding weight on the box, combining both variations, and altering the initial orientation, as illustrated in (\subref{fig:ttts_robustness}).}
\end{figure*}

We validated our approach in the real world through the whole-body bimanual manipulation task. Two 7-DoF Franka robots and a RealSense D435 camera were used. The manipulated object was a large box with the size of $36 \text{cm} \times 26 \text{cm} \times 34 \text{cm}$, which is representative of objects commonly found in warehouse applications. The task was to rotate the box to a target orientation and then lift it. It involved complex contact interactions among the robots, the object, and the table, as well as the full-body surface geometry of the two robot arms, making the system particularly difficult to model.

To address this, Genesis \citep{Genesis} is utilized to predict the box trajectory given a sequence of control commands, thereby eliminating the need for manual modeling of the complex system dynamics. Initially, a TT approximation augmented with the object pose is obtained via TT-Cross. During execution, conditioned on the current object pose, a tree search is performed for 3 iterations. This typically yields effective results due to the guidance provided by the TT approximation.

Following a model predictive control (MPC) paradigm to bridge the sim-to-real gap, a 9-second trajectory is generated at each 1-second time step. Figure \ref{fig:keyframe} presents keyframes from a representative task of rotating the box by $90^\circ$. Notably, the robot's geometry is actively utilized to establish contacts with the object, enabling whole-body bimanual manipulation. Furthermore, a comparative evaluation between TTTS and CMA-ES was conducted across five trials each, with the results summarized in Figure \ref{fig:ttts_real_res} and Figure \ref{fig:cmaes_real_res}. Given a tolerance of $10^\circ$, TTTS achieves a $100\%$ success rate, whereas CMA-ES achieves only $40\%$ under similar computational budgets. This highlights the superior sampling efficiency of TTTS, which stems directly from the tensor factorization.

\begin{figure}[t]
    \centering
    \includegraphics[width=0.45\textwidth]{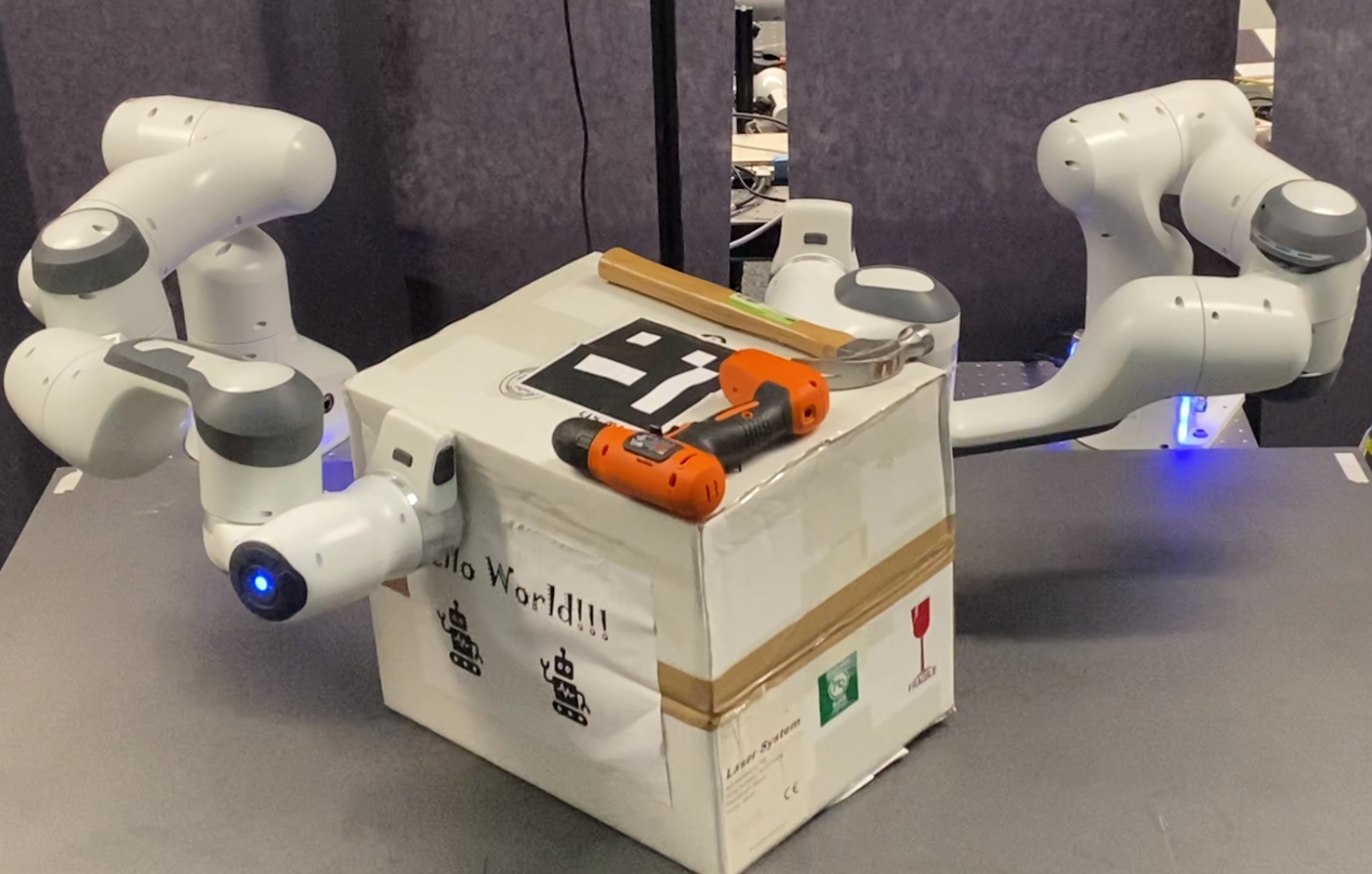}
    \caption{\textbf{Experimental setup with mass perturbation.} A 1.2~kg external mass is introduced to evaluate the robustness of TTTS against model mismatch between the real world and the Genesis simulation.}
    \label{fig:real_add_weight}
\end{figure}

To evaluate the robustness of the proposed method, we further tested it under several variations, including changes in the initial position, adding weight to the box (Figure~\ref{fig:real_add_weight}), combining both variations, and altering the initial orientation. The corresponding box trajectories are presented in Figure~\ref{fig:ttts_robustness}. We observe that changes in the initial position and orientation have little effect on TTTS performance, thanks to the feedback mechanism embedded in MPC. However, adding weight results in slightly worse performance because of the sim-to-real gap. Nevertheless, even with this gap, the box can still be reoriented by more than $70^\circ$, which demonstrates the robustness of our approach under model uncertainty. More tasks and comparisons are presented in the accompanying video.

\section{Conclusion and Future Work}

In this work, we present \textit{Tensor Train Tree Search} (TTTS), which combines tensor factorization with Monte Carlo Tree Search for generalized robot optimization. The key idea is to represent the full decision tree as a high-dimensional tensor and obtain a compact TT approximation covering all candidate branches. This TT model serves as an informative global guide, in which node values at every layer are efficiently computed via TT contractions, concentrating early exploration near high-value regions. Furthermore, UCB-driven strategic search in MCTS is employed for asymptotic global convergence. We demonstrate the effectiveness of TTTS across diverse domains in simulation, including inverse kinematics, motion planning around obstacles, planar pushing with face switching, and legged robot manipulation. Furthermore, a real-world experiment on bimanual whole-body manipulation highlights the practical efficiency of TTTS in enabling fast sampling-based model predictive control.

While these results establish the promise of TTTS, several open challenges remain. TTTS is currently formulated under a deterministic action execution assumption, where each action sequence yields a consistent system output, which is a natural assumption for standard model-based planning and control. However, this assumption may not hold in the presence of real-world stochasticity. Looking ahead, extending TTTS to stochastic settings (e.g., via chance-constraint formulations, robust cost objectives, or belief-space tree search) represents a promising direction for further broadening its applicability.

In this work, we apply TT-Cross to obtain the TT model by actively querying function values, but TT-Cross faces scalability issues in very high-dimensional settings (e.g., with visual observations). To address this limitation, TT cores can be learned via neural networks in a data-driven manner \citep{dolgov2023data, steinlechner2016riemannian, novikov2021tensor}.

Beyond TT, we can also explore improvements to the tree representation itself. In this work we employ the standard TT format to approximate the decision tree and exploit its separable structure, which improves both search efficiency and memory usage. For future work, we aim to investigate Quantized Tensor Train (QTT) decomposition \citep{dolgov2012fast}, which provides a multi-resolution embedding by reshaping each dimension into a sequence of binary (or small) modes. This hierarchical representation aligns naturally with the layered structure of decision trees, enabling coarse-to-fine reasoning and facilitating more efficient branch-and-bound procedures that can significantly reduce computation time for global optimization.

In our current implementation, Genesis \citep{Genesis} is used as the simulation model for planning and control. While it offers general-purpose flexibility and reliable physics, its rollout speed remains a computational bottleneck, limiting scalability for tasks that require extensive sampling or rapid execution. A promising direction for future work is to integrate learned dynamics models as lightweight surrogates, enabling significantly faster rollouts. Such models could greatly improve computational efficiency and expand the applicability of our approach.


Finally, our current representation of discrete modes is limited to integer encoding, which could be extended to symbolic forms such as first-order logic. Since symbolic transitions often introduce geometric constraints, decision-making requires integrated logic-geometric programming \citep{toussaint2015logic}. TTTS naturally supports the joint modeling of logic and geometric variables, offering improved computational efficiency over traditional hierarchical frameworks.

\begin{acks}
This work was supported by the State Secretariat for Education, Research and Innovation in Switzerland for participation in the European Commission's Horizon Europe Program through the INTELLIMAN project (\url{https://intelliman-project.eu/}, HORIZON-CL4-Digital-Emerging Grant 101070136) and the SESTOSENSO project (\url{http://sestosenso.eu/}, HORIZON-CL4-Digital-Emerging Grant 101070310), and by the China Scholarship Council (grant No.202106230104). 
\end{acks}

\bibliographystyle{plainnat}
\bibliography{references}

@string{IROS    		= "Proc. {IEEE/RSJ} Intl Conf. on Intelligent Robots and Systems ({IROS})"}

@string{ICRA    		= "Proc. {IEEE} Intl Conf. on Robotics and Automation ({ICRA})"}

@string{ICML				= "Proc. Intl Conf. on Machine Learning ({ICML})"}

@string{RSS				= "Proc. Robotics: Science and Systems ({R:SS})"}

@string{AAAI				= "Proc. {AAAI} Conference on Artificial Intelligence"}

@String{IJRR = "International Journal of Robotics Research (IJRR)"}

@String{RAL = "{IEEE} Robotics and Automation Letters ({RA-L})"}

@String{TRO = "IEEE Transactions on Robotics"}

@article{oseledets2011tensor,
	title={Tensor-train decomposition},
	author={Oseledets, Ivan V},
	journal={SIAM Journal on Scientific Computing},
	volume={33},
	number={5},
	pages={2295--2317},
	year={2011},
	publisher={SIAM}
}

@article{shetty2016tensor,
	author={Shetty, Suhan and Lembono, Teguh and L\"ow, Tobias and Calinon, Sylvain},
	title={Tensor Train for Global Optimization Problems in Robotics},
	journal=IJRR,
	year={2024},
	volume={43},
	number={6},
	pages={811--839},
	doi={10.1177/02783649231217527}
}

@INPROCEEDINGS{Xue24RSS, 
    AUTHOR    = {Xue, Teng and Razmjoo, Amirreza and Shetty, Suhan and Calinon, Sylvain}, 
    TITLE     = {{Logic-Skill Programming: An Optimization-based Approach to Sequential Skill Planning}}, 
    BOOKTITLE = {Proc.\ Robotics: Science and Systems ({RSS})}, 
    YEAR      = {2024}
}

@inproceedings{xue2023guided,
  title={Demonstration-guided optimal control for long-term non-prehensile planar manipulation},
  author={Xue, Teng and Girgin, Hakan and Lembono, Teguh Santoso and Calinon, Sylvain},
  booktitle=ICRA,
  pages={4999--5005},
  year={2023}
}

@article{dolgov2023data,
  title={Data-driven tensor train gradient cross approximation for hamilton--jacobi--bellman equations},
  author={Dolgov, Sergey and Kalise, Dante and Saluzzi, Luca},
  journal={SIAM Journal on Scientific Computing},
  volume={45},
  number={5},
  pages={A2153--A2184},
  year={2023},
  publisher={SIAM}
}

@article{oseledets2010_ttcross1,
	title={{TT}-cross approximation for multidimensional arrays},
	author={Oseledets, Ivan and Tyrtyshnikov, Eugene},
	journal={Linear Algebra and its Applications},
	volume={432},
	number={1},
	pages={70--88},
	year={2010},
	publisher={Elsevier}
}

@Article{savostyanov2011_ttcross2,
	Title                    = {Fast adaptive interpolation of multi-dimensional arrays in tensor train format},
	Author                   = {Dmitry V. Savostyanov and Ivan V. Oseledets},
	Journal                  = {The 2011 International Workshop on Multidimensional (nD) Systems},
	Year                     = {2011},
	Pages                    = {1-8}
}

@article{pang2023global,
  title={Global planning for contact-rich manipulation via local smoothing of quasi-dynamic contact models},
  author={Pang, Tao and Suh, HJ Terry and Yang, Lujie and Tedrake, Russ},
  journal=TRO,
  year={2023},
  publisher={IEEE}
}

@inproceedings{Xue24CORL,
	author={Xue, Teng and Razmjoo, Amirreza and Shetty, Suhan and Calinon, Sylvain},
	title={Robust Manipulation Primitive Learning via Domain Contraction},
	booktitle={Proc.\ Conference on Robot Learning ({CoRL})},
	year={2024}
}

@article{mason1986mechanics,
  title={Mechanics and planning of manipulator pushing operations},
  author={Mason, Matthew T},
  journal=IJRR,
  volume={5},
  number={3},
  pages={53--71},
  year={1986},
  publisher={Sage Publications Sage CA: Thousand Oaks, CA}
}

@article{mason1999progress,
  title={Progress in nonprehensile manipulation},
  author={Mason, Matthew T},
  journal=IJRR,
  volume={18},
  number={11},
  pages={1129--1141},
  year={1999},
  publisher={SAGE Publications}
}

@inproceedings{li2004iterative,
  title={Iterative linear quadratic regulator design for nonlinear biological movement systems},
  author={Li, Weiwei and Todorov, Emanuel},
  booktitle={International Conference on Informatics in Control, Automation and Robotics},
  volume={2},
  pages={222--229},
  year={2004},
  organization={SciTePress}
}

@article{mayne1966second,
  title={A second-order gradient method for determining optimal trajectories of non-linear discrete-time systems},
  author={Mayne, David},
  journal={International Journal of Control},
  volume={3},
  number={1},
  pages={85--95},
  year={1966},
  publisher={Taylor \& Francis}
}

@inproceedings{tassa2012synthesis,
  title={Synthesis and stabilization of complex behaviors through online trajectory optimization},
  author={Tassa, Yuval and Erez, Tom and Todorov, Emanuel},
  booktitle={2012 IEEE/RSJ International Conference on Intelligent Robots and Systems},
  pages={4906--4913},
  year={2012},
  organization={IEEE}
}

@inproceedings{ratliff2009chomp,
  title={CHOMP: Gradient optimization techniques for efficient motion planning},
  author={Ratliff, Nathan and Zucker, Matt and Bagnell, J Andrew and Srinivasa, Siddhartha},
  booktitle=ICRA,
  pages={489--494},
  year={2009}
}

@article{schulman2014motion,
  title={Motion planning with sequential convex optimization and convex collision checking},
  author={Schulman, John and Duan, Yan and Ho, Jonathan and Lee, Alex and Awwal, Ibrahim and Bradlow, Henry and Pan, Jia and Patil, Sachin and Goldberg, Ken and Abbeel, Pieter},
  journal=IJRR,
  volume={33},
  number={9},
  pages={1251--1270},
  year={2014},
  publisher={SAGE Publications Sage UK: London, England}
}

@book{lavalle2006planning,
  title={Planning algorithms},
  author={LaValle, Steven M},
  year={2006},
  publisher={Cambridge university press}
}

@article{kavraki1996probabilistic,
  title={Probabilistic roadmaps for path planning in high-dimensional configuration spaces},
  author={Kavraki, Lydia E and Svestka, Petr and Latombe, J-C and Overmars, Mark H},
  journal={IEEE transactions on Robotics and Automation},
  volume={12},
  number={4},
  pages={566--580},
  year={1996}
}

@article{karaman2011sampling,
  title={Sampling-based algorithms for optimal motion planning},
  author={Karaman, Sertac and Frazzoli, Emilio},
  journal=IJRR,
  volume={30},
  number={7},
  pages={846--894},
  year={2011},
  publisher={Sage Publications Sage UK: London, England}
}

@inproceedings{gammell2015batch,
  title={Batch informed trees (BIT*): Sampling-based optimal planning via the heuristically guided search of implicit random geometric graphs},
  author={Gammell, Jonathan D and Srinivasa, Siddhartha S and Barfoot, Timothy D},
  booktitle=ICRA,
  pages={3067--3074},
  year={2015}
}

@article{hart1968formal,
  title={A formal basis for the heuristic determination of minimum cost paths},
  author={Hart, Peter E and Nilsson, Nils J and Raphael, Bertram},
  journal={IEEE transactions on Systems Science and Cybernetics},
  volume={4},
  number={2},
  pages={100--107},
  year={1968}
}

@inproceedings{coulom2006efficient,
  title={Efficient selectivity and backup operators in Monte-Carlo tree search},
  author={Coulom, R{\'e}mi},
  booktitle={International conference on computers and games},
  pages={72--83},
  year={2006},
  organization={Springer}
}

@article{browne2012survey,
  title={A survey of monte carlo tree search methods},
  author={Browne, Cameron B and Powley, Edward and Whitehouse, Daniel and Lucas, Simon M and Cowling, Peter I and Rohlfshagen, Philipp and Tavener, Stephen and Perez, Diego and Samothrakis, Spyridon and Colton, Simon},
  journal={IEEE Transactions on Computational Intelligence and AI in games},
  volume={4},
  number={1},
  pages={1--43},
  year={2012},
  publisher={IEEE}
}

@article{silver2017mastering,
  title = {Mastering the game of go without human knowledge},
  author = {
    Silver, David and Schrittwieser, Julian and Simonyan, Karen and Antonoglou, Ioannis
    and Huang, Aja and Guez, Arthur and Hubert, Thomas and Baker, Lucas and Lai, Matthew
    and Bolton, Adrian and Chen, Yutian and Lillicrap, Timothy P. and Hui, Fan and Sifre, Laurent
    and van den Driessche, George and Graepel, Thore and Hassabis, Demis
  },
  journal = {Nature},
  volume = {550},
  number = {7676},
  pages = {354–359},
  year = {2017},
  publisher = {Nature Publishing Group UK London}
}

@book{choset2005principles,
  title={Principles of robot motion: theory, algorithms, and implementations},
  author={Choset, Howie and Lynch, Kevin M and Hutchinson, Seth and Kantor, George A and Burgard, Wolfram},
  year={2005},
  publisher={MIT press}
}

@article{posa2014direct,
  title={A direct method for trajectory optimization of rigid bodies through contact},
  author={Posa, Michael and Cantu, Cecilia and Tedrake, Russ},
  journal=IJRR,
  volume={33},
  number={1},
  pages={69--81},
  year={2014},
  publisher={Sage Publications Sage UK: London, England}
}

@inproceedings{deits2015computing,
  title={Computing large convex regions of obstacle-free space through semidefinite programming},
  author={Deits, Robin and Tedrake, Russ},
  booktitle={Workshop on the Algorithmic Foundations of Robotics (WAFR)},
  pages={109--124},
  year={2015},
  organization={Springer}
}

@article{anthony2017thinking,
  title={Thinking fast and slow with deep learning and tree search},
  author={Anthony, Thomas and Tian, Zheng and Barber, David},
  journal={Advances in neural information processing systems},
  volume={30},
  year={2017}
}

@inproceedings{kim2020monte,
  title={Monte carlo tree search in continuous spaces using voronoi optimistic optimization with regret bounds},
  author={Kim, Beomjoon and Lee, Kyungjae and Lim, Sungbin and Kaelbling, Leslie and Lozano-P{\'e}rez, Tom{\'a}s},
  booktitle={Proceedings of the AAAI Conference on Artificial Intelligence},
  volume={34},
  number={06},
  pages={9916--9924},
  year={2020}
}

@inproceedings{doshi2020hybrid,
  title={Hybrid differential dynamic programming for planar manipulation primitives},
  author={Doshi, Neel and Hogan, Francois R and Rodriguez, Alberto},
  booktitle=ICRA,
  pages={6759--6765},
  year={2020},
}

@article{Xue25IJRR,
	author={Xue, Teng and Razmjoo, Amirreza and Shetty, Suhan and Calinon, Sylvain},
	title={Robust Contact-rich Manipulation through Implicit Motor Adaptation},
	journal={International Journal of Robotics Research ({IJRR})},
	year={2025},
}

@article{riviere2024monte,
  title={Monte Carlo tree search with spectral expansion for planning with dynamical systems},
  author={Rivi{\`e}re, Benjamin and Lathrop, John and Chung, Soon-Jo},
  journal={Science Robotics},
  volume={9},
  number={97},
  pages={eado1010},
  year={2024},
  publisher={American Association for the Advancement of Science}
}

@article{marcucci2023motion,
  title={Motion planning around obstacles with convex optimization},
  author={Marcucci, Tobia and Petersen, Mark and von Wrangel, David and Tedrake, Russ},
  journal={Science robotics},
  volume={8},
  number={84},
  pages={eadf7843},
  year={2023},
  publisher={American Association for the Advancement of Science}
}

@article{goldenberg2003complete,
  title={A complete generalized solution to the inverse kinematics of robots},
  author={Goldenberg, Andrew and Benhabib, Beno and Fenton, Robert},
  journal={IEEE Journal on Robotics and Automation},
  volume={1},
  number={1},
  pages={14--20},
  year={2003}
}

@article{lembono2020memory,
  title={Memory of motion for warm-starting trajectory optimization},
  author={Lembono, Teguh Santoso and Paolillo, Antonio and Pignat, Emmanuel and Calinon, Sylvain},
  journal=RAL,
  volume={5},
  number={2},
  pages={2594--2601},
  year={2020}
}

@article{lin2017sampling,
  title={Sampling-based path planning for UAV collision avoidance},
  author={Lin, Yucong and Saripalli, Srikanth},
  journal={IEEE Transactions on Intelligent Transportation Systems},
  volume={18},
  number={11},
  pages={3179--3192},
  year={2017}
}

@inproceedings{moura2022non,
  title={Non-prehensile planar manipulation via trajectory optimization with complementarity constraints},
  author={Moura, Jo{\~a}o and Stouraitis, Theodoros and Vijayakumar, Sethu},
  booktitle=ICRA,
  pages={970--976},
  year={2022}
}

@article{garrett2021integrated,
  title={Integrated task and motion planning},
  author={Garrett, Caelan Reed and Chitnis, Rohan and Holladay, Rachel and Kim, Beomjoon and Silver, Tom and Kaelbling, Leslie Pack and Lozano-P{\'e}rez, Tom{\'a}s},
  journal={Annual review of control, robotics, and autonomous systems},
  volume={4},
  number={1},
  pages={265--293},
  year={2021},
  publisher={Annual Reviews}
}

@inproceedings{toussaint2015logic,
	title={Logic-geometric programming: an optimization-based approach to combined task and motion planning},
	author={Toussaint, Marc},
	booktitle={Proceedings of the 24th International Conference on Artificial Intelligence},
	pages={1930--1936},
	year={2015}
}

@inproceedings{Xue24ICRA,
	author={Xue, Teng and Razmjoo, Amirreza and Calinon, Sylvain},
	title={{D-LGP}: Dynamic Logic-Geometric Program for Reactive Task and Motion Planning},
	booktitle=ICRA,
	year={2024},
	pages={14888--14894}
}

@article{holladay2024robust,
  title={Robust planning for multi-stage forceful manipulation},
  author={Holladay, Rachel and Lozano-P{\'e}rez, Tom{\'a}s and Rodriguez, Alberto},
  journal=IJRR,
  volume={43},
  number={3},
  pages={330--353},
  year={2024},
  publisher={SAGE Publications Sage UK: London, England}
}

@software{Genesis,
  author = {
    Zhou, Xian and Qiao, Yiling and Xu, Zhenjia and Wang, Tsun-Hsuan and Chen, Zhehuan and Zheng, Juntian and Xiong, Ziyan and Wang, Yian and Zhang, Mingrui and Ma, Pingchuan and Wang, Yufei and Dou, Zhiyang and Kim, Byungchul and Tian, Yunsheng and Chen, Yipu and Qiu, Xiaowen and Lin, Chunru and He, Tairan and Si, Zilin and Zhang, Yunchu and Yang, Zhanlue and Liu, Tiantian and Li, Tianyu and Yamazaki, Kashu and Zhang, Hongxin and Ha, Huy and Zhang, Yu and Liu, Michael and Zheng, Shaokun and Fu, Zipeng and Wu, Qi and Geng, Yiran and Chen, Feng and Hu, Yuanming and Shi, Guanya and Liu, Lingjie and Komura, Taku and Erickson, Zackory and Held, David and Li, Minchen and Fan (Jim), Linxi and Zhu, Yuke and Matusik, Wojciech and Gutfreund, Dan and Song, Shuran and Rus, Daniela and Lin, Ming and Zhu, Bo and Fragkiadaki, Katerina and Gan, Chuang
  },
  title = {Genesis: A Generative and Universal Physics Engine for Robotics and Beyond},
  month = {December},
  year = {2024},
  url = {https://github.com/Genesis-Embodied-AI/Genesis}
}

@article{hansen2003reducing,
  title={Reducing the time complexity of the derandomized evolution strategy with covariance matrix adaptation (CMA-ES)},
  author={Hansen, Nikolaus and M{\"u}ller, Sibylle D and Koumoutsakos, Petros},
  journal={Evolutionary computation},
  volume={11},
  number={1},
  pages={1--18},
  year={2003},
  publisher={MIT Press}
}

@article{wang2017constrained,
  title={Constrained trajectory optimization for planetary entry via sequential convex programming},
  author={Wang, Zhenbo and Grant, Michael J},
  journal={Journal of Guidance, Control, and Dynamics},
  volume={40},
  number={10},
  pages={2603--2615},
  year={2017},
  publisher={American Institute of Aeronautics and Astronautics}
}

@article{malyuta2022convex,
  title={Convex optimization for trajectory generation: A tutorial on generating dynamically feasible trajectories reliably and efficiently},
  author={Malyuta, Danylo and Reynolds, Taylor P and Szmuk, Michael and Lew, Thomas and Bonalli, Riccardo and Pavone, Marco and A{\c{c}}{\i}kme{\c{s}}e, Beh{\c{c}}et},
  journal={IEEE Control Systems Magazine},
  volume={42},
  number={5},
  pages={40--113},
  year={2022},
  publisher={IEEE}
}

@article{dolgov2012fast,
  title={Fast solution of parabolic problems in the tensor train/quantized tensor train format with initial application to the Fokker--Planck equation},
  author={Dolgov, Sergey V and Khoromskij, Boris N and Oseledets, Ivan V},
  journal={SIAM Journal on Scientific Computing},
  volume={34},
  number={6},
  pages={A3016--A3038},
  year={2012},
  publisher={SIAM}
}

@inproceedings{chaslot2008parallel,
  title={Parallel monte-carlo tree search},
  author={Chaslot, Guillaume MJ -B and Winands, Mark HM and van Den Herik, H Jaap},
  booktitle={Computers and Games: 6th International Conference, CG 2008, Beijing, China, September 29-October 1, 2008. Proceedings 6},
  pages={60--71},
  year={2008},
  organization={Springer}
}

@inproceedings{todorov2012mujoco,
  title={Mujoco: A physics engine for model-based control},
  author={Todorov, Emanuel and Erez, Tom and Tassa, Yuval},
  booktitle={2012 IEEE/RSJ international conference on intelligent robots and systems},
  pages={5026--5033},
  year={2012},
  organization={IEEE}
}

@inproceedings{liang2018gpu,
  title={Gpu-accelerated robotic simulation for distributed reinforcement learning},
  author={Liang, Jacky and Makoviychuk, Viktor and Handa, Ankur and Chentanez, Nuttapong and Macklin, Miles and Fox, Dieter},
  booktitle={Conference on Robot Learning},
  pages={270--282},
  year={2018},
  organization={PMLR}
}

@article{cichocki2016tensor,
  title={Tensor networks for dimensionality reduction and large-scale optimization: Part 1 low-rank tensor decompositions},
  author={Cichocki, Andrzej and Lee, Namgil and Oseledets, Ivan and Phan, Anh-Huy and Zhao, Qibin and Mandic, Danilo P.},
  journal={Foundations and Trends{\textregistered} in Machine Learning},
  volume={9},
  number={4-5},
  pages={249--429},
  year={2016},
  publisher={Now Publishers, Inc.}
}

@article{virtanen2020scipy,
  title={SciPy 1.0: fundamental algorithms for scientific computing in Python},
  author={Virtanen, Pauli and Gommers, Ralf and Oliphant, Travis E and Haberland, Matt and Reddy, Tyler and Cournapeau, David and Burovski, Evgeni and Peterson, Pearu and Weckesser, Warren and Bright, Jonathan and van der Walt, St{\'e}fan J and Brett, Matthew and Wilson, Joshua and Millman, K Jarrod and Mayorov, Nikolay and Nelson, Andrew R J and Jones, Eric and Kern, Robert and Larson, Eric and Carey, C J and {SciPy 1.0 Contributors}},
  journal={Nature Methods},
  volume={17},
  number={3},
  pages={261--272},
  year={2020},
  publisher={Nature Publishing Group US New York}
}

@article{marcucci2024shortest,
  title={Shortest paths in graphs of convex sets},
  author={Marcucci, Tobia and Umenberger, Jack and Parrilo, Pablo and Tedrake, Russ},
  journal={SIAM Journal on Optimization},
  volume={34},
  number={1},
  pages={507--532},
  year={2024},
  publisher={SIAM}
}

@inproceedings{jankowski2023vp,
  title={VP-STO: Via-point-based Stochastic Trajectory Optimization for Reactive Robot Behavior},
  author={Jankowski, Julius and Bruderm{\"u}ller, Lara and Hawes, Nick and Calinon, Sylvain},
  booktitle=ICRA,
  pages={10125--10131},
  year={2023}
}

@article{gasparetto2015path,
  title={Path planning and trajectory planning algorithms: A general overview},
  author={Gasparetto, Alessandro and Boscariol, Paolo and Lanzutti, Albano and Vidoni, Renato},
  journal={Motion and operation planning of robotic systems: Background and practical approaches},
  pages={3--27},
  year={2015},
  publisher={Springer}
}

@article{steinlechner2016riemannian,
  title={Riemannian optimization for high-dimensional tensor completion},
  author={Steinlechner, Michael},
  journal={SIAM Journal on Scientific Computing},
  volume={38},
  number={5},
  pages={S461--S484},
  year={2016},
  publisher={SIAM}
}

@inproceedings{novikov2021tensor,
  title={Tensor-train density estimation},
  author={Novikov, Georgii S and Panov, Maxim E and Oseledets, Ivan V},
  booktitle={Uncertainty in artificial intelligence},
  pages={1321--1331},
  year={2021},
  organization={PMLR}
}

@inproceedings{zhu2023efficient,
  title={Efficient object manipulation planning with monte carlo tree search},
  author={Zhu, Huaijiang and Meduri, Avadesh and Righetti, Ludovic},
  booktitle=IROS,
  pages={10628--10635},
  year={2023}
}

@article{cheng2023enhancing,
  title={Enhancing dexterity in robotic manipulation via hierarchical contact exploration},
  author={Cheng, Xianyi and Patil, Sarvesh and Temel, Zeynep and Kroemer, Oliver and Mason, Matthew T},
  journal=RAL,
  volume={9},
  number={1},
  pages={390--397},
  year={2023}
}

@article{eisert2010colloquium,
  title={Colloquium: Area laws for the entanglement entropy},
  author={Eisert, Jens and Cramer, Marcus and Plenio, Martin B},
  journal={Reviews of modern physics},
  volume={82},
  number={1},
  pages={277--306},
  year={2010},
  publisher={APS}
}

@article{roy2021machine,
  title={From machine learning to robotics: Challenges and opportunities for embodied intelligence},
  author={
    Roy, Nicholas and
    Posner, Ingmar and
    Barfoot, Tim and
    Beaudoin, Philippe and
    Bengio, Yoshua and
    Bohg, Jeannette and
    Brock, Oliver and
    Dépatie, Isabelle and
    Fox, Dieter and
    Koditschek, Dan and
    Lozano-Pérez, Tomás and
    Mansinghka, Vikash and
    Pal, Christopher and
    Richards, Blake and
    Sadigh, Dorsa and
    Schaal, Stefan and
    Sukhatme, Gaurav and
    Thérien, Denis and
    Toussaint, Marc and
    Van de Panne, Michiel
  },
  journal={arXiv preprint arXiv:2110.15245},
  year={2021}
}

@article{storn1997differential,
  title={Differential evolution--a simple and efficient heuristic for global optimization over continuous spaces},
  author={Storn, Rainer and Price, Kenneth},
  journal={Journal of global optimization},
  volume={11},
  number={4},
  pages={341--359},
  year={1997},
  publisher={Springer}
}

@inproceedings{guez2018learning,
  title={Learning to search with mctsnets},
  author={Guez, Arthur and Weber, Th{\'e}ophane and Antonoglou, Ioannis and Simonyan, Karen and Vinyals, Oriol and Wierstra, Daan and Munos, R{\'e}mi and Silver, David},
  booktitle=ICML,
  pages={1822--1831},
  year={2018}
}

@article{kemmerling2024beyond,
  title={Beyond games: a systematic review of neural Monte Carlo tree search applications},
  author={Kemmerling, Marco and L{\"u}tticke, Daniel and Schmitt, Robert H},
  journal={Applied Intelligence},
  volume={54},
  number={1},
  pages={1020--1046},
  year={2024},
  publisher={Springer}
}

@inproceedings{kocsis2006bandit,
  title={Bandit based {M}onte-{C}arlo planning},
  author={Kocsis, Levente and Szepesv{\'a}ri, Csaba},
  booktitle={European Conference on Machine Learning (ECML)},
  pages={282--293},
  year={2006},
  publisher={Springer}
}

@article{goreinov1997theory,
  title={A theory of pseudoskeleton approximations},
  author={Goreinov, Sergei A. and Tyrtyshnikov, Eugene E. and Zamarashkin, Nickolai L.},
  journal={Linear Algebra and its Applications},
  volume={261},
  number={1--3},
  pages={1--21},
  year={1997},
  publisher={Elsevier}
}

\appendix
\section{Appendix}

\subsection{Experimental hyperparameter}

In our experiments, we utilized an NVIDIA GeForce RTX 3090 GPU with 24 GB of memory. The tolerance for the TT-Cross approximation was set to $\epsilon = 10^{-3}$. Table~\ref{tab:hyperparam} summarizes the hyperparameters applied across different tasks. \textit{Batch size} indicates the maximum number of parallel environments; \textit{Num. MCTS} denotes the number of independent MCTS instances executed concurrently; and \textit{Num. Sim.} specifies the number of parallel environments used during the simulation phase. \textit{Num. Discret.} refers to the discretization granularity of the state and action spaces. The parameter $r_{\text{max}}$ defines the maximum TT rank employed during TT-Tree initialization via TT-Cross. \textit{Pop. Size} and \textit{CMA-ES Iter.} are the population size and iteration count used in the CMA-ES refinement step, respectively.

\begin{table*}
\centering
\caption{\textbf{Hyperparameters employed for different tasks.}}
\resizebox{\textwidth}{!}{\begin{tabular}{cc|c|c|c|c|c|c|c} \label{tab:learning_data}
      & Task & Batch Size & Num. MCTS & Num. Discret. & Num. Sim. &$r_{max}$ & Pop. Size & CMA-ES Iter.\\
     \hline
     & {3-joint IK} & {$1000$}& {{$5$}} & {{$20$}} & {{$1000$}} & {$21$} & $25$ & $20$\\
     &{7-joint IK} & {$500$} & {{$5$}} & {{$20$}} & {{$1000$}} & {$21$} & $25$ & $20$\\
     &{3-joint MP} & {$1000$} & {{$5$}} & {{$20$}} & $1000$ & {$41$} & $25$ & $20$\\
     &7-joint MP    & $500$   & $5$       & {$20$} & $1000$ & $21$ & {$25$} & $20$\\
     &LegMani & $500$ &$5$ &$20$ &$1000$ &$21$ &$25$ & $20$\\
     & Multi-stage MP & $1000$ & $5$ & $50$ & $1000$ & {$41$} & {$25$} & $20$\\
     &Whole-body Manipulation & $500$ & $1$& $20$ & $500$ & {$21$} & {$25$} & $20$\\
     \hline
\end{tabular}}
\label{tab:hyperparam}
\end{table*}

\subsection{\minew{Implementation Details of the \minew{AlphaZero-MCTS} Baseline}}
\label{sec:app_neural_mcts}

\minew{Table~\ref{tab:neural_mcts_details} lists the implementation details of the \minew{AlphaZero-MCTS} baseline of Fig.~\ref{fig:cost_time_comparison} side by side with those of TTTS. The baseline follows AlphaZero \citep{silver2017mastering} on the same decision tree as TTTS, where each level assigns one decision variable. Its policy and value networks are pretrained by self-play and are kept fixed during online search. As in TTTS, the best solutions found by the search are refined by CMA-ES over the continuous B-spline weights, with the contact-face IDs fixed.}

\begin{table*}
\centering
\caption{\minew{\textbf{Implementation details of TTTS and the \minew{AlphaZero-MCTS} baseline on the planar pushing task} (Fig.~\ref{fig:cost_time_comparison}). Query counts refer to evaluations of the true objective $J$, i.e., simulator rollouts.}}
\label{tab:neural_mcts_details}
\small
\renewcommand{\arraystretch}{1.15}
\begin{tabular}{>{\raggedright\arraybackslash}p{0.19\textwidth}|>{\raggedright\arraybackslash}p{0.36\textwidth}|>{\raggedright\arraybackslash}p{0.36\textwidth}}
\hline
 & \minew{\textbf{TTTS (ours)}} & \minew{\textbf{\minew{AlphaZero-MCTS}}} \\
\hline
\minew{Learned model} & \minew{TT approximation of the objective $J(\bm{z}, \bm{x})$} & \minew{Policy network $P_\theta(a \mid s, \bm{z})$ and value network $V_\phi(s, \bm{z})$} \\
\minew{State space} & \minew{$(x, y, \theta, p_x, p_y, c)$: slider pose, pusher position, current contact face; initial $(x, y, \theta)$ used as context} & \minew{Same} \\
\minew{Action space} & \minew{$c_1,\dots,c_4 \in \{0,1,2,3\}$ (contact face per segment) and $4$ B-spline weights}
 & \minew{Same variables} \\
\minew{Tree search rule} & \minew{UCB, Eq.~\eqref{eq:ucb}, $c=3$} & \minew{PUCT, $Q(s,a) + c\,P_\theta(a \mid s)\sqrt{N(s)}/(1+N(s,a))$, $c=3$} \\
\minew{Architecture and model size} & \minew{TT with maximal rank 41, $1.8\times10^5$ parameters} & \minew{Policy and value MLPs with $9.2\times10^4$ (policy) and $8.8\times10^4$ (value) parameters} \\
\minew{Offline budget: queries of $J$ / time} & \minew{$4.7\times10^5$ / 10.5\,s (TT-Cross)} & \minew{$\approx 4.1\times10^7$ / 4.77\,h (self-play)} \\
\minew{Online budget: queries of $J$ per scenario} & \minew{$\approx 2\times10^4$} & \minew{$\approx 2.6\times10^4$} \\
\hline
\end{tabular}
\end{table*}

\subsection{\minew{Setup of the Controlled Comparison of TT and NN Surrogates}}
\label{sec:app_tt_vs_nn}

\minew{Table~\ref{tab:tt_vs_nn_setup} summarizes the experimental setup for the comparison in Section~\ref{sec:tt_vs_nn}. We record all evaluations of $J$ queried during TT-Cross construction and then use these data to train the NN surrogates. The 1$\times$ NN is an MLP with three hidden layers and the same number of parameters as the TT. The 10$\times$ and 20$\times$ NNs are MLPs with six hidden layers and ten and twenty times as many parameters, respectively, and the 50$\times$ NN is a residual MLP with twelve hidden layers and fifty times as many parameters.}

\minew{For the node-value accuracy in Fig.~\ref{fig:tt_vs_nn}(b, d), we evaluate nodes along high-quality solutions and compare the estimated values of their children with reference values computed using the true objective. Accuracy is measured by the best-child accuracy, i.e., the fraction of nodes for which the child with the highest estimated value is also the child with the highest reference value.}



\begin{table*}[t]
\centering
\small
\renewcommand{\arraystretch}{1.15}
\caption{\minew{\textbf{Setup of the controlled comparison of TT and NN surrogates} (Section~\ref{sec:tt_vs_nn}).}}
\label{tab:tt_vs_nn_setup}
\begin{tabular}{>{\raggedright\arraybackslash}p{0.22\textwidth}|>{\raggedright\arraybackslash}p{0.34\textwidth}|>{\raggedright\arraybackslash}p{0.34\textwidth}}
\hline
 & \minew{\textbf{Face-switching planar pushing}} & \minew{\textbf{3-joint manipulator motion planning}} \\
\hline
\minew{Decision variables} & \minew{4 contact-face IDs (4 values) and 4 B-spline weights} & \minew{12 basis weights} \\
\minew{Training data (shared)} & \minew{$3.8\times10^5$ TT-Cross queries} & \minew{$1.2\times10^6$ TT-Cross queries} \\
\minew{Parameters (TT)} & \minew{$1.48\times10^5$} & \minew{$2.6\times10^5$} \\
\minew{Parameters (NN 1$\times$ / 10$\times$ / 20$\times$ / 50$\times$)} & \minew{$1.48\times10^5$ / $1.48\times10^6$ / $2.96\times10^6$ / $7.4\times10^6$} & \minew{$2.6\times10^5$ / $2.6\times10^6$ / $5.2\times10^6$ / $1.3\times10^7$} \\
\minew{NN hidden layers (1$\times$ / 10$\times$ / 20$\times$ / 50$\times$)} & \minew{3 / 6 / 6 / 12 (residual)} & \minew{3 / 6 / 6 / 12 (residual)} \\
\minew{Search (shared)} & \multicolumn{2}{>{\raggedright\arraybackslash}p{0.68\textwidth}}{\minew{UCB of Eq.~\eqref{eq:ucb} with $c=3$}} \\
\hline
\end{tabular}
\end{table*}

\subsection{Cost Function Used in Experiments}
\label{sec: cost_functions}

\paragraph{\textbf{Continuous nonconvex function.}} To illustrate the capability of TTTS in handling nonconvex optimization problems and its advantage over TTGO, we construct the following full-rank two-dimensional function as a toy example. The performance of TTTS on this function is reported in Section~\ref{sec:cont_opt}.
\begin{equation}
\begin{aligned}
    f_1(\bm{x}) &= -\tfrac{1}{2}\,(0.8z - 2\,\operatorname{sign}(z))^2 + 0.1 \|\bm{x}\|_2^2 + 2, \\ 
    \bm{x} &= (x_1, x_2)^\trsp, \quad z = \tfrac{x_1+x_2}{\sqrt{2}}
\end{aligned}
\label{eq: cont_func}
\end{equation}

\paragraph{\textbf{Mixed-integer nonconvex function.}}
MsMP problems can be generally formulated as mixed-integer nonconvex programs.  
To demonstrate the effectiveness of TTTS in addressing such problems,  
we consider the following piecewise-defined mixed-integer nonconvex function:  

\[
\tilde{\bm{x}} =
\begin{bmatrix}
\tilde x_1 \\ \tilde x_2
\end{bmatrix}
= \tfrac{1}{\sqrt{2}}
\begin{bmatrix}
1 & -1 \\ 1 & 1
\end{bmatrix} \bm{x}, 
\]

\begin{equation}
f_2(\bm{x}) = \tfrac{1}{2}\,g(\tilde x_1) + 0.1\,h(\tilde x_1,\tilde x_2),
\label{eq: mix_func}
\end{equation}

where the piecewise components are defined as
\[
g(\tilde x_1) =
\begin{cases}
|-(\tilde x_1+5)^2+8|, & -7 \le \tilde x_1 < -3,\\
|-(\tilde x_1+1)^2+3|, & -3 \le \tilde x_1 < 1,\\
|-(\tilde x_1-3)^2+4|, & 1 \le \tilde x_1 \le 5,\\
|-(\tilde x_1-7)^2+5|, & 5 < \tilde x_1 \le 9,\\
|-(\tilde x_1-11)^2+10|, & 9 < \tilde x_1 \le 13,\\
0, & \text{otherwise},
\end{cases}
\]

\[
h(\tilde x_1,\tilde x_2) =
\begin{cases}
(\tilde x_1+5)^2+(\tilde x_2+5)^2, & -7 \le \tilde x_1 < -3,\\
(\tilde x_1+1)^2+(\tilde x_2+1)^2, & -3 \le \tilde x_1 < 1,\\
(\tilde x_1-3)^2+(\tilde x_2-3)^2, & 1 \le \tilde x_1 \le 5,\\
(\tilde x_1-7)^2+(\tilde x_2-7)^2, & 5 < \tilde x_1 \le 9,\\
(\tilde x_1-11)^2+(\tilde x_2-11)^2, & 9 < \tilde x_1 \le 13,\\
0, & \text{otherwise}.
\end{cases}
\]

subject to the constraints
\[
 \bm{x} = (x_1, x_2)^\trsp, \quad x_1 \in \{0,1,\dots,10\}, \quad x_2 \in [-5,5].
\]

The performance of TTTS on this function is reported in Section~\ref{sec:mi_opt}.

\paragraph{\textbf{Cost function for inverse kinematics.}}

In the inverse kinematics (IK) setting, the goal is to find a joint configuration \( \bm{q} \in \mathbb{R}^n \) such that the robot's end-effector reaches a desired target position while avoiding collisions and maintaining reasonable deviation from the current posture. The total cost function used is defined as:
\[
c_{\text{total}} = 50\, c_{\text{goal}} +  c_{\text{obst}}.
\]

The term \( c_{\text{goal}} \) penalizes the distance between the forward kinematics output of the proposed joint configuration and the desired end-effector position:
\[
c_{\text{goal}} = \left\| \text{eef}_{\text{pos}}(\bm{q}) - \bm{x}_{\text{goal}} \right\|_2.
\]

Here, \( \bm{x}_{\text{goal}} \) is the target pose of the end-effector, and \( \text{eef}_{\text{pos}}(\bm{q}) \) denotes the pose of the end-effector computed using forward kinematics at joint configuration \( \bm{q} \).

\bigskip

The term \( c_{\text{obst}} \) penalizes collisions by summing binary collision indicators for the final robot configuration:
\[
c_{\text{obst}} = \sum_{i=1}^{N_{\text{links}}} \text{collides}_i(\bm{q}), 
\]
where \( \text{collides}_i(\bm{q}) \in \{0, 1\} \) indicates whether the \( i \)-th link is in collision.

\paragraph{\textbf{Cost function for motion planning.}}

The total cost function used for trajectory optimization is composed of three terms: a goal-reaching cost, a collision avoidance cost, and a control smoothness cost. The overall cost is defined as:
\[
c_{\text{total}} = 50 \, c_{\text{goal}} + c_{\text{obst}} + 0.1 \, c_{\text{control}}.
\]

The first term, \( c_{\text{goal}} \), encourages the robot to reach the desired target position. It is computed as the Euclidean distance between the current end-effector position and the goal:
\[
c_{\text{goal}} = \left\| \text{eef}_{\text{pos}} - \bm{x}_{\text{goal}} \right\|_2, 
\]
where \( \text{eef}_{\text{pos}} \) denotes the Cartesian position of the robot's end-effector at the final time step, and \( \bm{x}_{\text{goal}} \) is the desired target position.

The second term, \( c_{\text{obst}} \), penalizes trajectories that result in collisions. It is computed by summing binary collision indicators over the trajectory:
\[
c_{\text{obst}} = \sum_{t=1}^{T} \text{collisions}_t, 
\]
where \( \text{collisions}_t \in \{0, 1\} \) is a binary variable indicating whether a collision occurs at time step \( t \).

The third term, \( c_{\text{control}} \), measures the smoothness of the joint trajectory. It compares the total length of the path in joint space with the direct distance between the start and end configurations:
\[
c_{\text{control}} = 
\frac{ \sum_{t=1}^{T} \left\| \bm{q}_{t} - \bm{q}_{t-1} \right\|_2}
{ \left\| \bm{q}_T - \bm{q}_0 \right\|_2 + \varepsilon }.
\]

Here, \( \bm{q}_t \) represents the robot's joint configuration at time step \( t \), and \( \varepsilon \) is a small positive constant added for numerical stability. We set $\varepsilon = 10^{-6}$ in our experiments.

\paragraph{\textbf{Cost function for legged robot manipulation.}}

In the legged robot manipulation setting, the objective is to find optimal robot trajectories which move the manipulated box toward the target pose. The total cost is defined as
\[
c_{\text{total}} \;=\; c_{\text{orn}} + c_{\text{vel}} + 5\,c_{\text{pos}},
\]
where the three components are given below.

The first term penalizes deviation between the final box orientation and the desired orientation:
\[
c_{\text{orn}} \;=\; 0.1 \, \big\| \mathbf{o}_T^{\text{box}} - \mathbf{o}_{\text{target}} \big\|_2,
\]
where \(o_T^{\text{box}}\) is the yaw angle of the final box orientation.

The second term encourages stable manipulation by penalizing abrupt orientation changes of the box during the trajectory:
\[
c_{\text{vel}} \;=\; \frac{0.1}{T-1} \sum_{t=0}^{T-2} \big|\, o_{t+1}^{\text{box}} - o_t^{\text{box}} \,\big|,
\]
where $o_t^{\text{box}}$ is the yaw angle of the box at time $t$.

The third term penalizes final position error of the box in the horizontal plane:
\[
c_{\text{pos}} \;=\; 0.1 \,\big\| \mathbf{p}_T^{\text{box}} - \mathbf{p}_{\text{target}}\big\|_2,
\]
where \(\mathbf{p}_T^{\text{box}}\) denotes the final box position projected onto the \(xy\)-plane.

\bigskip

The cost function therefore balances final pose accuracy (position and orientation) with the smoothness of the manipulated box trajectory, promoting stable and efficient robot–box interaction.

\paragraph{\textbf{Cost function for multi-stage motion planning.}}

In the planar pushing setting, the objective is to compute a sequence of discrete contact modes and continuous robot velocities that move an object from its initial state to a target pose. The total cost combines two components: a state reaching cost and a control effort cost. It is defined as:
\[
c_{\text{total}} = c_{\text{state}} + 0.01 \, c_{\text{action}}.
\]

\bigskip

The state cost \( c_{\text{state}} \) evaluates how far the final state \( \bm{x}_T \in \mathbb{R}^3 \) (position and orientation of the object) is from the target pose \( \bm{x}_\text{target} \in \mathbb{R}^3 \). It is defined as:
\[
c_{\text{state}} = \left\| \bm{x}_T^{\text{pos}} - \bm{x}_{\text{target}}^{\text{pos}} \right\|_2 + 0.1 \, \left| \theta_T - \theta_{\text{target}} \right|,
\]
where:
\begin{itemize}
  \item \( \bm{x}_T^{\text{pos}} = [x_{1_T}, x_{2_T}]^\top \) is the object's final position
  \item \( \theta_T \) is the final orientation
  \item \( \bm{x}_{\text{target}}^{\text{pos}} \) and \( \theta_{\text{target}} \) are the desired position and orientation
\end{itemize}

\bigskip

The action cost \( c_{\text{action}} \) penalizes the total effort of the pushing trajectory, where each control input \( \mathbf{u}_t \in \mathbb{R}^2 \) represents a planar velocity vector. The total control cost is the sum of the norms of all pushing actions:
\[
c_{\text{action}} = \sum_{k=1}^K \sum_{t=1}^T \left\| \mathbf{u}_t \right\|_2.
\]

These actions are generated based on a discrete contact face selection and continuous parameters defining a trajectory. The discrete mode \( i_t \in \{1, 2, 3, 4\} \) specifies which face of the object is being pushed at each time.

\bigskip

The final cost balances reaching the target pose accurately with minimizing the pushing effort, with a small weight on the latter to avoid excessive motion without overconstraining the optimization.

\paragraph{\textbf{Cost function for whole-body manipulation.}}

In the bimanual whole-body manipulation setting, the objective is to control two arms collaboratively to manipulate an object (e.g., a box) toward a target pose while maintaining effective contact and avoiding awkward configurations. The total cost function is composed of five terms:
\[
c_{\text{total}} = c_{\text{pos}} + 50 \, c_{\text{orn}} + 0.1 \, c_{\text{control}} - c_{\text{contact}} + 5 \, \delta_{\text{eef}}, 
\]
where $c_{\text{pos}}$ penalizes deviation of the box's final position from the target:
\[
c_{\text{pos}} = \left\| \mathbf{p}_T^{\text{box}} - \mathbf{p}_{\text{target}} \right\|_2. 
\]
\( \mathbf{p}_T^{\text{box}} \in \mathbb{R}^3 \) is the final box position and \( \mathbf{p}_{\text{target}} \) is the desired position. $ c_{\text{orn}}$ measures orientation alignment between the final box orientation and the target orientation, namely 
\[
c_{\text{orn}} =  \left|{\theta}_T^{\text{box}}- {\theta}_{\text{target}} \right|.
\]

$c_{\text{control}}$ is a regularization term that penalizes excessive joint movement across the trajectory:
\[
c_{\text{control}} = \sum_{t=0}^{T} \left\| \bm{q}_{t+1} - \bm{q}_{t} \right\|_2,
\]
where \( \bm{q}_t \in \mathbb{R}^n \) is the full-body joint configuration at time step \( t \).

\bigskip

$c_{\text{contact}}$ rewards the maintenance of valid bimanual contact. Let \( c_t^L, c_t^R \in \{0, 1\} \) be binary contact flags for the left and right hands. Then:
\[
c_{\text{contact}} = \sum_{t=1}^{T} \mathbb{I}\left[c_t^L = 1 \land c_t^R = 1\right],
\]
where \( \mathbb{I}[\cdot] \) is the indicator function. This term is subtracted from the total cost to encourage simultaneous dual-arm contact.

\bigskip

$\delta_{\text{eef}}$ is  a binary term that penalizes configurations where the two arms are too close to each other, in order to avoid self-collision:
\[
\delta_{\text{eef}} = \mathbb{I}\left[ \left\| \mathbf{x}^{\text{eef}_0} - \mathbf{x}^{\text{eef}_1} \right\|_2 < 0.3 \right],
\]
where \( \mathbf{x}^{\text{eef}_i} \) is the Cartesian position of the \( i \)-th end-effector.

The cost function encourages accurate placement and orientation of the manipulated object, smooth and efficient joint trajectories, persistent bimanual contact, and physically feasible arm configurations.






\end{document}